\documentclass[Afour,sageh,times]{template/sagej}

\usepackage{moreverb,url}
\usepackage[nolist]{acronym}
 \usepackage{multirow}
 \usepackage{longtable}
 \usepackage{float}
 \usepackage{dsfont}
 \usepackage{algorithmicx}
 \usepackage{algorithm}
 \usepackage{algpseudocode} 
 \usepackage{dsfont} 
 \usepackage{url}
 \usepackage{dblfloatfix}
 \usepackage{placeins}
\usepackage[colorlinks,bookmarksopen,bookmarksnumbered,citecolor=red,urlcolor=red]{hyperref}
\usepackage[dvipsnames]{xcolor}

\usepackage{tabularx}   %
\usepackage{booktabs}   %
\usepackage{makecell}   %
\usepackage{graphicx}   %
\usepackage{subcaption}
\newcolumntype{Y}{>{\centering\arraybackslash}X}
\newcommand{\cmark}{\checkmark}
\newcommand{\xmark}{\(\times\)}
\newcommand{\pmark}{\(\triangle\)} %

\newcommand\BibTeX{{\rmfamily B\kern-.05em \textsc{i\kern-.025em b}\kern-.08em
T\kern-.1667em\lower.7ex\hbox{E}\kern-.125emX}}

\def\volumeyear{2025}

\makeatletter
\renewcommand{\@seccntformat}[1]{\csname the#1\endcsname.\quad}
\makeatother

\begin{document}

\begin{acronym}
    \acro{AR}[AR]{Aerial Vehicle}
    \acro{API}[API]{Application Program Interface}
    \acro{CoM}[CoM]{Center of Mass}
    \acro{DOPE}[DOPE]{Deep Object Pose Estimation}
    \acro{DoF}[DoF]{Degree-of-Freedom}
    \acro{HITL}{Hardware-In-The-Loop}
    \acro{HSI}{Human-Swarm Interaction}
    \acro{ICRA}{International Conference on Robotics and Automation}
    \acro{ML}[ML]{Machine Learning}
    \acro{RL}[RL]{Reinforcement Learning}
    \acro{ROS}[ROS]{Robot Operating System}
    \acro{SITL}{Software-In-The-Loop}
    \acro{UAV}[UAV]{Uncrewed Aerial Vehicle}
    \acro{wrt}[w.r.t.]{with respect to}
    \acro{VTOL}[VTOL]{Vertical Take-Off and Landing}
    \acro{RC}[RC]{Radio Controlled}
    \acro{ODE}[ODE]{Open Dynamics Engine}
    \acro{OS}[OS]{Operating System}
\end{acronym}

\runninghead{Garimella et al.}

\title{HERCULES: An Open-Source Simulation Framework for Heterogeneous Multi-Robot SLAM, Collaborative Perception, and Exploration}

\author{Sandilya Sai Garimella\affilnum{1}, Daniel Chase Butterfield\affilnum{1}, Sean Wilson\affilnum{1,2} and Lu Gan\affilnum{1}}

\affiliation{\affilnum{1}Georgia Institute of Technology, USA\\
\affilnum{2}Georgia Tech Research Institute, USA}

\corrauth{Lu Gan, 
Georgia Institute of Technology, 
Atlanta, 
GA, 
USA.}

\email{lgan@gatech.edu}

\begin{abstract}
We present HERCULES, an open-source simulator and data-collection pipeline for heterogeneous multi-robot autonomy. Built upon the Unreal Engine 5 (UE5)-based simulators AirSim and Cosys-AirSim, HERCULES resolves key architectural limitations of prior frameworks to enable concurrent unmanned aerial and ground vehicle (UAV-UGV) operation in large-scale, photorealistic, dynamic environments. On top of this capability, HERCULES introduces a new waypoint-tracking UGV controller that mirrors existing UAV control interfaces, and provides a shared navigation stack for mapping, traversability analysis, planning, and control across heterogeneous platforms. Expanding upon inherited multi-modal sensor suites, the simulator introduces physics-based long-wave infrared (LWIR) cameras and configurable night-vision modes for degraded visual environments. To support multi-robot autonomy research, HERCULES provides lightweight APIs, ROS 2 wrappers, and rigorous time synchronization across multi-modal sensors and heterogeneous platforms. HERCULES further brings state-of-the-art game-engine capabilities into robotics simulation by integrating intelligent agents, such as pedestrians, traffic, and wildlife, together with high-fidelity dynamic environmental phenomena, including fire, flooding, and crop disease spread, offering a comprehensive, photorealistic testbed for autonomous robot deployment. HERCULES can be run in two modes to support a variety of multi-robot research: \emph{passively}, replaying offline-designed robot trajectories to generate reproducible multi-modal datasets, and \emph{actively}, running an online planner in closed loop from live observations. Our extensive experiments in heterogeneous multi-robot SLAM, collaborative perception, and exploration, using both HERCULES-generated data and active closed-loop execution, demonstrate its capability and utility as a versatile testbed for advancing heterogeneous multi-robot autonomy. We publicly release our simulation source code, experiment code, documentation, and datasets, including a heterogeneous multi-robot SLAM benchmark collected with two UAVs and two UGVs across kilometer-scale desert, forest, and city environments, at \texttt{https://lunarlab-gatech.github.io/HERCULES-website}.

\end{abstract}


\maketitle

\begin{figure}[ht]
    \centering

    \includegraphics[width=0.495\linewidth]{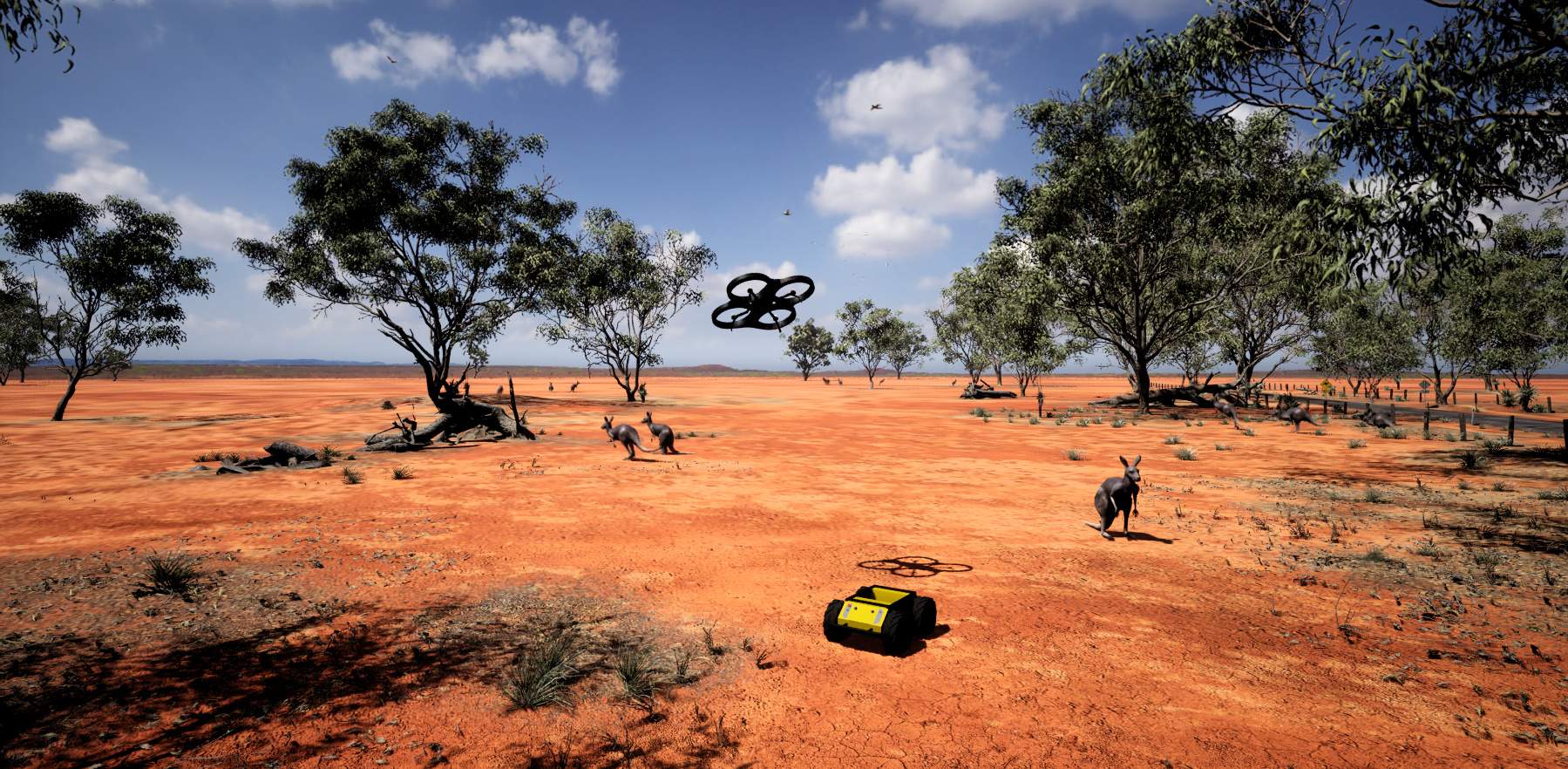}%
    \hspace{0.2em}%
    \includegraphics[width=0.495\linewidth]{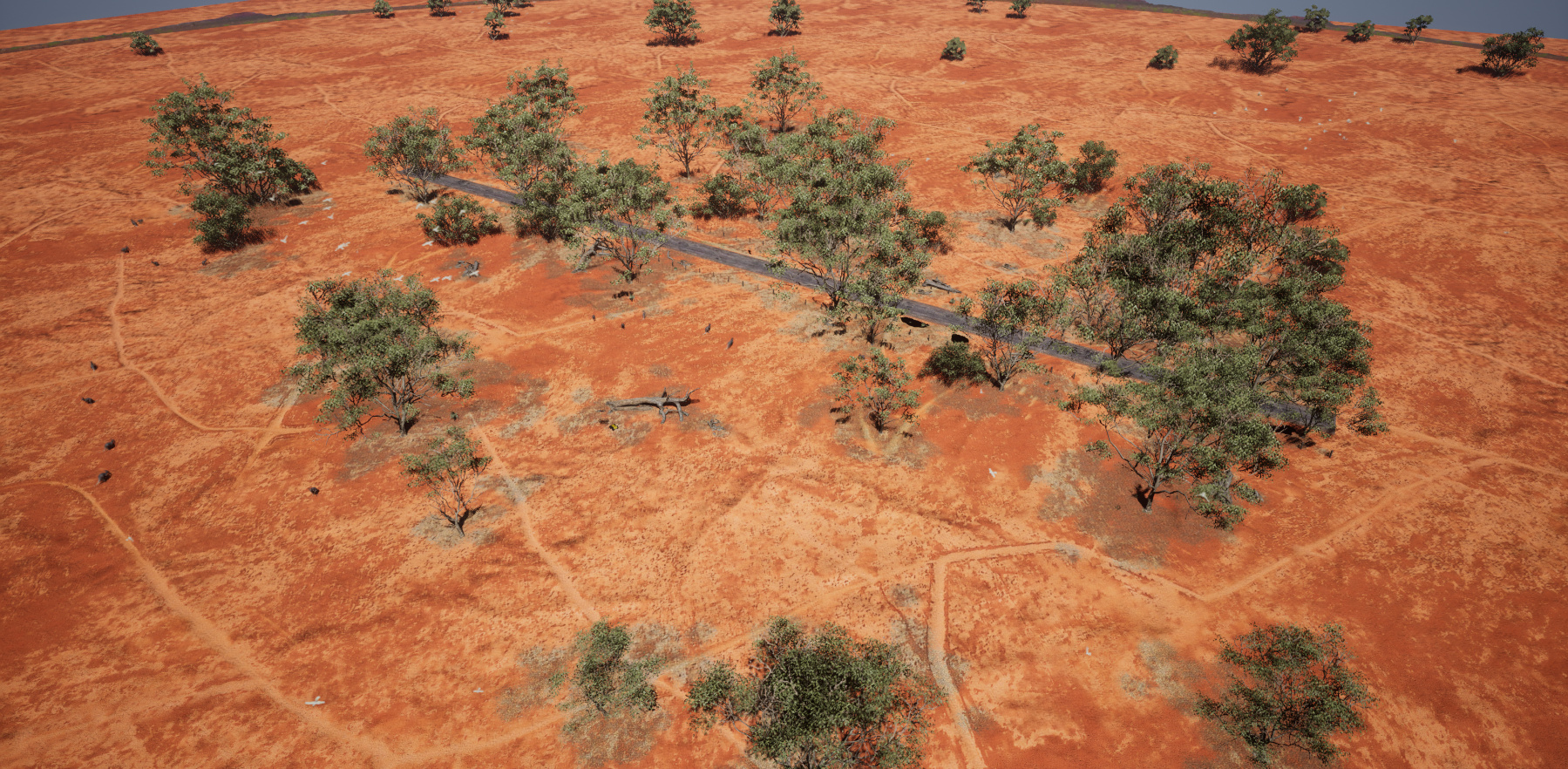}

    \vspace{0.4em}

    \includegraphics[width=0.495\linewidth]{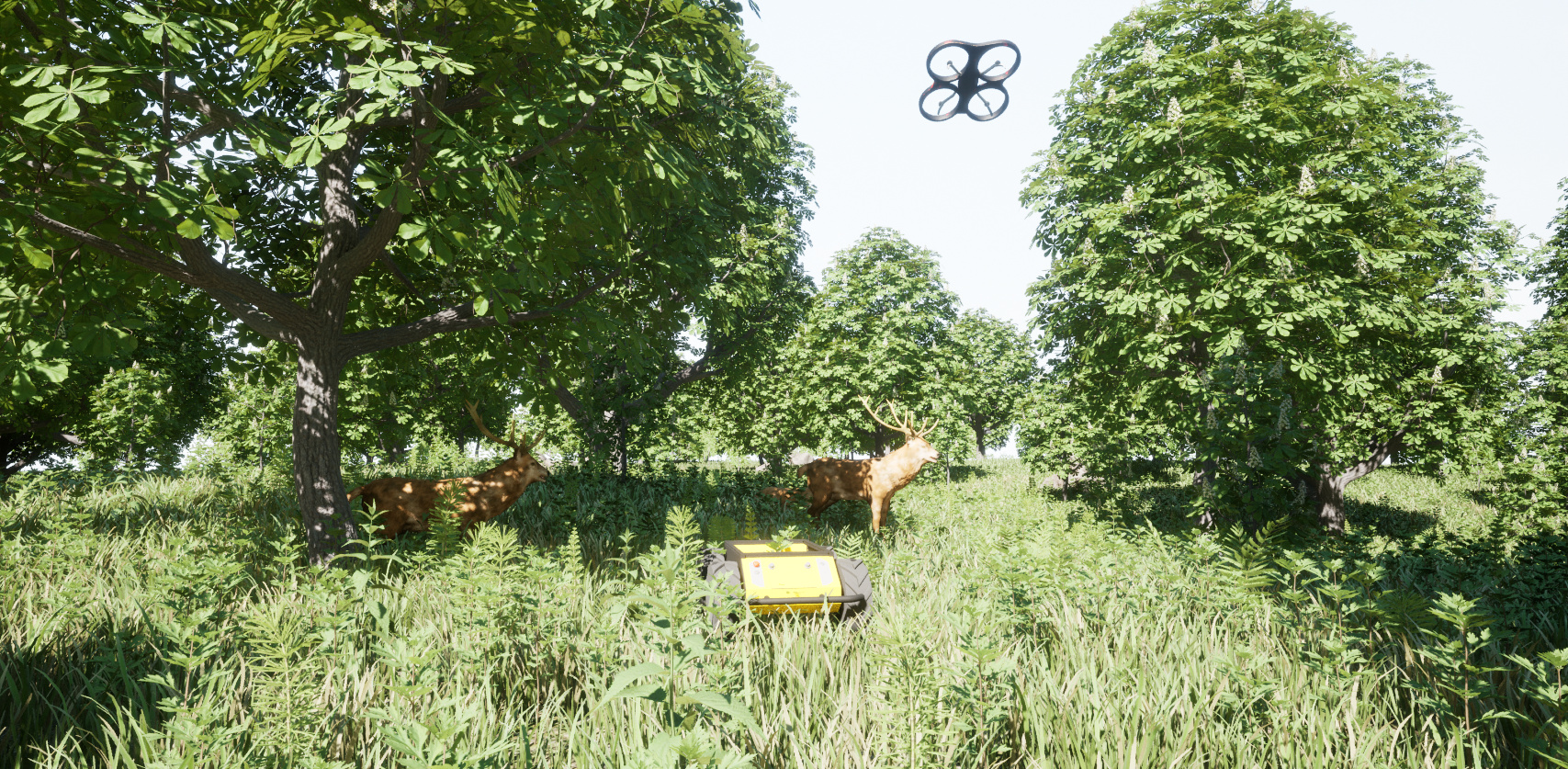}%
    \hspace{0.2em}%
    \includegraphics[width=0.495\linewidth]{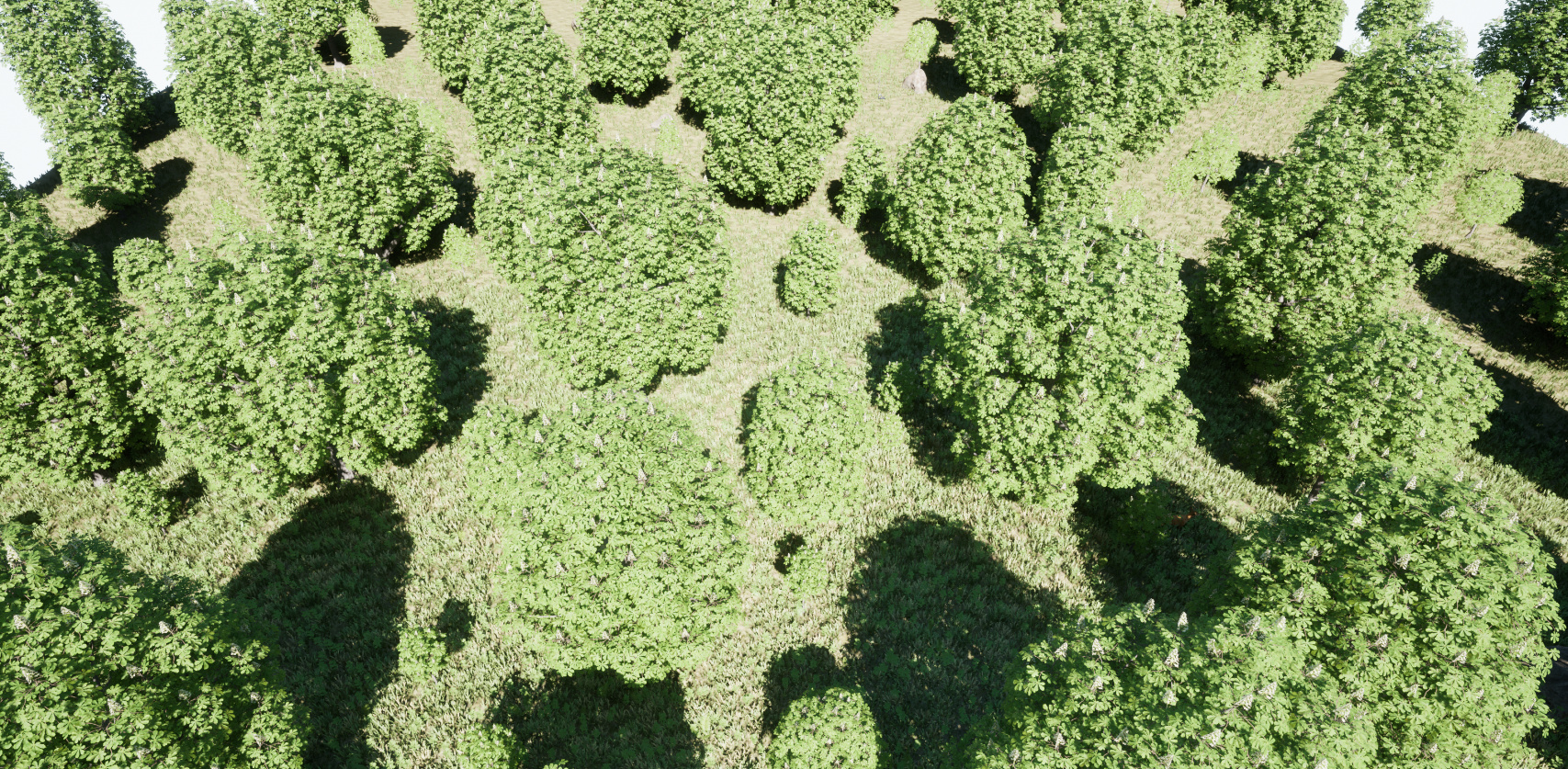}

    \vspace{0.4em}

    \includegraphics[width=0.495\linewidth]{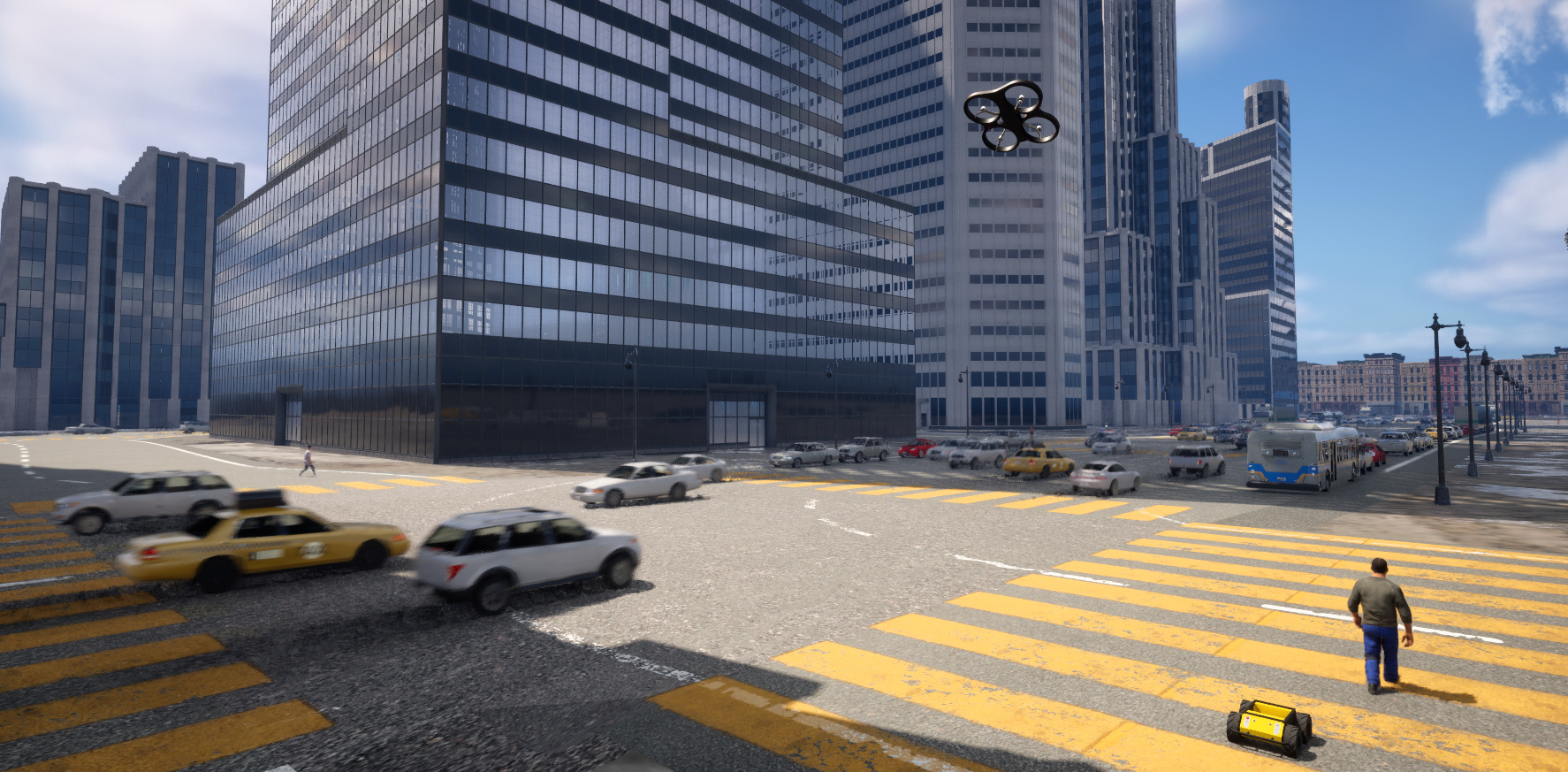}%
    \hspace{0.2em}%
    \includegraphics[width=0.495\linewidth]{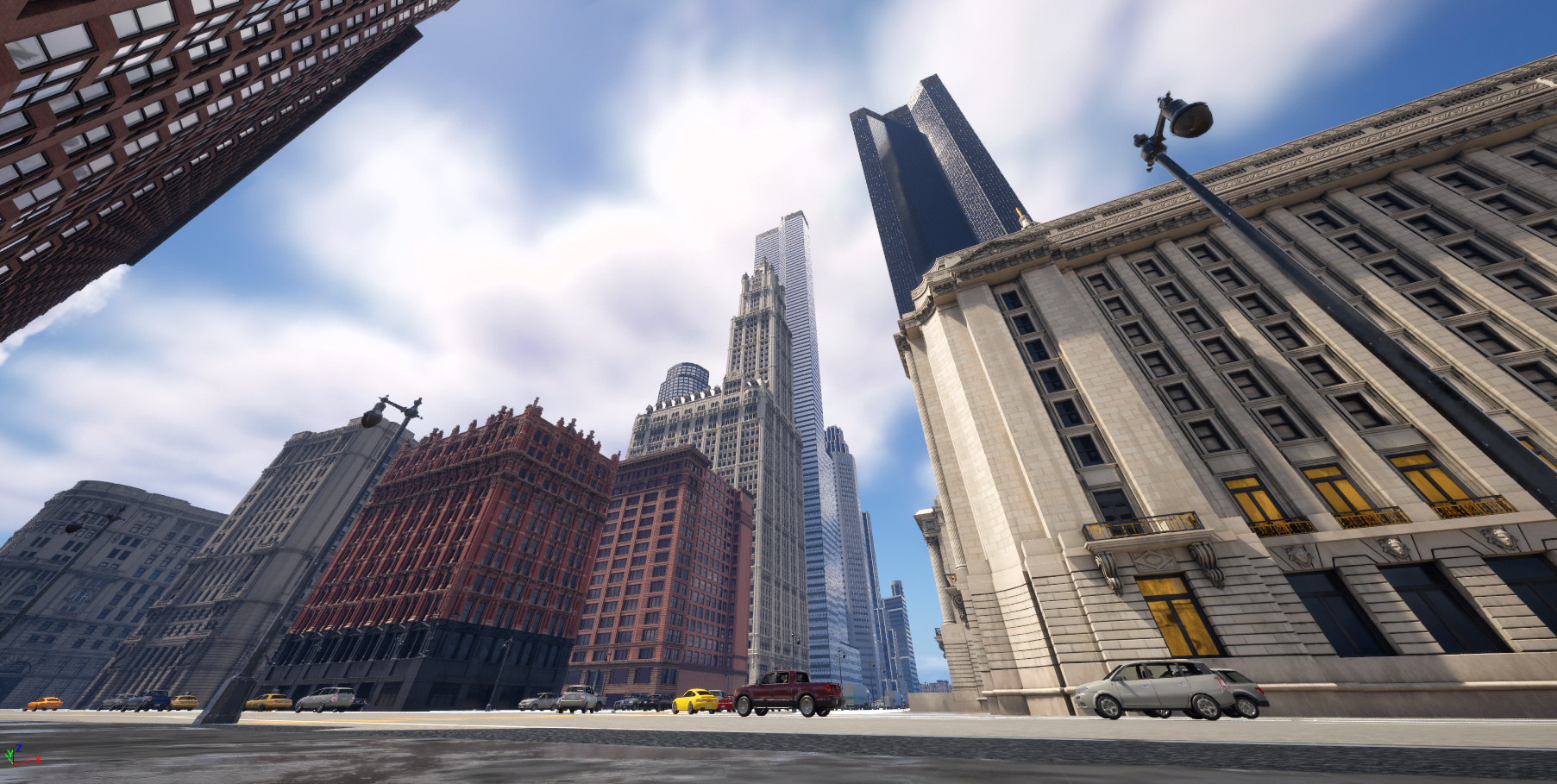}

    \caption{Environment diversity in HERCULES at two operational scales. (Left column) Ground-level detail. (Right column) High-level overview. (Top row) Desert. (Middle row) Forest. (Bottom row) City. Dynamic agents (AnimalAI wildlife, MetaHuman pedestrians, VehicleAI traffic) can be toggled based on experimental requirements.}
    \label{fig:fig1_env_diversity}
    \end{figure}

\section{Introduction}
Multi-robot systems are increasingly deployed in disaster response, environmental monitoring, and infrastructure inspection~\citep{queralta2020collaborative, tiwari2019multi, li2025review}. 
In these missions, heterogeneous teams of UAVs and UGVs are particularly compelling due to their complementary sensing capabilities, endurance, and mobility: UAVs provide rapid coverage and elevated viewpoints, while UGVs offer detailed close-range sensing and persistent ground access while being less constrained by size, weight, and power (SWaP)~\citep{munasinghe2024comprehensive}. 
However, rigorous progress on heterogeneous collaborative simultaneous localization and mapping (SLAM), scene understanding, and exploration remains difficult without simulation platforms that (i) support concurrent aerial-ground operations at kilometer scale, (ii) expose realistic, dynamic environments where coordination matters (e.g., occlusions, moving agents, and evolving hazards), and (iii) provide instrumentation to evaluate algorithms under bandwidth, line-of-sight, and GPS-denied conditions. 
Existing heterogeneous multi-robot datasets (e.g., CoPED~\citep{ZhouLoiannoCoPeD2024}, AirMuseum~\citep{DuboisFremont2020AirMuseum}, GRACO~\citep{zhu2023graco}), while valuable, are fixed recordings and therefore cannot support closed-loop coordination or controlled re-collection under new conditions; widely used simulators (e.g., Gazebo~\citep{Gazebo}, AirSim~\citep{ShahKapoor2018AirSim}, Isaac Sim~\citep{isaac_sim}), in turn, lack an out-of-the-box, research-ready stack for heterogeneous robot coordination and exploration in large-scale outdoor settings.

We present HERCULES (\emph{HE}terogeneous multi-\emph{R}obot simulator for \emph{Co}ordination, scene \emph{U}nderstanding, \emph{L}arge-scale \emph{E}xploration and \emph{S}LAM): an open-source \emph{simulator} and \emph{data-collection pipeline} built on Unreal Engine~5 (UE5)~\citep{epicgames_unrealengine5} by extending AirSim~\citep{ShahKapoor2018AirSim} and Cosys-AirSim~\citep{jansen2023Cosys}. The simulator supports concurrent heterogeneous UAV--UGV operation and can be driven in two modes: passively, replaying designed trajectories, or actively, with an online planner in closed loop. In either mode, the data-collection tool records the resulting runs as synchronized, standard-format datasets.
HERCULES extends these foundations by resolving a fundamental SimMode architecture conflict in AirSim that prevents concurrent UAV--UGV operation, enabling heterogeneous multi-robot autonomy in photorealistic, kilometer-scale environments (see Figure~\ref{fig:fig1_env_diversity}); a shared high-level waypoint-control interface across aerial and ground platforms; online and offline planning using OctoMap~\citep{hornung2013octomap} and elevation maps; and multi-modal sensing, including long-wave infrared (LWIR) and night-vision goggle (NVG) cameras, under diverse weather conditions and in the presence of dynamic agents such as vehicles, animals, and humans.
With HERCULES, UAV--UGV teams can fuse complementary observations, map unknown environments, and systematically test a range of coordination algorithms, from centralized to decentralized, before field deployment.

Additionally, HERCULES bridges the gap between what robotics researchers need and what state-of-the-art game engines offer. 
Rather than using UE5 assets only as passive scenery, HERCULES wraps MetaHuman pedestrians, VehicleAI traffic, AnimalAI wildlife, and environment processes (e.g., fire spread, urban flooding, crop-disease propagation) with parameterized, robotics-ready APIs that expose their configuration (e.g., spawn regions, paths, rates of spread) to the simulation pipeline. These interfaces lower the game-engine barrier for robotics researchers while preserving the visual realism and controllability needed for perception and planning experiments.

We validate the capabilities and practical utility of HERCULES through experiments in heterogeneous collaborative SLAM (ROMAN~\citep{peterson2025roman}) and cooperative 3D object detection (DAIR-V2X-style late fusion with PointPillars~\citep{lang2019pointpillars, yu2022dair}), together with a closed-loop multi-robot exploration demonstration in which a heterogeneous UAV--UGV team plans online. 
The collaborative SLAM benchmark spans all three environments, desert, forest, and city, which respectively stress sparse landmarks, perceptual aliasing, and dynamic agents. The collaborative SLAM and cooperative perception suites both rely on the synchronized multi-robot dataset collection and heterogeneous UAV--UGV operation uniquely enabled by HERCULES. To the best of our knowledge, HERCULES is the first open-source simulator that provides out-of-the-box support for heterogeneous multi-robot SLAM, collaborative perception, and closed-loop coordination within a single framework.

{
\renewcommand{\arraystretch}{1.05}
\setlength{\tabcolsep}{2pt}
\begin{table*}[ht]
\centering
\tiny
\caption{Platform facts, heterogeneity, environment realism, evaluation/dataset tools, and experiment design support. \cmark~=~native; \pmark~=~with effort/partial; \xmark~=~absent.}
\label{tab:simulator_comparison}
\begin{tabularx}{\textwidth}{|l|c|c|Y|Y|Y|Y|Y|Y|}
\hline
\multirow{2}{*}{\textbf{Simulator}} &
\multirow{2}{*}{\makecell{\textbf{Engine}\\(Renderer/Physics)}} &
\multirow{2}{*}{\textbf{Scope}} &
\multirow{2}{*}{\makecell{\textbf{Concurrent}\\\textbf{UAV+UGV}\\\textbf{operation}}} &
\multicolumn{3}{c|}{\makecell{\textbf{Env. realism}}} &
\multirow{2}{*}{\makecell{\textbf{Evaluation/}\\\textbf{Dataset}\\\textbf{collection tools}}} &
\multirow{2}{*}{\makecell{\textbf{Out-of-box}\\\textbf{experiment design}\\\textbf{and verification tools}}}
\\
\cline{5-7}
 & & & &
\makecell{\textbf{Outdoor}\\\textbf{km-scale envs}} &
\makecell{\textbf{Dynamic}\\\textbf{agents}} &
\makecell{\textbf{Dynamic}\\\textbf{phenomena}} &
 & \\
\textbf{HERCULES (ours)} & UE5/PhysX      & \textbf{Heterogeneous} & \cmark & \cmark & \cmark & \cmark & \cmark & \cmark \\
AirSim (UE4/5)           & UE/PhysX       & Aerial/Car             & \xmark & \pmark & \xmark & \xmark & \xmark & \xmark \\
Cosys-AirSim             & UE5/PhysX      & Aerial/Car/UGV         & \xmark & \pmark & \pmark & \pmark & \pmark & \pmark \\
Isaac Sim (Omniverse)    & Omniverse/Flex & Multi-domain           & \pmark & \pmark & \pmark & \pmark & \xmark & \pmark \\
CARLA (driving)          & UE/PhysX       & Ground (AV)            & \xmark & \cmark & \cmark & \pmark & \xmark & \cmark \\
FastSim (Unity; aerial)  & Unity/Flexible & Aerial                 & \xmark & \pmark & \xmark & \xmark & \pmark & \cmark \\
ARGoS (swarms)           & Custom         & Aerial/Ground          & \xmark & \xmark & \xmark & \xmark & \xmark & \pmark \\
Gazebo / Ignition        & ODE/DART+      & Multi-domain           & \pmark & \pmark & \pmark & \xmark & \xmark & \xmark \\
Webots                   & ODE            & Multi-domain           & \pmark & \xmark & \pmark & \xmark & \xmark & \cmark \\
\hline
\end{tabularx}
\end{table*}
}
To summarize, the main contributions of our HERCULES framework include:
\begin{itemize}
    \item \textbf{A heterogeneous-robot simulator.} We re-architect the AirSim/Cosys-AirSim SimMode layer to run UAVs and UGVs concurrently in a single session under a shared world state and unified simulation clock, by resolving a physics-engine conflict that previously restricted each session to one vehicle type.
    \item \textbf{A navigation stack.} We provide a unified navigation stack for mapping, planning, and control across heterogeneous platforms by introducing a waypoint-tracking UGV controller, a kinodynamic planner, and a ground-truth mapping (including OctoMap, elevation, and slope layers) and traversability analysis pipeline.
    \item \textbf{A synchronized data-collection tool.} We implement a global simulation clock with deterministic pause, step, and resume control that governs all sensor capture and pose logging, enabling time synchronization across multi-modal sensors and heterogeneous robots. We also provide tools for exporting data to KITTI-style and ROS~2 formats, with trajectory design patterns for controlled, repeatable data collection.
    \item \textbf{A collaborative SLAM benchmark.} We release a UAV--UGV collaborative SLAM benchmark dataset collected with two UAVs and two UGVs across desert, forest, and urban scenes, and evaluate it using multiple state-of-the-art SLAM methods.
    \item \textbf{An open release with three demonstrations.} We publicly release the simulator, datasets, and runnable experiment code, and validate the platform on three heterogeneous tasks: the collaborative SLAM benchmark above, a cooperative 3D detection study showing sim-to-real transfer to DAIR-V2X, and a closed-loop multi-robot exploration demonstration.
\end{itemize}

\section{Related Work}
We review representative open-source robotics simulators, highlighting their impact on robotics research as well as the gaps that motivate our work. Table~\ref{tab:simulator_comparison} provides an at-a-glance comparison of their key capabilities and summarizes the additional functionality offered by HERCULES.

\subsection{Classical Robotics Simulators}

Gazebo/Ignition~\citep{Gazebo} and Webots~\citep{michel2004cyberbotics} are long-standing robotics simulators widely used for control, navigation, and manipulation, with broad support for robot models, sensors, and ROS integration. 
ARGoS~\citep{pinciroli2012argos} targets large swarm experiments with efficient CPU-based physics, and numerous Gazebo-based forks and distributions adapt the core engine to specific domains such as aerial SITL. 
However, these tools are not optimized for perception-driven research in photorealistic, kilometer-scale worlds with rich dynamics: constructing such environments is cumbersome, and the resulting scenes typically lack the visual realism and variability demanded by modern vision and mapping pipelines. 
While multi-robot deployments are supported, there is no out-of-the-box stack for tightly coupled UAV-UGV coordination, shared autonomy across robot types, and evaluation under evolving, visually complex outdoor conditions.

Gazebo-based packages such as RotorS~\citep{rotorS} and Hector~\citep{kohlbrecher2014hector} specialize in aerial robotics, providing IMU, GPS, and camera models for navigation and control research, and are often used with PX4~\citep{meier2015px4} or ArduPilot ~\citep{ArduPilotSITL} SITL. 
These frameworks excel at single-drone flight dynamics but remain confined to aerial domains with simple, static scenes. 
They inherit Gazebo’s limitations in rendering realism and computational scalability, and offer no built-in support for heterogeneous teams, terrain-aware planning, or joint UAV-UGV experiments.

\subsection{Photorealistic High-Fidelity Simulators}

These photorealistic simulators achieve far higher visual fidelity than the classical robotics simulators above. 
AirSim, built on Unreal Engine 4, supports photorealistic physics-based simulation for drones and cars. 
Its UE5 port, Cosys-AirSim~\citep{jansen2023Cosys}, improves sensor types and their realism but largely retains AirSim’s single-vehicle structure and provides no coordinated multi-robot autonomy or tools for synchronized multi-robot dataset collection. 
Beyond Cosys-AirSim, several AirSim forks target specific domains such as wildfire monitoring and multi-robot image covering experiments~\citep{xue2023research, xu2025communication}, but they mostly extend environments, sensors, and networking, rather than providing a unified heterogeneous navigation stack. 
CARLA~\citep{dosovitskiy2017carla} provides rich urban driving scenarios but remains vehicle-centric. 
NVIDIA Isaac Sim~\citep{isaac_sim}, part of the Omniverse platform, delivers advanced multi-robot simulation and high-quality rendering tailored for reinforcement-learning workflows. Its strong coupling with the NVIDIA GPU ecosystem enables high performance and realism, albeit with reduced cross-platform flexibility. 
FastSim~\citep{cui2024fastsim}, based on Unity, offers modular low-level control for multi-drone research but is limited to homogeneous aerial swarms. 
Together, these platforms emphasize either realism or scale, but none offer a unified, open, and extensible stack for heterogeneous ground-aerial coordination in large outdoor worlds. 
In contrast, HERCULES combines the visual realism of UE5 with open, scalable multi-robot autonomy and integrated planning tools.

\subsection{Task-Focused Perception and Navigation Simulators}

Frameworks such as BenchBot/BEAR~\citep{hall2022benchbot}, Habitat~\citep{puig2023habitat}, and iGibson ~\citep{li2022igibson} focus on active perception in structured indoor domains, offering standardized benchmarks for semantic mapping and navigation. 
In aerial robotics, FlightGoggles~\citep{GuerraKaraman2019FlightGoggles} and Flightmare~\citep{flightmare} enable high-speed vision-based flight and learning by decoupling rendering and physics. 
These systems excel at perception benchmarking but remain limited to single-agent or homogeneous setups in static scenes. 
They cannot model dynamic, kilometer-scale, photorealistic outdoor environments or support the coordinated planning and control required for heterogeneous UAV--UGV research. HERCULES addresses these gaps by enabling realistic, large-scale simulation for multi-robot SLAM, exploration, and cooperative perception.

\begin{figure*}[ht]
    \centering
    \includegraphics[width=\textwidth,height=\textheight,keepaspectratio]{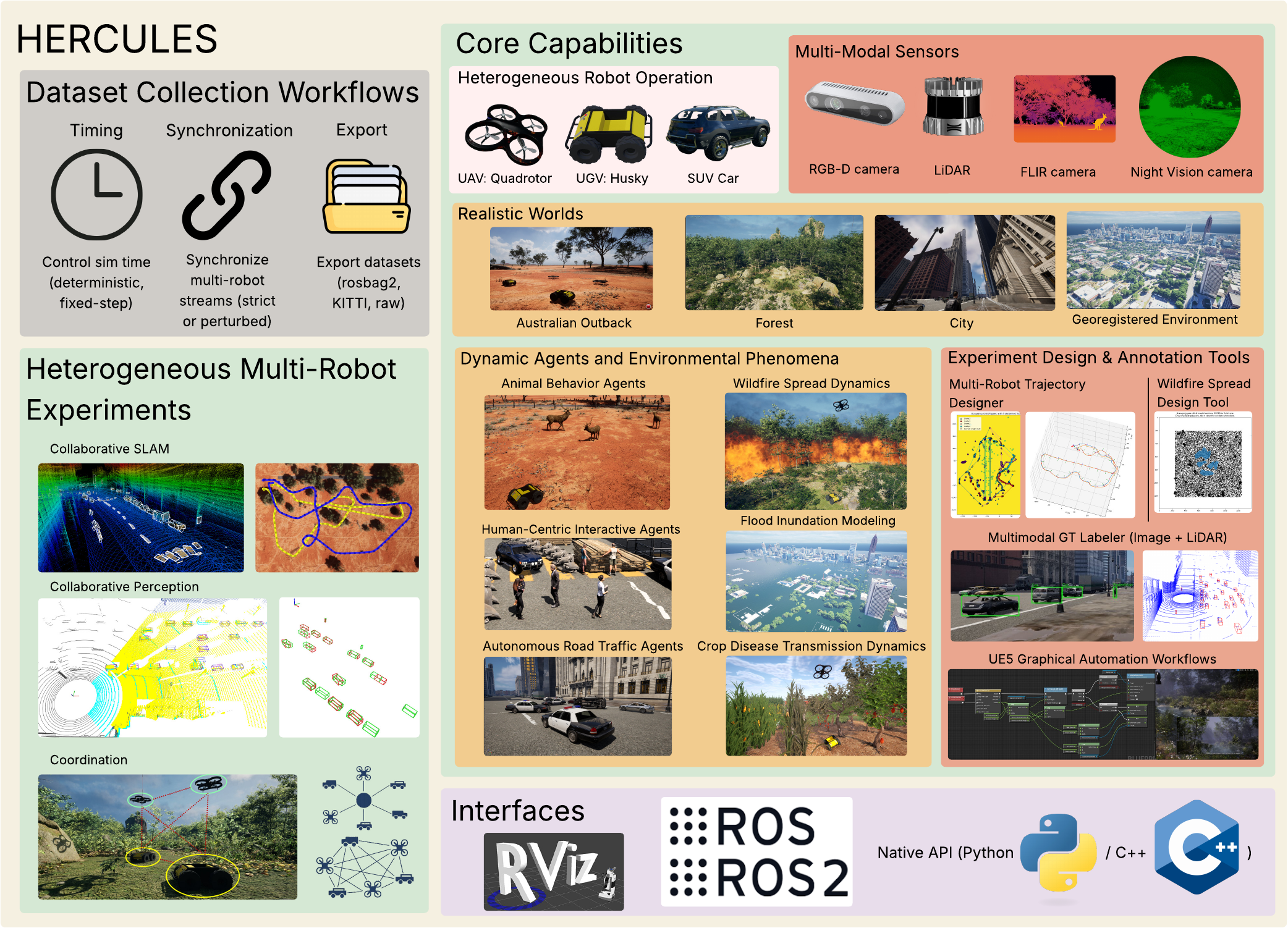}
    \caption{An overview of HERCULES, a UE5-based simulator and experimentation stack for heterogeneous UAV--UGV autonomy. HERCULES provides photorealistic large-scale worlds and synchronized sensing with ready-to-run interfaces, benchmarks, and dataset export for collaborative SLAM, cooperative perception, and exploration. The Heterogeneous Multi-Robot Workflows panel marks the capabilities quantitatively evaluated in this paper (Sec.~\ref{sec:experiments}); the Dynamic Agents and Environmental Phenomena modules are demonstrated as functional extensibility and are not included in the quantitative experiments.}
    \label{fig:system_overview}
\end{figure*}

\section{The HERCULES Simulator}
\label{sec:system_overview}
This section describes the functionality HERCULES provides; our experiments (Sec.~\ref{sec:experiments}) then evaluate representative capabilities. HERCULES is a simulator paired with a data-collection tool, and its functionality falls into three groups: (i)~a robot \emph{navigation stack}, including mapping, traversability analysis, planning, and control (Sec.~\ref{sec:autonomousnav}), used regardless of how the simulator is driven; (ii)~\emph{data-collection capabilities} for passive use, generating reproducible datasets (Sec.~\ref{sec:dataset_collection}); and (iii)~\emph{closed-loop operation} for active use, in which an online planner is stepped from live observations (Sec.~\ref{sec:closedloop}). Underlying all three is the concurrent heterogeneous simulation core (Sec.~\ref{sec:hetero_robot_op}). Figure~\ref{fig:system_overview} gives an overview of the simulator and its capabilities.

\subsection{Concurrent Heterogeneous Operation}
\label{sec:hetero_robot_op}
HERCULES currently supports multirotor UAVs (quadrotors), UGVs (Husky-style differential-drive), and SUV car platforms.
Enabling true heterogeneous multi-robot operation requires solving a fundamental architectural conflict in AirSim and Cosys-AirSim. These platforms use a \emph{SimMode} abstraction that binds each simulation session to a single vehicle type: multirotor mode uses a simplified fast-physics engine for rapid aerodynamic integration, while car mode uses PhysX for wheel-surface contact dynamics. Because SimMode is set globally at startup and configures the physics pipeline, spawn logic, and API dispatch for all vehicles, it is not possible to run UAVs and UGVs concurrently without deep modifications to the simulation core. We resolved this by re-architecting the SimMode layer to support concurrent heterogeneous vehicle types within a single simulation session, routing each platform to its appropriate physics backend while maintaining a shared world state and a unified simulation clock.

On top of this unified execution layer, we implement a waypoint-level command interface that abstracts platform-specific control so that high-level planners can issue commands to UAVs and UGVs through the same API. We additionally develop a new autonomous UGV controller (Sec.~\ref{sec:ugv_controller}) that mirrors the existing UAV waypoint-tracking interface, enabling symmetric treatment of aerial and ground platforms. We also enabled multi-robot sensor logging with globally synchronized timestamps for dataset generation (Sec.~\ref{sec:dataset_collection}). Realizing concurrent heterogeneous operation also requires fixing low-level sensor behavior that had assumed a single vehicle type; for example, ensuring that each robot's LiDAR completes a full revolution on the shared, synchronized physics tick rather than being interrupted when multiple heterogeneous vehicles are stepped together.
Physical and dynamic properties (e.g., mass, drag, wheel friction, thrust coefficients) remain configurable through UE5 Blueprints and AirSim configuration files, as in Cosys-AirSim.

\begin{figure}[ht]
    \centering
    \includegraphics[width=\linewidth]{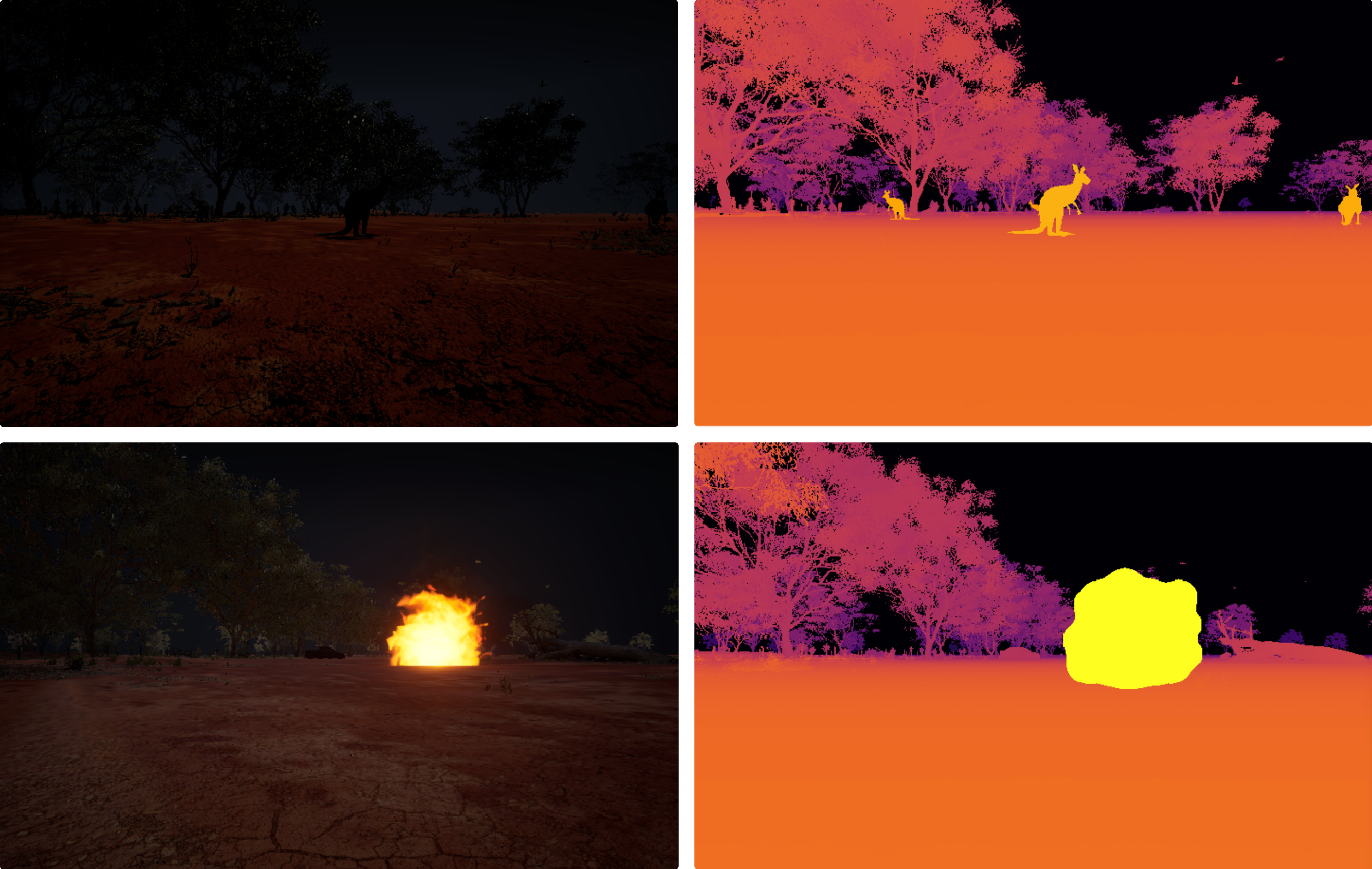}
    \caption{
    Long-wave infrared (LWIR) rendering in the desert environment.
    (Top left) RGB image under low illumination showing limited visibility.
    (Top right) Corresponding LWIR output where warm-bodied animals (kangaroos) appear as high-temperature regions with clear contrast against the background.
    (Bottom left) RGB image containing an active fire source.
    (Bottom right) LWIR response to the same configuration, exhibiting high-temperature saturation and a sharply defined thermal plume while preserving detail and contrast in cooler surroundings.
    }
    \label{fig:flir_kangaroo_fire}
\end{figure}

\subsection{Interfaces}
\label{sec:interfaces}
HERCULES provides two complementary interfaces: a ROS~2 interface for integration with existing autonomy stacks, and a lightweight Python/C++ API for prototyping and machine learning.
The ROS~2 interface publishes all simulated sensor and ground-truth streams and accepts commands through standard ROS~2 message types, so it integrates directly with mainstream robotics autonomy stacks.
Its multi-threaded publishers stream the sensor and ground-truth topics concurrently, which is necessary to keep pace with high-rate data from multiple robots in real time.
Standard ROS~2 tooling such as RViz2 can therefore be used directly for visualization and debugging.
The Python/C++ API, inherited from Cosys-AirSim, exposes all sensor streams, ground-truth poses, environment metadata, and robot control commands, supporting fast prototyping and machine-learning workflows.
Both interfaces share the same simulation clock and support deterministic pause, step, and resume control. Runs are therefore exactly repeatable, and the two interfaces can be used concurrently (for example, RViz2 visualization alongside Python control).

\subsection{Sensing and Environments}
\label{sec:sensing_env}
\subsubsection{Multi-Modal Sensors.}
\label{sec:multimodal_sensors}
HERCULES inherits existing Cosys-AirSim sensor suite, including RGB/depth/stereo cameras, LiDAR, IMU, GPS, barometer, magnetometer, pulse-echo, and UWB. On top of this, we add two derived imaging modalities for adverse sensing conditions, i.e., a long-wave infrared (LWIR) view and a night-vision goggle (NVG) view, computed from the simulator's existing RGB, depth, and ground-truth segmentation outputs. 
When captured through the data-collection tool (Sec.~\ref{sec:dataset_collection}), all streams share a common time base, and per-sensor delays can be injected to emulate latency. Ground-truth instance segmentation also provides unique per-object labels across multi-robot platforms for consistent data association.

\begin{figure}[ht!]
    \centering
    \includegraphics[width=0.97\linewidth]{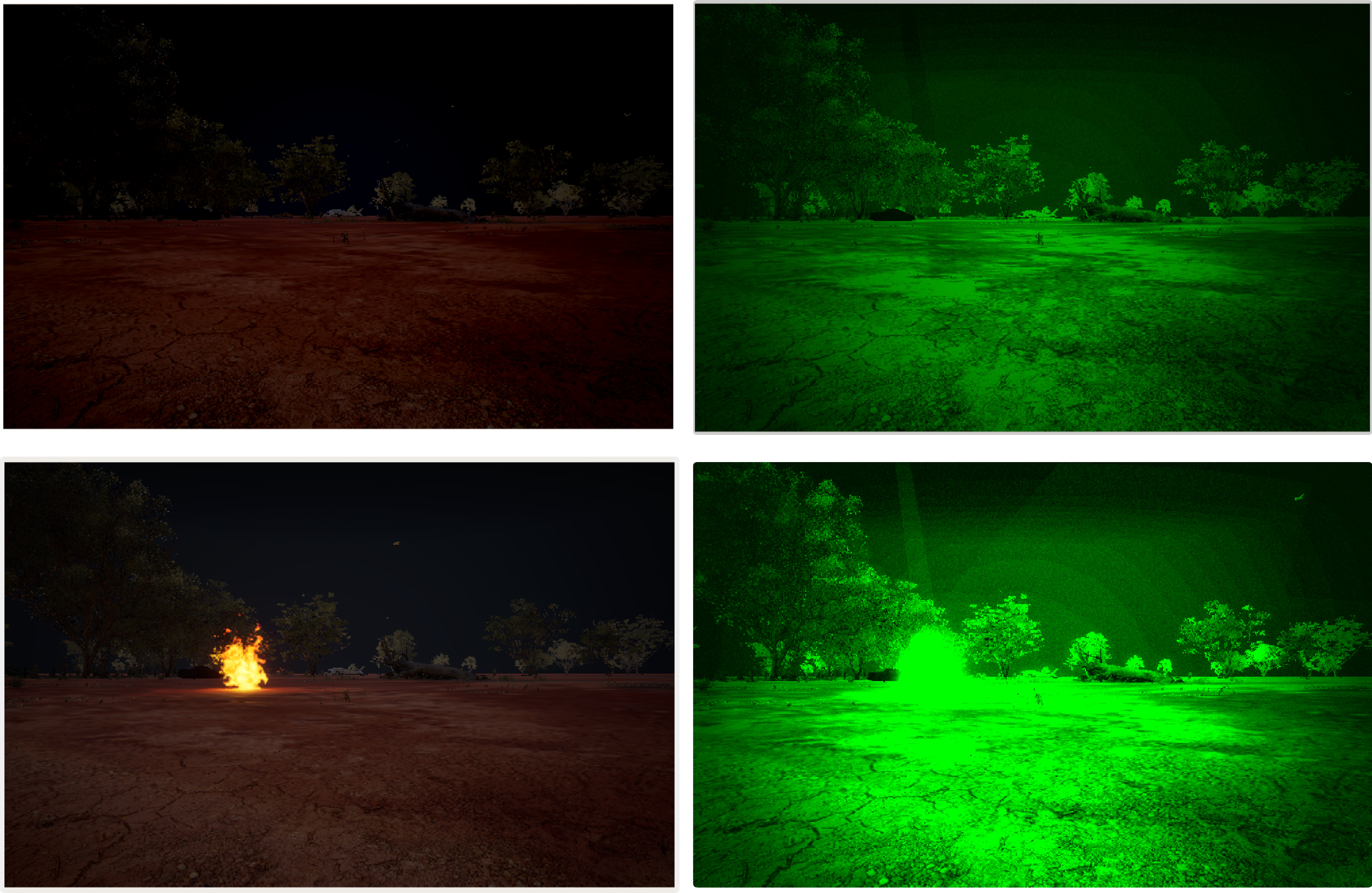} 
    \caption{
    Night-vision goggle (NVG) rendering in the desert environment.
    (Top left) RGB image under near-zero ambient illumination where the scene is essentially dark.
    (Top right) Corresponding NVG image revealing terrain and vegetation structure.
    (Bottom left) RGB image with an active fire source.
    (Bottom right) NVG response to the same configuration, showing realistic saturation, blooming, and brightness clipping around the fire while still preserving detail in the surrounding low-light regions.
    }
    \label{fig:nvg_fire}
\end{figure}

The LWIR camera synthesizes thermal imagery by integrating a Planck-law spectral radiance model over the 8--14\,$\mu$m band using material emissivity and temperature profiles~\citep{modest2021radiative}.
Ground-truth instance segmentation labels are paired with per-class material and nominal-temperature priors so that distinct object categories (e.g., animals, vehicles, terrain) exhibit physically interpretable thermal contrast (see Figure~\ref{fig:flir_kangaroo_fire}). Optional pseudocolor rendering (e.g., Inferno) enhances visual interpretability while preserving the underlying radiometric values. For low-light conditions, a night-vision goggle (NVG) mode approximates image-intensifier behavior using an empirical photometric transfer function that integrates adaptive gain, gamma correction, sensor noise, and a phosphor-style green colormap, following established approaches to night-vision-goggle appearance simulation~\citep{kooi2005crucial} (see Figure~\ref{fig:nvg_fire}). This model is tuned for realistic appearance in our environments rather than an exact hardware replication.
Both LWIR and NVG views are generated by lightweight processing modules layered on the existing camera outputs, so they can be enabled or re-parameterized at runtime without modifying the underlying rendering pipeline.

\subsubsection{Realistic Worlds.} 
\label{sec:realistic_worlds}
Built on UE5, HERCULES can import any compatible environments available in the UE5 Fab marketplace~\citep{fab2026}. Figure~\ref{fig:fig1_env_diversity} shows three representative large-scale photorealistic 3D environments we use as testbeds: desert, forest, and city, with the desert modeled after the Australian Outback.
Each environment presents distinct perception challenges: sparse landmarks and long-range visibility (desert), perceptual aliasing from repetitive geometry and similar semantics (forest), and structured geometry with strong occlusions and dynamic obstacles (city).
All scenes use physically based materials, realistic lighting, and configurable time-of-day via UE5's sky and atmosphere system, and HERCULES uses UE5's Lumen global illumination~\citep{epicgames_lumen} and Nanite geometry~\citep{epicgames_nanite} for real-time photorealistic rendering.
HERCULES is implemented as a UE5 plugin that can be dropped into any Unreal project, allowing users to import or author new scenes with minimal effort.

HERCULES also supports geo-registered environments for sim-to-real transfer. Geo-registration itself is provided by the Cesium for Unreal plugin~\citep{cesium_for_unreal}, which streams real-world terrain, satellite imagery, and building models as tiled datasets aligned to true latitude-longitude coordinates. Our contribution here is practical: we resolved a build issue that prevented the Cesium plugin from working in our Linux-based UE5 setup, and we provide documented, step-by-step setup scripts so that users can easily import large-scale geo-registered environments and obtain correctly geo-referenced sensor and ground-truth data reproducibly.

\subsubsection{Dynamic Agents and Environmental Phenomena.} 
\label{sec:dynamic_agents_phenomena}
\begin{figure}[t!]
    \centering
    \includegraphics[width=\linewidth]{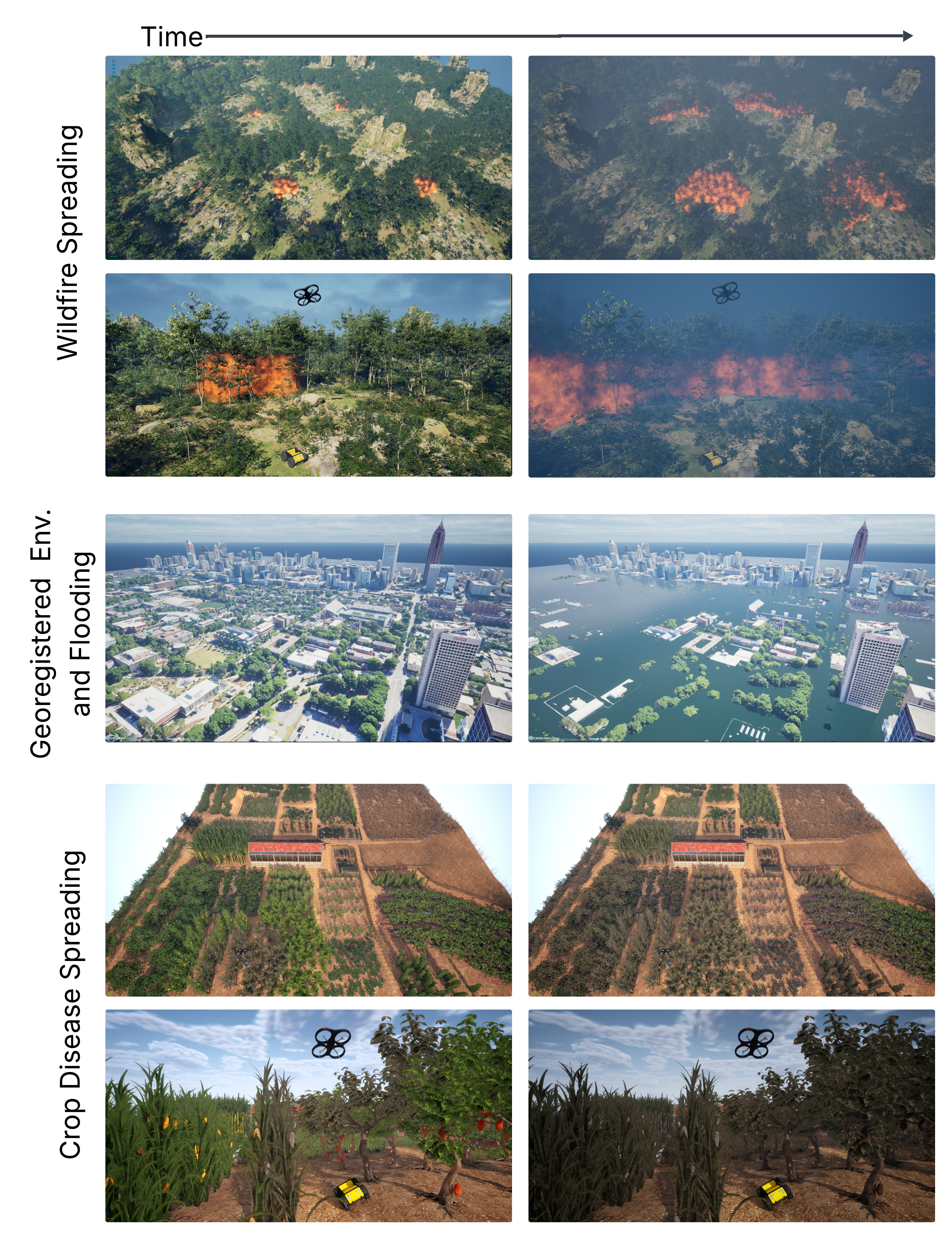}

    \caption{Dynamic environmental phenomena implemented in HERCULES. (Top) Wildfire spread with progressive smoke propagation. (Middle) Flood inundation on a geo-registered model of the Georgia Tech campus. (Bottom) Crop disease transmission across agricultural terrain.}
    \label{fig:fivexthree}
\end{figure}

Within these worlds, HERCULES simulates dynamic agents and evolving natural phenomena to enable realistic multi-robot interaction studies such as disaster response and environmental monitoring. On the dynamic agents side, we implement custom UE5 Blueprints for three categories of interactive agents: Animal Behavior Agents (e.g., kangaroos, deer), Human-Centric Interactive Agents (MetaHuman pedestrians), and Autonomous Road Traffic Agents (VehicleAI traffic), each with configurable paths and Unreal Engine AI logic~\citep{epicgames_ue5_ai}, rather than relying solely on off-the-shelf marketplace assets. On the environmental phenomena side, three classes of dynamic processes are implemented as independent, parameterized modules that update the UE5 world state at runtime: Wildfire Spread Dynamics, Flood Inundation Modeling, and Crop Disease Transmission Dynamics (see Figure~\ref{fig:fivexthree} for examples).
Each process can be seeded pseudo-randomly for reproducibility and configured by the user (e.g., rate of spread or affected area).
Robots perceive these changes through their sensors and can respond via the planning stack, supporting research in adaptive planning and situational awareness.
Multiple phenomena can coexist within a single simulation, as all modules share the same underlying UE5/Cosys-AirSim world state.

\section{Navigation Stack}
\label{sec:autonomousnav}
The navigation stack provides mapping, traversability analysis, planning, and control capabilities that underpin all HERCULES use cases, whether the simulator is operated passively for dataset collection (Sec.~\ref{sec:data_capabilities}) or actively in closed loop (Sec.~\ref{sec:closedloop}). AirSim and Cosys-AirSim provide a low-level waypoint-tracking controller for UAVs, but offer no UGV controller, no terrain-aware map representation, and no trajectory planning that accounts for ground-surface hazards such as slopes, overhangs, or negative obstacles. We add these components to support heterogeneous robot navigation.

The navigation stack consists of three components: (i) an environment map generation pipeline that converts UE5 geometry into OctoMaps and elevation maps for terrain-aware planning (Sec.~\ref{sec:envmapgen}); (ii) a kinodynamic trajectory planner that produces dynamically feasible paths for both UAVs and UGVs while respecting obstacle clearance and terrain constraints (Sec.~\ref{sec:kino-traj-planning}); and (iii) a pure-pursuit UGV controller that tracks planned paths using proportional steering and speed control, mirroring the existing UAV waypoint interface (Sec.~\ref{sec:ugv_controller}).

In the experiments reported here, we use this stack to plan trajectories offline on precomputed ground-truth maps and replay them deterministically for dataset collection; the same stack also runs online for closed-loop operation (Sec.~\ref{sec:closedloop}).

\subsection{Environment Map Generation and Processing}
\label{sec:envmapgen}
To support reliable motion planning in large-scale environments and provide ground-truth 3D structure for occupancy prediction and mapping research, we generate ground-truth maps of simulated environments through an offline preprocessing pipeline. 
For the highly detailed, kilometer-scale environments used in our experiments, we subdivide the environment model into manageable $100\mathrm{m}\times100\mathrm{m}$ tiles to ensure memory efficiency and scalability; this tiling keeps peak memory bounded so the full pipeline can run on low-RAM machines without exhausting memory.
Each tile's UE5 3D geometry (e.g., mesh or point cloud) is converted into an occupancy voxel grid via \texttt{binvox}~\citep{min_binvox}. 
Each voxelized tile is then converted into an octree representation using the OctoMap conversion utilities (\texttt{binvox2bt\_unique\_offsets}), producing a \texttt{.bt} octree where each leaf node stores log-odds of occupancy. 
During conversion, spurious voxels and small isolated clusters below a threshold are filtered, with optional dilation and smoothing applied to improve map consistency.

\begin{figure}[t!]
    \centering
    \includegraphics[width=\linewidth]{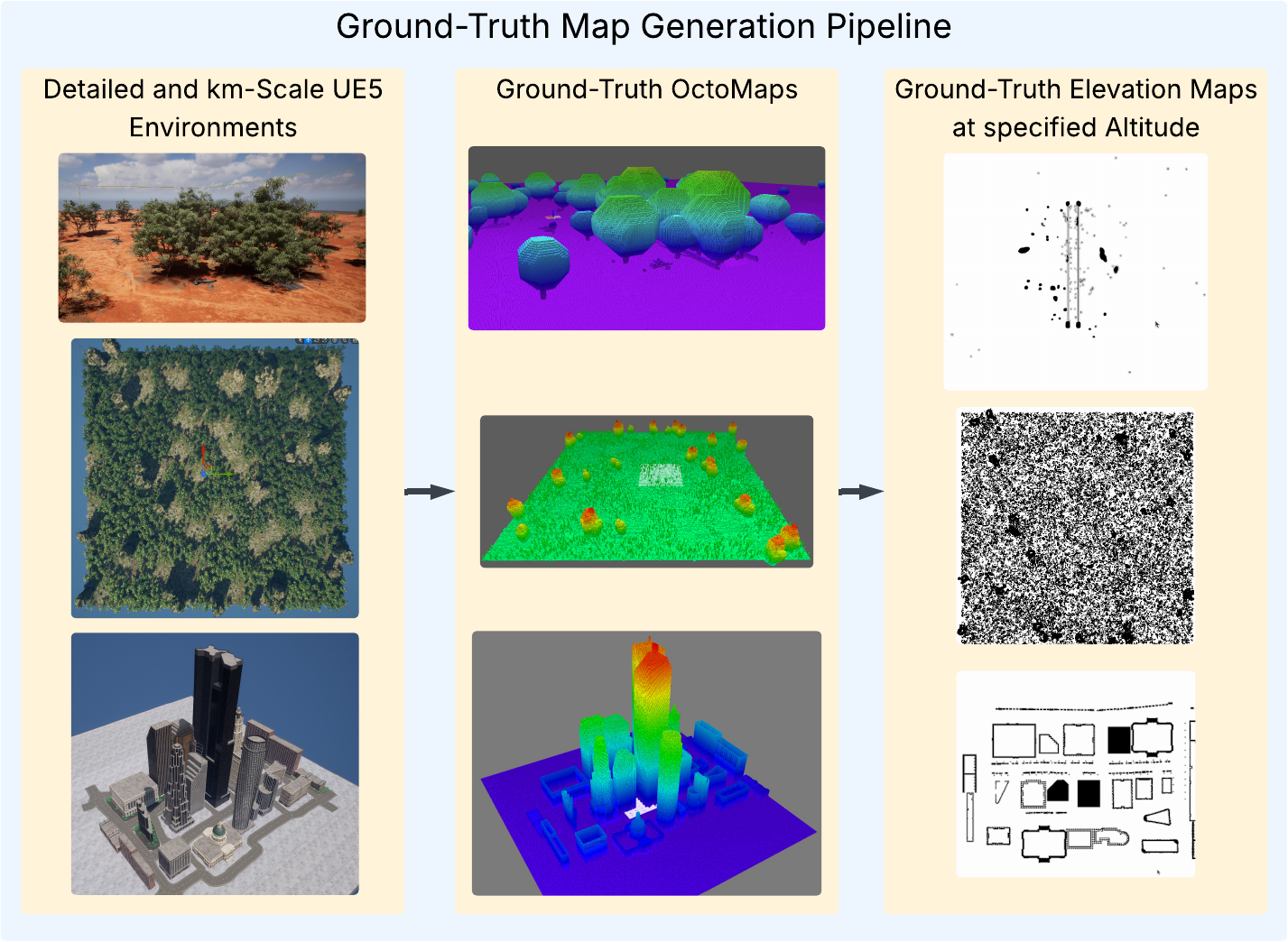}
    \caption{Ground-truth map generation pipeline. Detailed UE5 environments (left) are converted into ground-truth OctoMaps (center), followed by elevation maps at user-specified altitudes (right). This unified mapping pipeline can be run offline for full-environment preprocessing or online in a local robot-centric region for real-time planning.}
    \label{fig:gt_mapping_pipeline}
\end{figure}

On top of this, we derive a 2.5D elevation map by retaining the lowest traversable surface within each grid cell. 
In practice, for each $(x,y)$ cell in the horizontal plane, we identify the ground height, i.e., the lowest occupied voxel column that can support the UGV, and record that as the cell's elevation. 
Any occupied voxel above that ground height, such as walls, rocks, or overhangs, is marked separately so that the cell can be considered blocked for the UGV if the obstacle is tall. 
We then compute a slope for each cell by comparing the elevation with neighboring cells. The slope map $\nabla h(x,y)$ allows the system to classify terrain: regions where the incline exceeds a threshold, e.g., a slope corresponding to the UGV's tipping or wheel slip limit, are labeled non-traversable. 
This elevation-based occupancy grid, effectively a digital terrain model with an obstacle overlay, is used in conjunction with the full 3D OctoMap to plan safe ground paths for UGVs and inform UAVs of terrain relief.
Used together, the OctoMap and elevation map enable terrain-aware global planning that accounts for volumetric obstacles that are crucial for UAV flight, tall obstacles that may occlude the UAV's view or impede the UGV, and ground-surface hazards that are critical for UGV mobility.
Figure~\ref{fig:gt_mapping_pipeline} illustrates the generated map representations, including an OctoMap slice and the corresponding elevation map. 
Ground-truth maps can be used for trajectory generation during data collection, as perception inputs for closed-loop operation, and as ground truth labels for occupancy mapping evaluation.

\subsection{Kinodynamic Trajectory Planning}
\label{sec:kino-traj-planning}
HERCULES includes a randomized kinodynamic RRT (KRRT) planner, adapted from~\citep{lavalle2001randomized}, that produces time-parameterized, dynamically feasible trajectories for both UAVs and UGVs. The planner supports task-aware sampling by optionally biasing samples toward goals, frontiers, or previously visited landmarks, as well as clearance-aware expansion by penalizing proximity to obstacles using a signed distance field derived from the OctoMap. For SLAM-oriented dataset collection, users can mark checkpoints in previously mapped regions, and a revisit distribution then steers robots back through those checkpoints to induce intra- and inter-robot loop closures. The full mathematical formulation, including dynamics models, sampling distributions, and steering objectives, is provided in Appendix~\ref{app:krrt-details}. 

\begin{figure*}[t]
    \centering
    \vspace{0.3em}

    \includegraphics[width=0.92\textwidth]{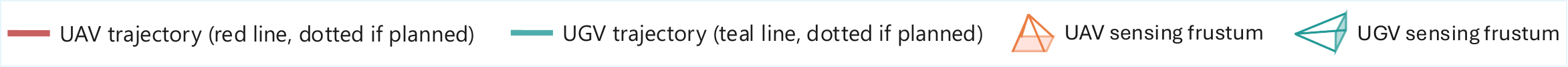}
    \vspace{0.35em}

    \begin{subfigure}[t]{0.48\textwidth}
        \centering
        \includegraphics[width=\linewidth]{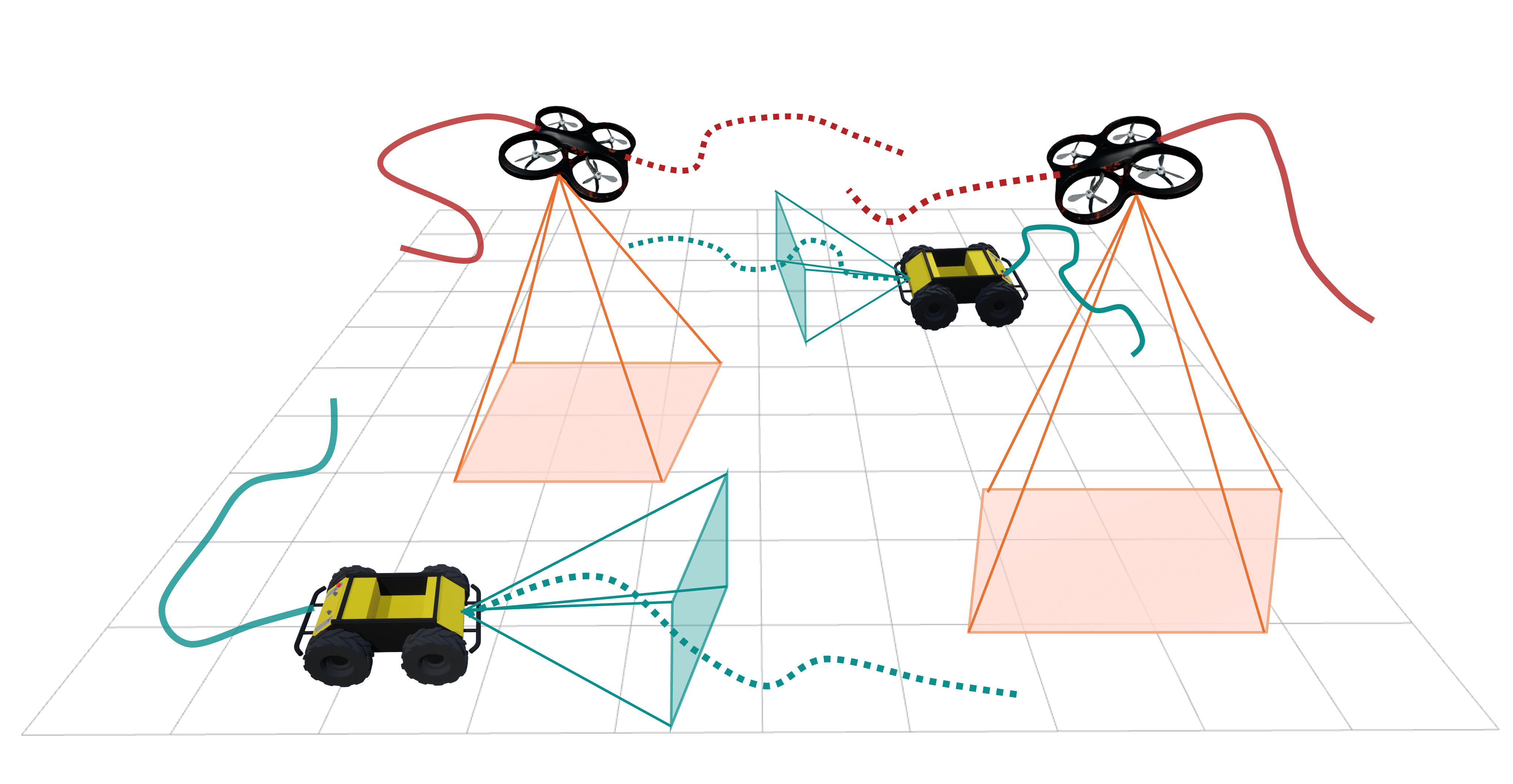}
        \caption{Complementary Coverage Mode. UAVs and UGVs disperse to maximize complementary coverage and information gain while reducing sensing redundancy.}
        \label{fig:coop_explore}
    \end{subfigure}
    \hfill
    \begin{subfigure}[t]{0.48\textwidth}
        \centering
        \includegraphics[width=\linewidth]{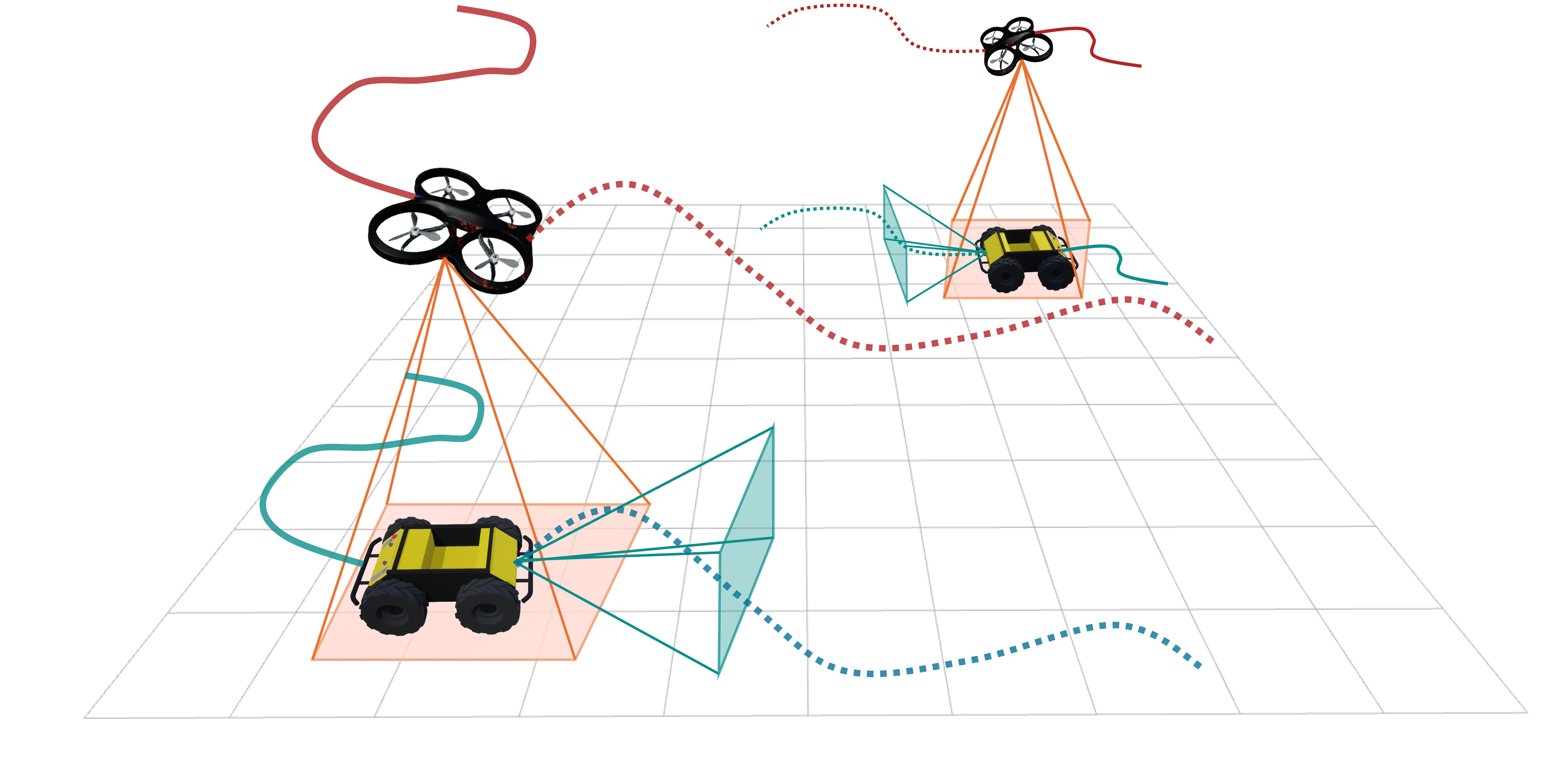}
        \caption{Leader-Follower Mode. UAVs maintain controlled overlap with UGV sensing footprints along a shared route to support joint perception.}
        \label{fig:convoy_mode}
    \end{subfigure}

    \caption{Operational modes for heterogeneous UAV--UGV teams in HERCULES. (a) Complementary Coverage disperses agents to improve coverage and reduce uncertainty. (b) Leader-Follower enforces spatiotemporal overlap for cooperative perception. Legend applies to both panels.}
    \label{fig:coop_vs_convoy}
\end{figure*}

\subsection{UGV Low-Level Controller}
\label{sec:ugv_controller}
We implement a geometric pure-pursuit path tracker for the UGV with proportional steering and speed control. The controller selects a forward lookahead target on the reference path and computes heading error and speed commands with saturation limits, consistent with the unicycle dynamics used in planning. The same controller is used for both the Husky-style UGV and the simulated SUV, with only geometric parameters (i.e., wheelbase, steering limits) adjusted. The full formulation is provided in Appendix~\ref{app:ugv-controller-details}.

\section{Data-Collection Workflows}
\label{sec:data_capabilities}
Used \emph{passively}, HERCULES is a synthetic data generator: users design multi-robot trajectories, execute the heterogeneous team, and export synchronized multi-modal datasets for offline SLAM and perception research. This section describes the synchronized logging and export machinery (Sec.~\ref{sec:dataset_collection}), the trajectory design patterns that enable meaningful, controllable, and repeatable data collection (Sec.~\ref{sec:motion_patterns}), and the experiment-design tooling built on top.

\subsection{Synchronized Logging and Export}
\label{sec:dataset_collection}
A valuable capability for collaborative SLAM and cooperative perception research is access to time-synchronized data streams across multi-modal sensors on each robot and, when needed, across multiple robots. Although collaborative SLAM systems can often operate asynchronously, precise cross-robot synchronization is important for controlled benchmarking, tightly coupled multi-robot fusion, and cooperative perception in dynamic scenes. AirSim and Cosys-AirSim trigger each sensor independently and do not guarantee synchronized data streams across heterogeneous sensors and platforms.
HERCULES provides a data-collection tool that drives the simulator through deterministic pause-step-resume cycles and gates all sensor capture and pose logging on a single global clock, ensuring that every robot's data streams share a common time base. Achieving this requires fixing sensor behavior under stepped execution, most notably ensuring that each LiDAR scan completes a full revolution across robots under pause/resume rather than being cut off. This enables deterministic and repeatable multi-robot dataset collection with consistent inter-sensor timing.

\subsubsection{Synchronization modes.}
Timing can be configured either in a strictly synchronized mode, where all sensors are triggered at exact global-clock ticks, or with controlled perturbations to emulate realistic clock drift and communication delay for robustness testing.

\subsubsection{Export.}
Exporters write multi-modal sensor data together with metadata YAML files.
Any experiment can be recorded as a ROS or ROS~2 bag with synchronized `/clock', or as a raw file set.
HERCULES supports generic PNG, TXT, NPY, and CSV exports, as well as converters to KITTI-style dataset formats commonly used by SLAM and 3D object detection pipelines.
Datasets for the collaborative SLAM benchmark (Sec.~\ref{sec:experiments-collab-slam}) and cooperative-perception experiments (Sec.~\ref{sec:experiments-collab-perception}) are generated directly through this dataset layer, providing a repeatable path from simulation runs to standardized evaluation.

\subsection{Trajectory Design Patterns}
\label{sec:motion_patterns}

To collect meaningful datasets for multi-robot research, we design two complementary trajectory-generation modes that represent different collaborative robot applications. For environmental exploration and monitoring, robots tend to spread out to maximize information gain, whereas shared fields of view are critical for collaborative object detection, tracking, and scene understanding.
This subsection describes the two motion patterns we use to design trajectories for our UAV--UGV teams. These trajectories are executed to generate the datasets used in Sec.~\ref{sec:experiments-collab-slam} and Sec.~\ref{sec:experiments-collab-perception}. As shown in~Figure~\ref{fig:coop_vs_convoy}, we define two UAV--UGV trajectory patterns:

\subsubsection{Complementary Coverage Mode.} 
The UGV and UAV follow trajectories with minimal field-of-view overlap, spreading out to observe different parts of the scene. This pattern maximizes spatial coverage and viewpoint diversity, and is used in our collaborative SLAM experiments (Sec.~\ref{sec:experiments-collab-slam}) to ensure that inter-robot loop closures arise from shared landmarks rather than direct co-observation. Imposing shared landmarks is achieved by specifying revisited checkpoints, as described in Sec.~\ref{sec:kino-traj-planning}.

\subsubsection{Leader-Follower Mode.} 
The UGV follows a ground-level route while the UAV maintains overhead coverage along the same path, ensuring \emph{spatiotemporal} overlap between the two platforms' sensor fields of view. We use this pattern in our cooperative perception experiments (Sec.~\ref{sec:experiments-collab-perception}) to generate paired UAV--UGV frames for 3D object detection.

Trajectories for both patterns are designed offline and replayed deterministically to collect synchronized multi-modal datasets. Decoupling trajectory design from data collection makes runs exactly repeatable across experiments.

\subsection{Experiment Design Tools}
\label{sec:exp_design_tools}
HERCULES provides tools to configure environments, design robot motions, and generate ground-truth annotations to support customized experimental needs.
A Multi-Robot Trajectory Designer enables users to specify team paths and waypoint sequences for coordinated motion.
A Wildfire Spread Design Tool allows users to define spatial regions of interest that drive dynamic environmental phenomena.
A Multimodal Ground-Truth Labeler exports aligned raw sensor data with instance/semantic segmentation for RGB images, 2D/3D bounding boxes for objects, and instance/semantic annotations for LiDAR point clouds.
In addition, UE5 Graphical Automation Workflows (Blueprint automations) support batch generation of scenarios with different parameter settings. 
Together, these utilities make it straightforward to construct reproducible experiments and systematically vary environmental conditions. Users retain full access to low-level scenario setup when needed, while avoiding the tedious, game-development-heavy setup steps that would otherwise be required.

\begin{figure*}[t]
    \centering

    \begin{subfigure}[t]{0.245\textwidth}
        \centering
        \includegraphics[width=\linewidth]{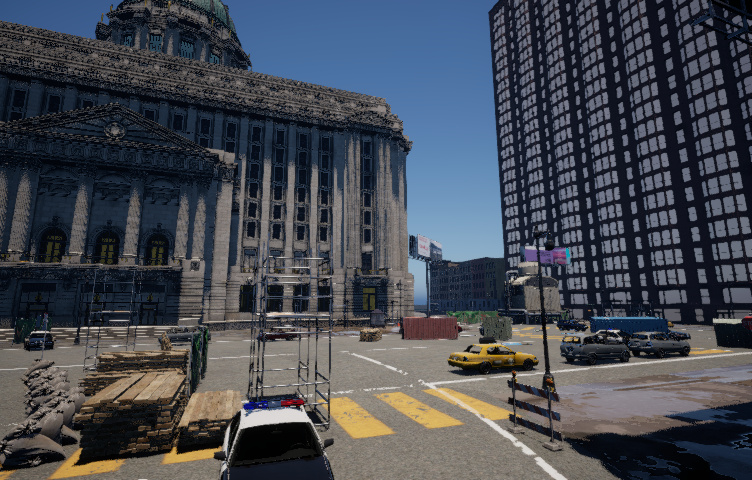}\hspace{1pt}
        \includegraphics[width=\linewidth]{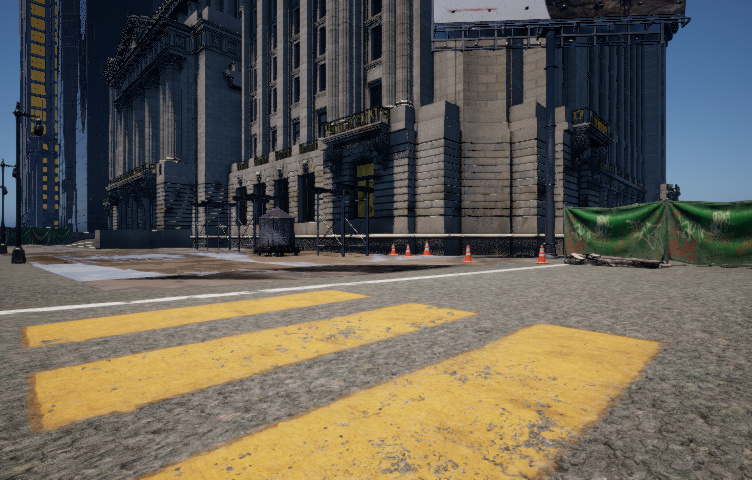}
        \caption{City Sequence.}
        \label{fig:pair1}
    \end{subfigure}
    \hfill
    \begin{subfigure}[t]{0.245\textwidth}
        \centering
        \includegraphics[width=\linewidth]{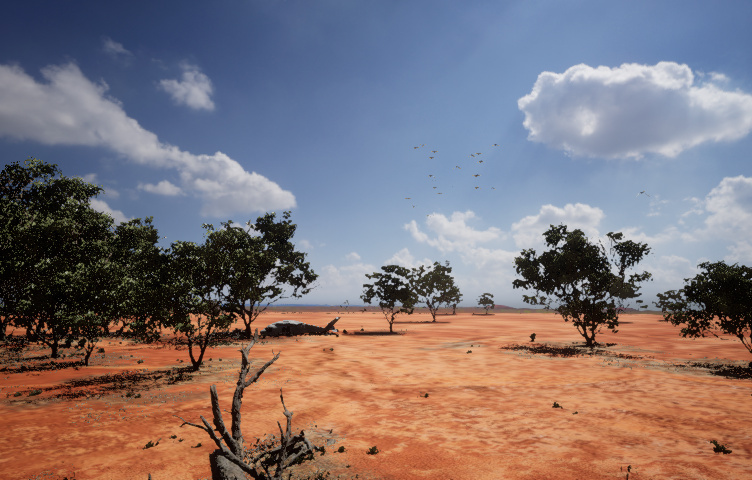}\hspace{1pt}
        \includegraphics[width=\linewidth]{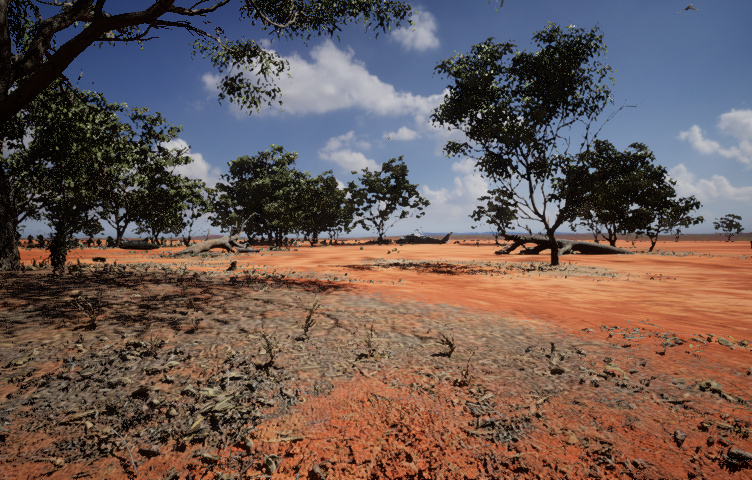}
        \caption{Desert-Perimeter Sequence.}
        \label{fig:pair2}
    \end{subfigure}
    \hfill
    \begin{subfigure}[t]{0.245\textwidth}
        \centering
        \includegraphics[width=\linewidth]{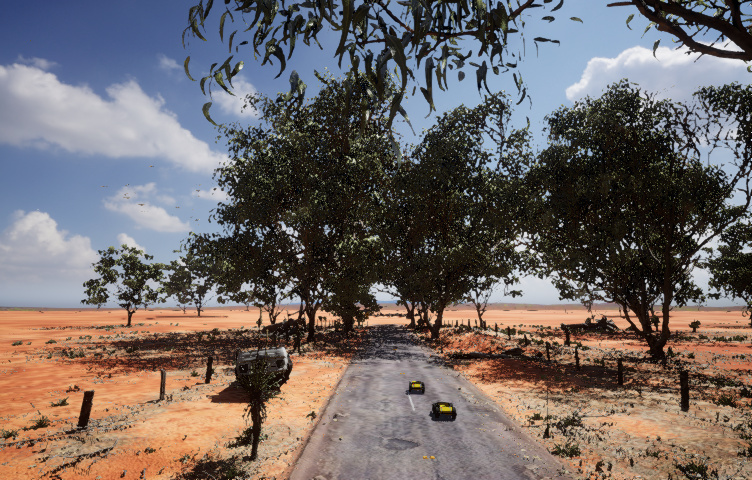}\hspace{1pt}
        \includegraphics[width=\linewidth]{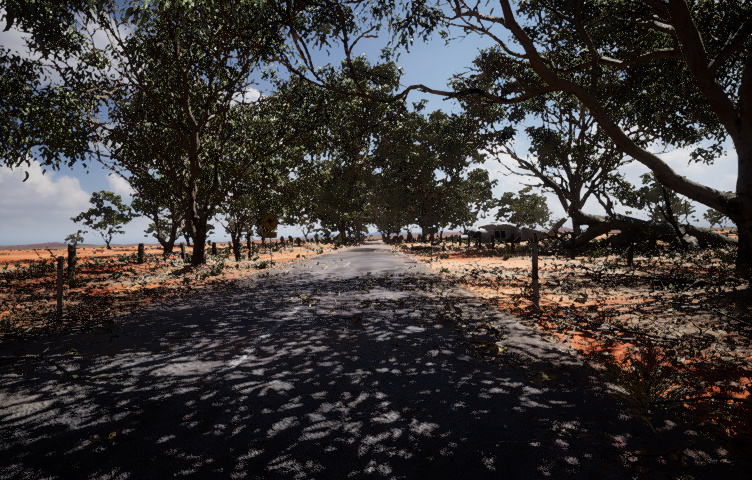}
        \caption{Desert-Center Sequence.}
        \label{fig:pair3}
    \end{subfigure}
    \hfill
    \begin{subfigure}[t]{0.245\textwidth}
        \centering
        \includegraphics[width=\linewidth]{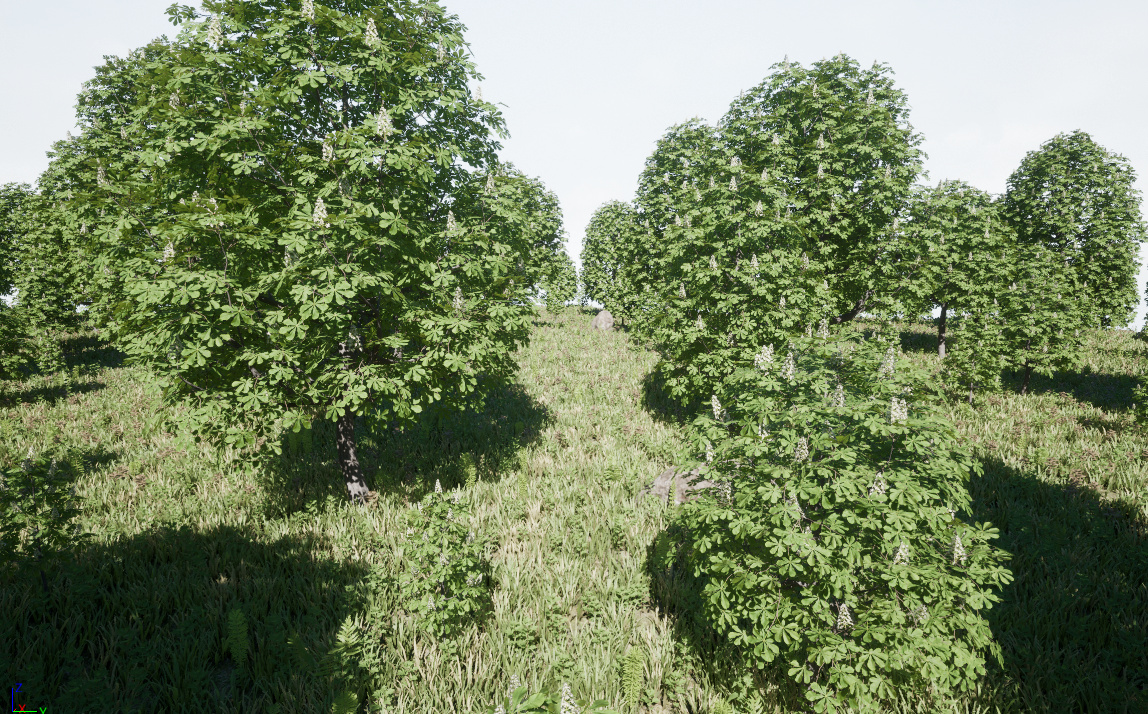}\hspace{1pt}
        \includegraphics[width=\linewidth]{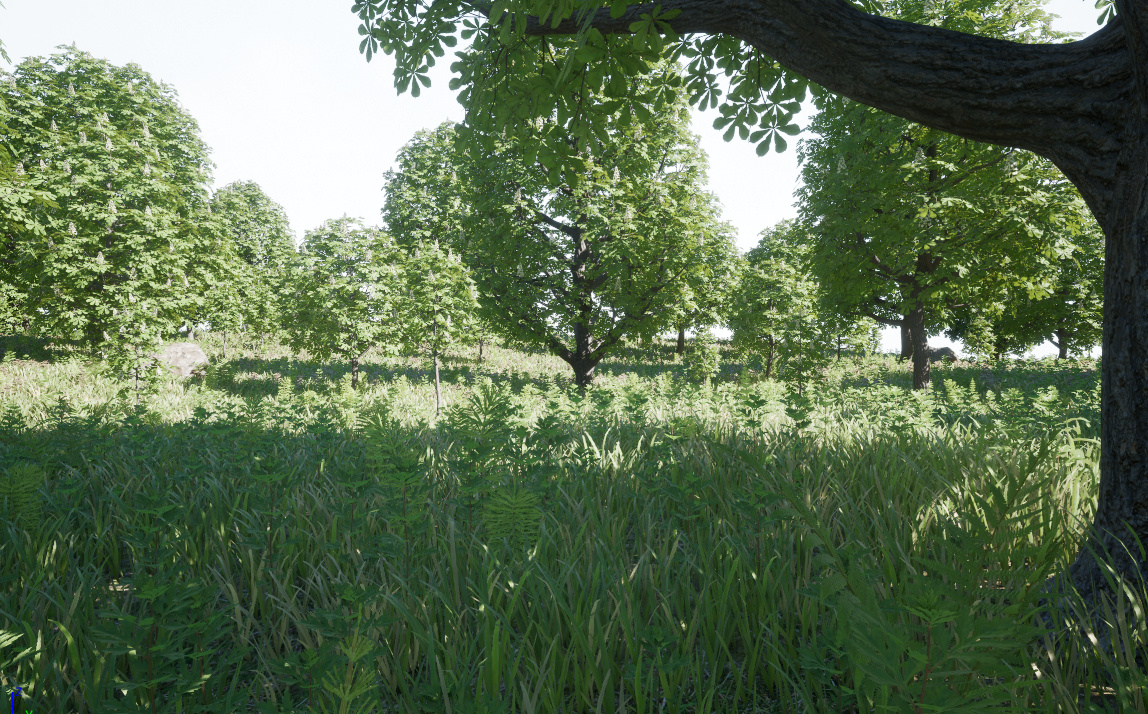}
        \caption{Forest Sequence.}
        \label{fig:pair4}
    \end{subfigure}

    \caption{
    Representative views from each sequence in the collaborative SLAM experiment, with (Top) UAV aerial views and (Bottom) UGV ground views. Each sequence poses different challenges for multi-agent localization as we rely on HERCULES's capabilities for photorealistic lighting and vegetation.
    }
    \label{fig:cslam_twoenvs_side}
\end{figure*}

\section{Closed-Loop Operation}
\label{sec:closedloop}
Used \emph{actively}, HERCULES is a closed-loop evaluation platform: an online planner receives synchronized observations at each step, updates a local map, issues control commands through the navigation stack (Sec.~\ref{sec:autonomousnav}), and advances the simulation under the real-time clock. The same heterogeneous core, mapping, planning, and control used for dataset collection are reused here; the only difference is that trajectories are generated online from live observations rather than designed offline. Concretely, closed-loop operation couples a global step that selects the next goal viewpoint (e.g., a frontier or an information-rich location) with the local planner in Sec.~\ref{sec:autonomousnav}, which generates a dynamically feasible trajectory to reach it; the cycle repeats as new observations arrive. We exercise this mode with our kinodynamic planner in frontier-biased exploration and demonstrate it on a heterogeneous UAV--UGV team in Sec.~\ref{sec:active_perception}. User-defined planners can be plugged in to replace the default global planner, supporting research on planning and multi-robot coordination.

\section{Experiments}
\label{sec:experiments}
We validate HERCULES through three experiments that demonstrate the platform's core capabilities: heterogeneous multi-robot SLAM across diverse environments (Sec.~\ref{sec:experiments-collab-slam}), cooperative 3D object detection with sim-to-real transfer (Sec.~\ref{sec:experiments-collab-perception}), and a closed-loop exploration demonstration with heterogeneous teams (Sec.~\ref{sec:active_perception}). The first two experiments rely on the synchronized multi-robot dataset collection pipeline and deterministic-replay trajectories; the third exercises the same heterogeneous substrate under online planning to demonstrate closed-loop coordination capability.
\subsection{Collaborative SLAM}
\label{sec:experiments-collab-slam}
To evaluate HERCULES as a testbed for large-scale heterogeneous multi-robot localization and mapping, we collect a collaborative SLAM benchmark dataset in HERCULES spanning a wide range of operational conditions. We evaluate our benchmark dataset on single-robot odometry/SLAM using \mbox{OpenVINS \citep{geneva2020openvins}}, \mbox{ORB-SLAM3 \citep{campos2021orbslam3}}, and \mbox{LIO-SAM \citep{shan2020lio}}, and on multi-robot collaborative SLAM using \mbox{ROMAN~\citep{peterson2025roman}} integrated with \mbox{Kimera-RPGO~\citep{tian2022kimera}} back-end.
Our goal is to demonstrate that HERCULES can elicit and surface the realistic failure modes that collaborative SLAM faces at kilometer scale, perceptual aliasing, repetitive semantics, and severe aerial--ground viewpoint baseline, rather than to provide a broad multi-method comparison. We show that these baselines perform well on short, feature-rich sequences with strong inter-robot viewpoint overlap, the regime in which they are typically benchmarked, but degrade on the kilometer-scale sequences with sparse or repetitive structure and large aerial--ground viewpoint baselines that HERCULES generates, demonstrating its value as a testbed for advancing heterogeneous multi-robot SLAM.

\begin{table*}[ht]
    \centering
    \footnotesize
    \setlength{\tabcolsep}{6pt}
    \renewcommand{\arraystretch}{1.25}

    \caption{RMS ATE (m) for single-robot odometry/SLAM baselines across four sequences. We report per-robot errors for UGV1/UGV2 and UAV1/UAV2. When two values are shown, they indicate results without loop closures / with loop closures.}
    \label{tab:collaborative_slam_odometry} 

    \begin{tabular}{|l|l|cc|cc|}
    \hline
    \multirow{2}{*}{\textbf{Baseline}} &
    \multirow{2}{*}{\textbf{Sequence}} 
    &
    \multicolumn{2}{c|}{\textbf{UGVs ATE [m]}} &
    \multicolumn{2}{c|}{\textbf{UAVs ATE [m]}} \\ 
    \cline{3-6}
      & &
     \textbf{UGV1} & \textbf{UGV2} &
     \textbf{UAV1} & \textbf{UAV2} \\
    \hline

    \multirow{4}{*}{\parbox{2.5cm}{OpenVINS \mbox{\citep{geneva2020openvins}}}}
          & City     & 4.90 & 6.80 & 15.40 & 10.21  \\
     \cline{2-2}
          & Desert-Perimeter & 3.92 & 4.53 & 17.70 & 14.05 \\
     \cline{2-2}
        & Desert-Center    & 9.76 & 7.66 & 27.55 & 10.72  \\[0.1em]
      \cline{2-2}
          & Forest    & 0.98 & 1.36 & 1.49 & 2.50   \\
    \hline

    \multirow{4}{*}{\parbox{2.5cm}{ORB-SLAM3 \mbox{\citep{campos2021orbslam3}}}}
          & City     & 1.27 / 1.03 & 0.68 / 1.66 & 0.70 / 0.91 & 8.64 / 7.15  \\
     \cline{2-2}
          & Desert-Perimeter & 3.53 / 5.40 & 10.88 / 10.56 & 4.16 / 4.34 & 23.14 / 24.13 \\
     \cline{2-2}
        & Desert-Center    & 1.06 / 1.82 & 1.72 / 1.19 & 4.10 / 6.71 & 4.12 / 37.85  \\[0.1em]
      \cline{2-2}
          & Forest    
          & 0.52 / 0.46 
          & 0.67 / 0.56 
          & 0.41 / 0.59 
          & 0.60 / 23.96 \\
    \hline

    \multirow{4}{*}{\parbox{2.5cm}{LIO-SAM \mbox{\citep{shan2020lio}}}}
          & City     & 1.33 & 1.44 & 0.18 & 0.06 \\
    \cline{2-2}
          & Desert-Perimeter & 0.23 & 0.29 & 0.06 & 0.10  \\
    \cline{2-2}  
          & Desert-Center    & 0.07 & 0.10 & 0.10 & 0.12   \\[0.1em]
    \cline{2-2}
          & Forest    & 0.43 & 0.53 & 0.04 & 0.07   \\
    \hline

    \end{tabular}
\end{table*}

\begin{table*}[t]
    \centering
    \footnotesize
    \setlength{\tabcolsep}{4pt}
    \renewcommand{\arraystretch}{1.25}
    \caption{RMS ATE (m) for multi-robot SLAM on four sequences using ROMAN integrated with Kimera-RPGO.}
    \label{tab:collaborative_slam_full}
    \resizebox{\textwidth}{!}{%
    \begin{tabular}{|l|l|cccccc|}
    \hline
    \multirow{2}{*}{\textbf{Baseline}} &
    \multirow{2}{*}{\textbf{Sequence}} &
    \multicolumn{6}{c|}{\textbf{Robot pairs ATE [m]}} \\
    \cline{3-8}
     & &
     \textbf{UGV1--UGV2} & \textbf{UGV1--UAV1} & \textbf{UGV1--UAV2} & \textbf{UGV2--UAV1} & \textbf{UGV2--UAV2} & \textbf{UAV1--UAV2} \\
    \hline
    \multirow{4}{*}{\parbox{2.5cm}{ROMAN \mbox{\citep{peterson2025roman}}}}
          & City             & 1.53 & 84.49 & 74.20 & 1.58 & 57.27 & 0.41 \\
      \cline{2-2}
          & Desert-Perimeter & 1.01 & 3.08 & 1.29 & 119.03 & 2.02 & 1.42 \\
      \cline{2-2}
          & Desert-Center    & 4.06 & 0.88 & 0.90 & 106.52 & 106.05 & 1.16 \\[0.1em]
      \cline{2-2}
          & Forest           & 1.08 & 43.88 & 1.43 & 26.48 & 0.93 & 49.17 \\
    \hline
    \end{tabular}%
    }
\end{table*}

\subsubsection{Dataset Collection.} 
We collect data in three representative large-scale environments, city, desert, and forest, each containing diverse visual textures, vegetation density, and illumination conditions (see \mbox{Figure~\ref{fig:cslam_twoenvs_side}}). Overall, four data sequences are collected to pose distinct sensing challenges:
\begin{itemize}
    \item \textbf{City}: UGVs and UAVs navigate street blocks in the city environment. The large scale of the environment complicates vision-based SLAM despite its rich visual details and abundance of objects.
    \item \textbf{Desert-Perimeter}: Robots traverse the perimeter of a grove of trees and weathered objects in the desert environment. This sequence is challenging due to sparse foliage, large scale, and dynamic birds. 
    \item \textbf{Desert-Center}: Similar to \textit{Desert-Perimeter}, but robots traverse a straight road inside the grove. Overlapping vegetation and increased shadows make object segmentation and tracking difficult.
    \item \textbf{Forest}: In the forest environment, UGVs traverse dense vegetation while UAVs fly through the trees. Visually repetitive structures and limited semantic diversity pose challenges to loop-closure detection.
\end{itemize}
For each sequence, data are collected from two UGVs and two UAVs, including synchronized stereo images, depth images, and LiDAR point clouds at 20 Hz, as well as IMU measurements at 500 Hz. Executed trajectories are designed using the Complementary Coverage mode and include both intra-robot and inter-robot loop closures, with lengths ranging from 359 to 945 meters. Each trajectory begins with a static period followed by calibration motions to support proper initialization when needed. Dynamic objects are disabled except for birds during data collection.

\subsubsection{Implementation.} 
For single-robot odometry/SLAM, we run OpenVINS, ORB-SLAM3, and LIO-SAM on all four robots individually to cover different sensor modalities in our dataset. OpenVINS uses stereo images and IMU data as input; ORB-SLAM3 is run in the \mbox{RGB-D} setting, both with and without loop-closure detection; LIO-SAM uses LiDAR and IMU measurements and is run with loop-closure detection. For multi-robot SLAM, we run ROMAN with LIO-SAM odometry and Kimera-RPGO as the back-end. ROMAN is run distributively between robot pairs for all combinations (i.e., UGV--UGV, UAV--UAV, UGV--UAV).
We use the official open-source implementation of each baseline throughout the experiments. Where feasible, parameters are shared across sequences and robot platforms, and default settings are preferred. In some cases, sequence- and robot-specific parameters are tuned for performance, such as the maximum depth range for ROMAN and the number of tracked features in OpenVINS.
For reproducibility, all collected datasets, environment configurations, sensor calibrations, and motion trajectories are publicly released with HERCULES.

\subsubsection{Results.} 
\emph{Single-robot SLAM.} We compute the root-mean-square absolute trajectory error (RMS ATE) for the single-robot odometry/SLAM methods using evo~\citep{grupp2017evo} and report the results in Table~\ref{tab:collaborative_slam_odometry}. 
Each method succeeds on all robots for at least one sequence, with success defined as an RMS ATE less than or equal to 2.5 meters. However, OpenVINS struggles in the city and desert environments due to the limited visual features within close range. This is especially apparent for the UAVs, where the RMS ATE for both drones exceeds 10\,m. ORB-SLAM3 exhibits a related failure mode: its loop closure module frequently accepts spurious matches due to perceptual aliasing in the repetitive desert and forest environments. As a result, enabling loop closures can sometimes cause the RMS ATE to increase catastrophically. For example, the RMS ATE for UAV 2 jumps from 4.12\,m to 37.85\,m on the Desert-Center sequence and from 0.60\,m to 23.96\,m on the Forest sequence. LIO-SAM succeeds and achieves the best performance on all sequences by leveraging LiDAR measurements, which provide accurate geometric information and are robust to visual aliasing. Therefore, we use the LIO-SAM as the odometry front-end for ROMAN in our multi-robot SLAM experiments.

\begin{figure*}[t!]
    \centering
    \captionsetup[subfigure]{justification=centering}

    \begin{minipage}{0.97\textwidth}
        \centering
        \begin{subfigure}{0.33\textwidth}
            \centering
            \includegraphics[width=\textwidth]{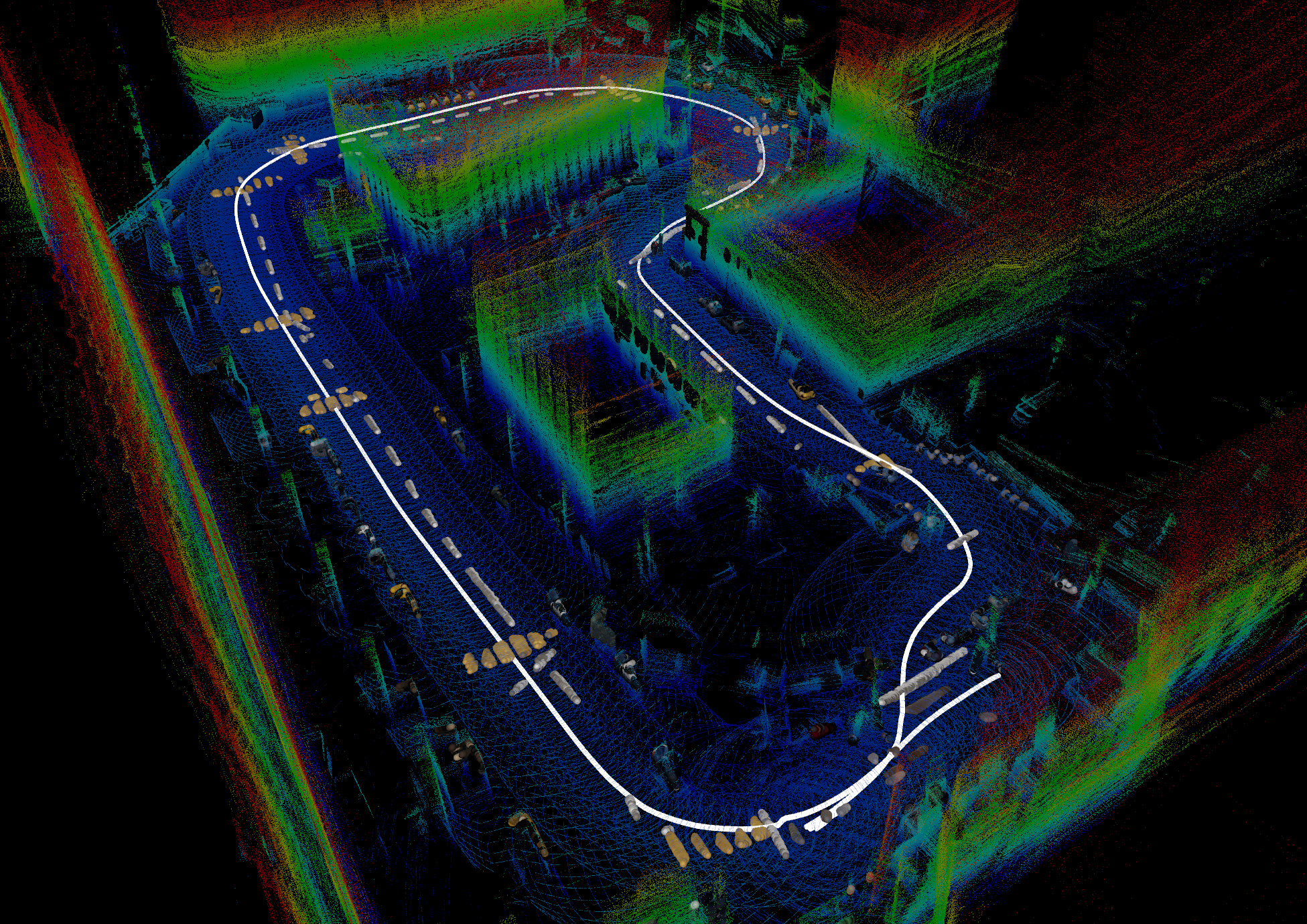}
        \end{subfigure}\hfill
        \begin{subfigure}{0.33\textwidth}
            \centering
            \includegraphics[width=\textwidth]{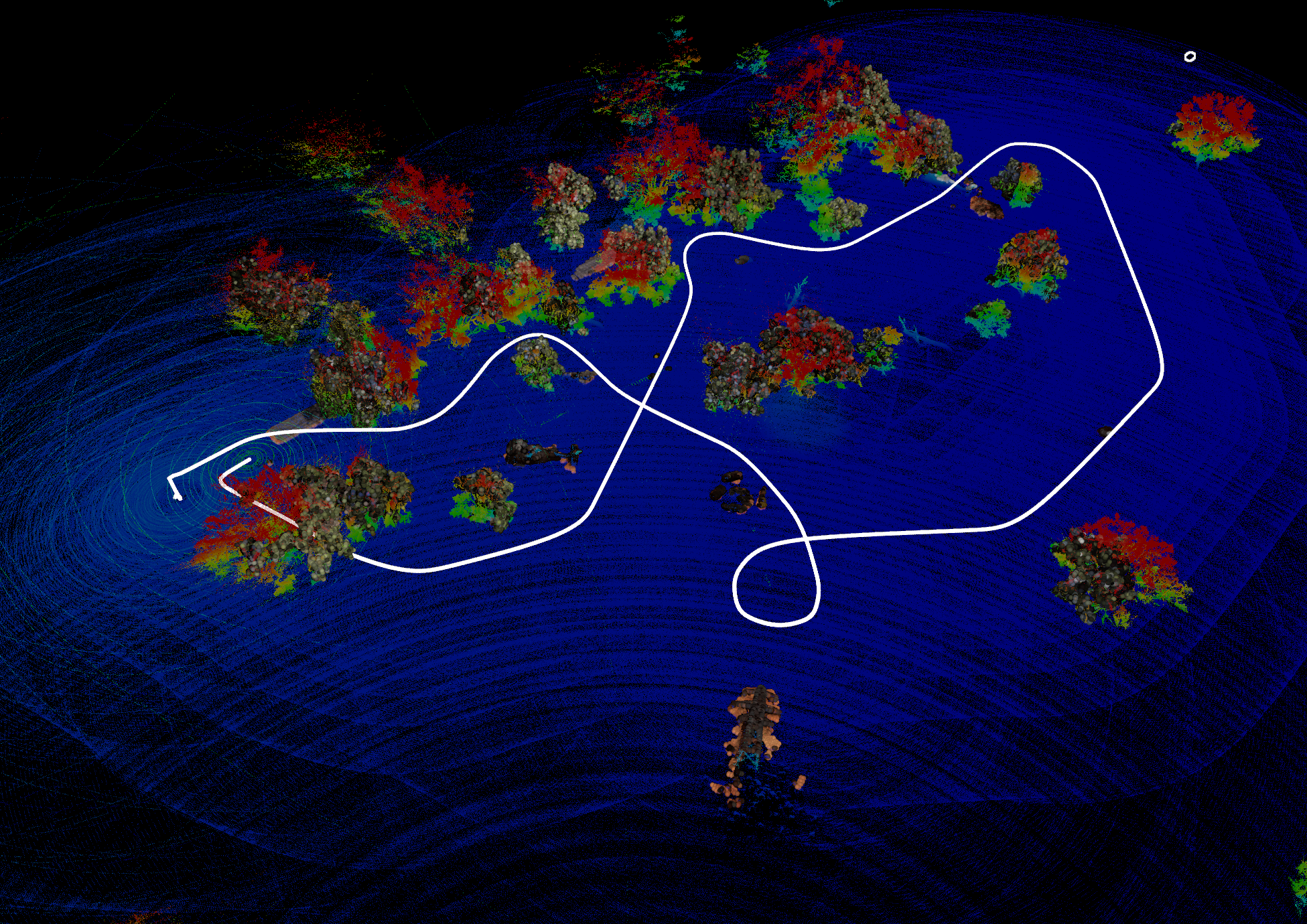}
        \end{subfigure}
        \begin{subfigure}{0.33\textwidth}
            \centering
            \includegraphics[width=\textwidth]{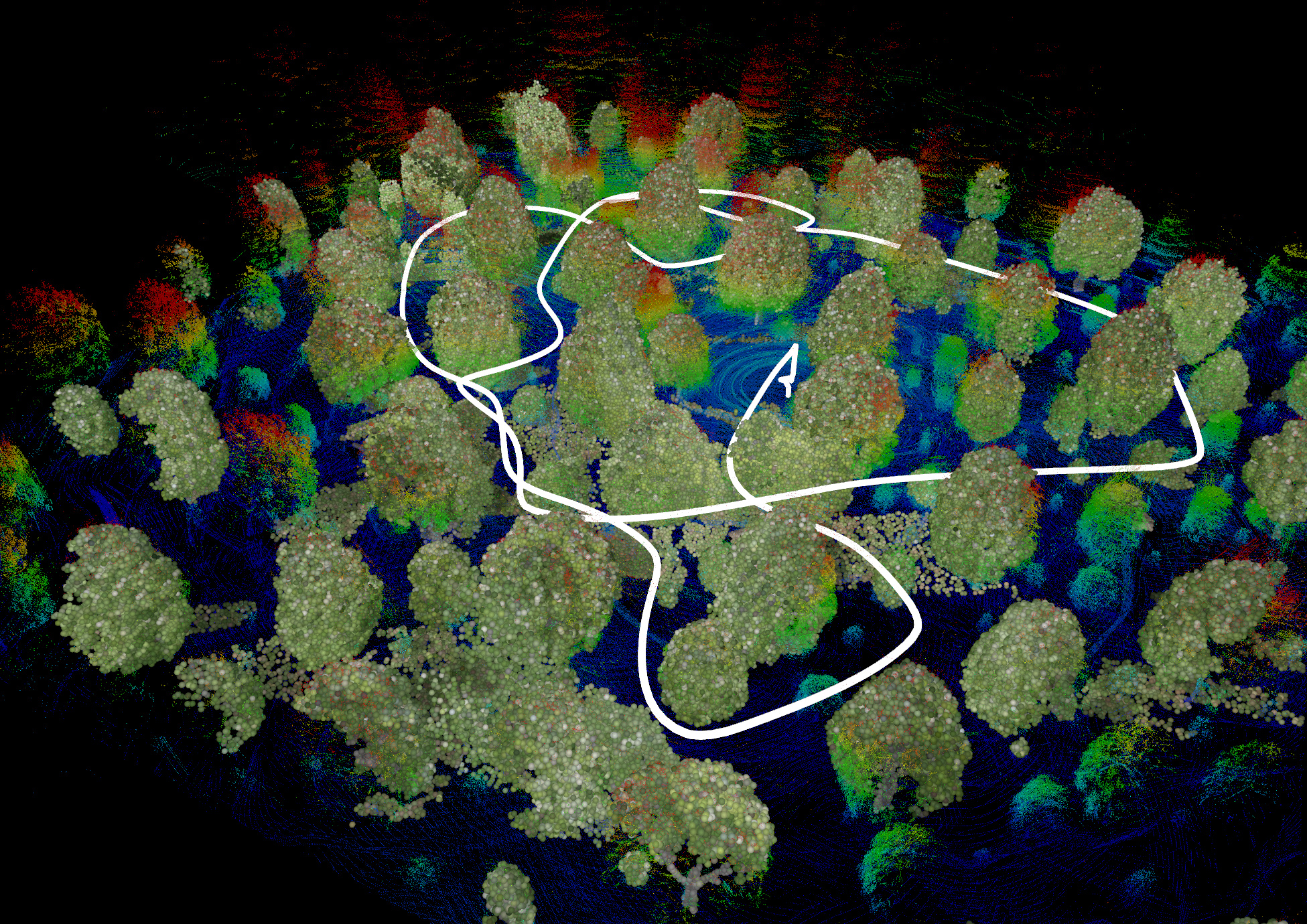}
        \end{subfigure}
    \end{minipage}

    \vspace{0.2em}

    \begin{minipage}{0.97\textwidth}
        \centering
        \begin{subfigure}{0.33\textwidth}
            \centering
            \includegraphics[width=\textwidth]{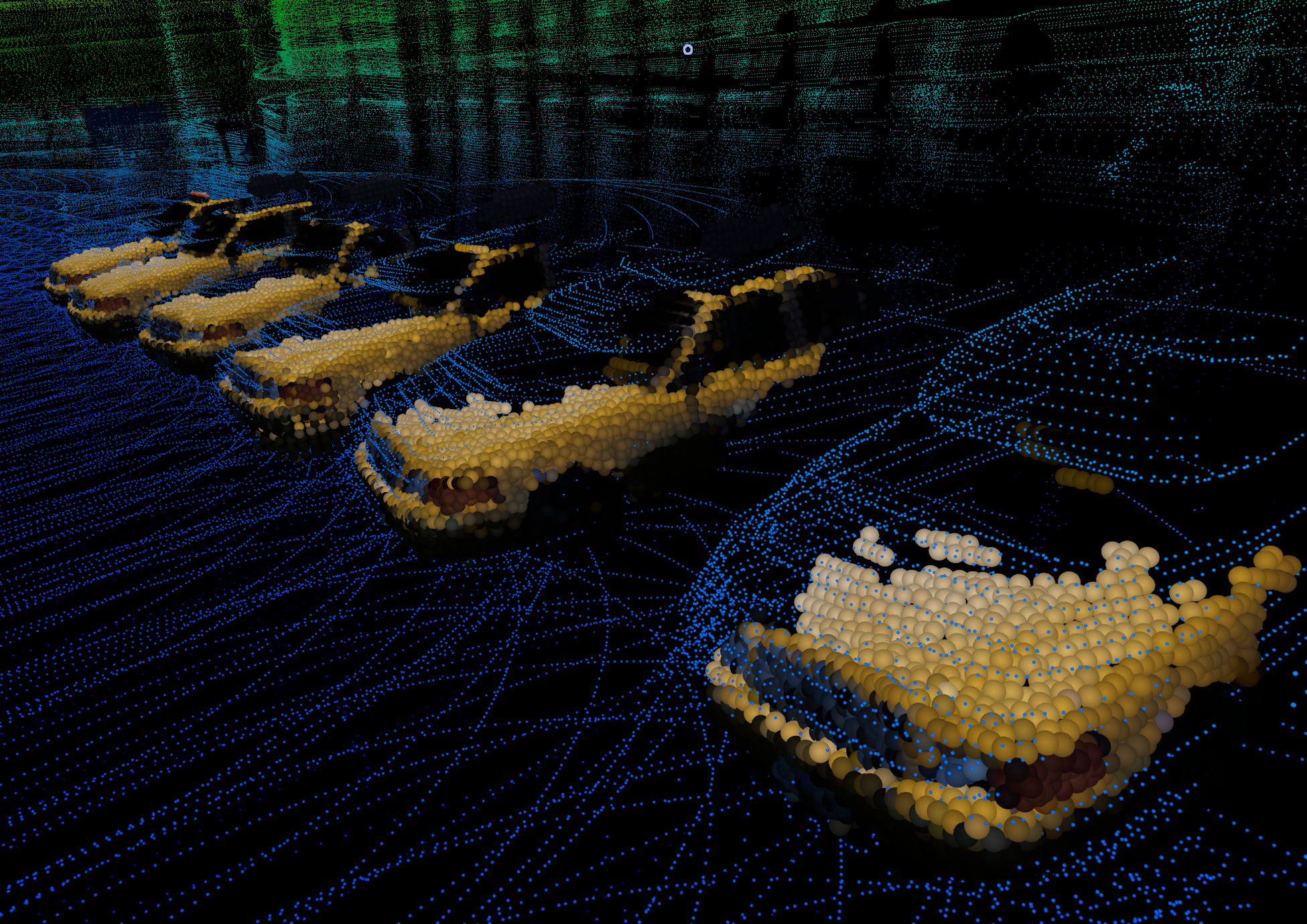}
        \end{subfigure}\hfill
        \begin{subfigure}{0.33\textwidth}
            \centering
            \includegraphics[width=\textwidth]{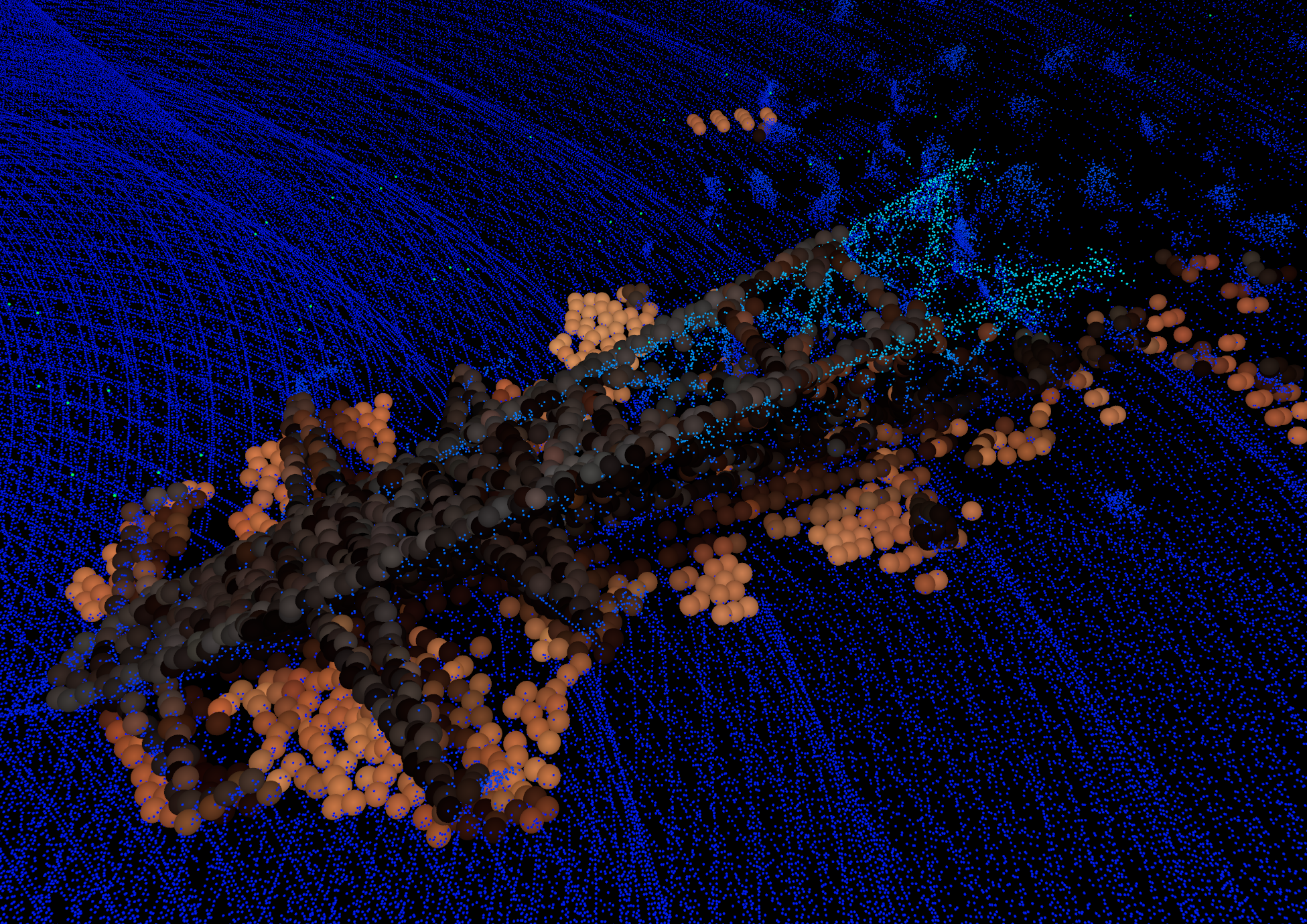}
        \end{subfigure}
        \begin{subfigure}{0.33\textwidth}
            \centering
            \includegraphics[width=\textwidth]{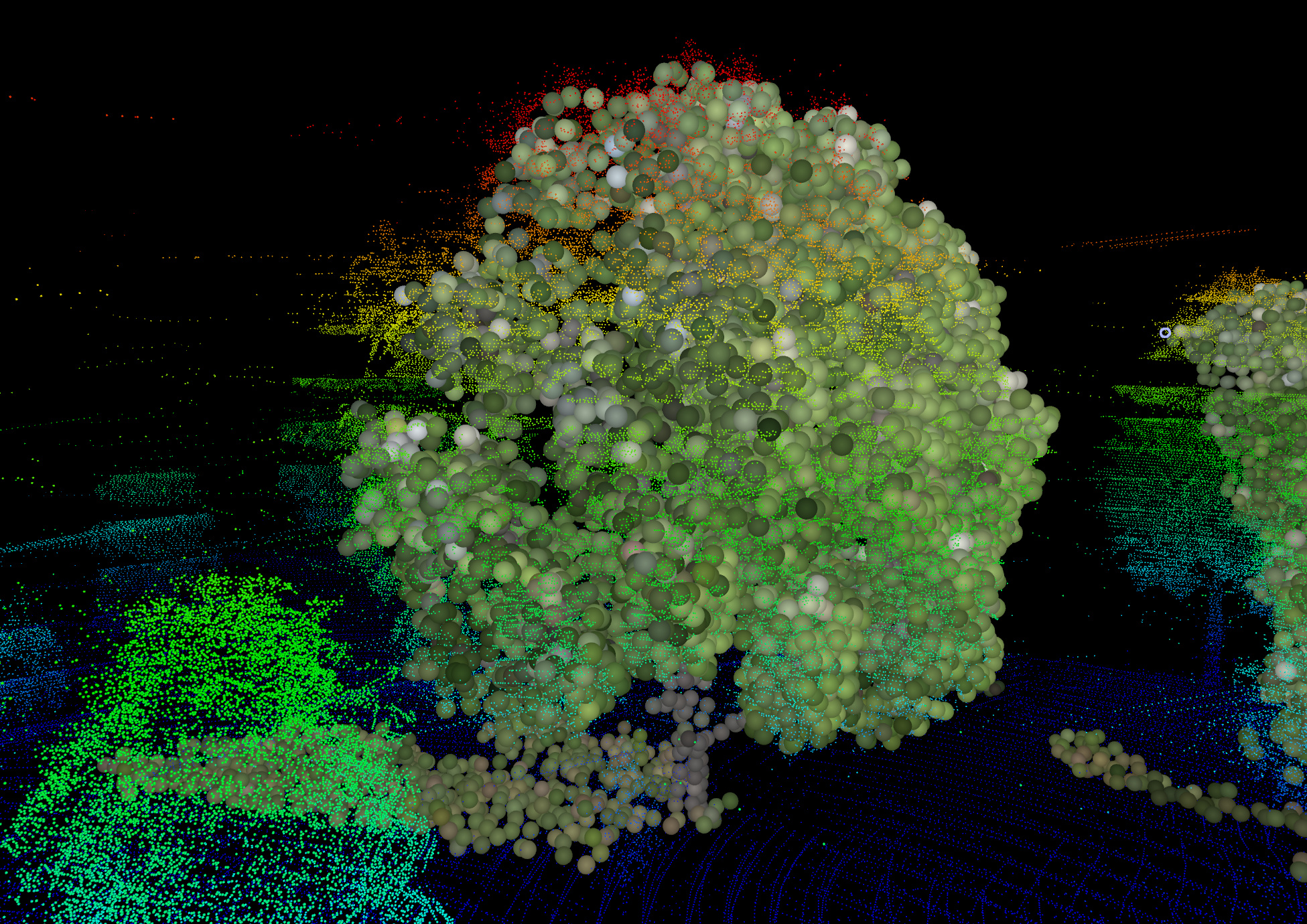}
        \end{subfigure}
    \end{minipage}

    \caption{Object maps constructed by ROMAN and the corresponding robot trajectory overlaid onto the environment point cloud. The environments are as follows: (Left) City, (Middle) Desert, and (Right) Forest. Object point colors are included for visual clarity. We provide views of the high-level environment structure (Top) and low-level objects (Bottom).}
    \label{fig:collaborative_slam_maps}
\end{figure*}

\begin{figure*}[t]
    \centering
    \captionsetup[subfigure]{justification=centering}

    \includegraphics[width=0.97\textwidth]{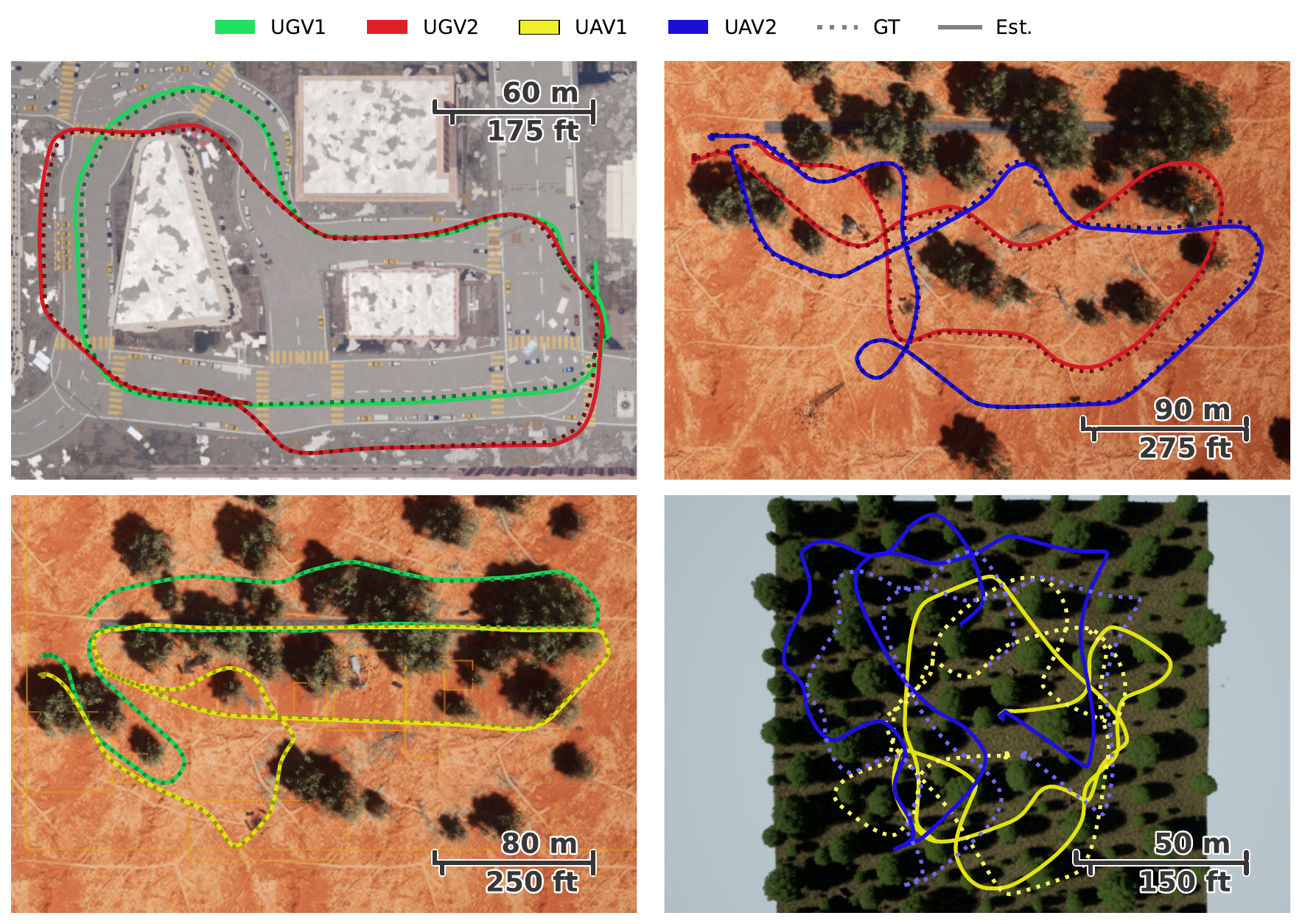}

    \caption{Ground truth (GT) versus estimated (Est.) trajectories of various sequences in Table~\ref{tab:collaborative_slam_full} overlaid on the HERCULES simulation environment. The sequences are as follows: (Top left) City UGV1-UGV2, (Top right) Desert-Perimeter UGV2-UAV2, (Bottom left) Desert-Center UGV1-UAV1, and (Bottom right) Forest UAV1-UAV2.}
    \label{fig:collaborative_slam_vis}
\end{figure*}

\emph{Multi-robot SLAM.} We align the ROMAN-estimated trajectories of each robot pair with the corresponding GT trajectories using SE(3) Umeyama alignment~\citep{umeyama1991least}, and report the RMS ATE in Table~\ref{tab:collaborative_slam_full}. With an RMS ATE of less than 5\,m as the success criterion for multi-robot SLAM, ROMAN successfully aligns trajectories for homogeneous robot pairs in all but one sequence: UAV1--UAV2 in Forest. Additionally, ROMAN achieves at least one successful alignment for every robot pair across all four sequences, despite being designed primarily for homogeneous multi-robot map alignment. In those cases, even as the heterogeneous robots view the same scene from drastically different angles, ROMAN is able to find object-based view-invariant loop closures to properly align any heterogeneous configuration. Example objects detected by ROMAN can be seen in Figure~\ref{fig:collaborative_slam_maps}, and various examples of successfully aligned trajectories are shown in Figure~\ref{fig:collaborative_slam_vis}.

Our results also show that ROMAN experiences failures in some sequences, occurring more frequently in the City and Forest sequences and for heterogeneous robot pairs. ROMAN leverages FastSAM~\citep{zhao2023fast} for open-set segmentation, which is prone to generating inconsistent or over-segmented objects~\citep{peterson2025roman}; especially under significant lighting variation and object occlusions. Additionally, as ROMAN uses semantic information for robust data association, it may struggle in environments with repetitive semantics. These issues tend to be exacerbated for heterogeneous robot pairs due to large viewpoint differences, scale changes, and partial overlap in observed objects, as well as in environments with repetitive visual structures, such as forests.
A failure case in the Forest sequence is shown in the bottom-right of Figure~\ref{fig:collaborative_slam_vis}. In contrast, the desert environment, with its spatially-separated and semantically-distinct objects, leads to more successful alignments.

\begin{figure*}[!t]
    \centering
    \begin{minipage}[c]{0.54\textwidth}
        \centering
        \includegraphics[width=\linewidth]{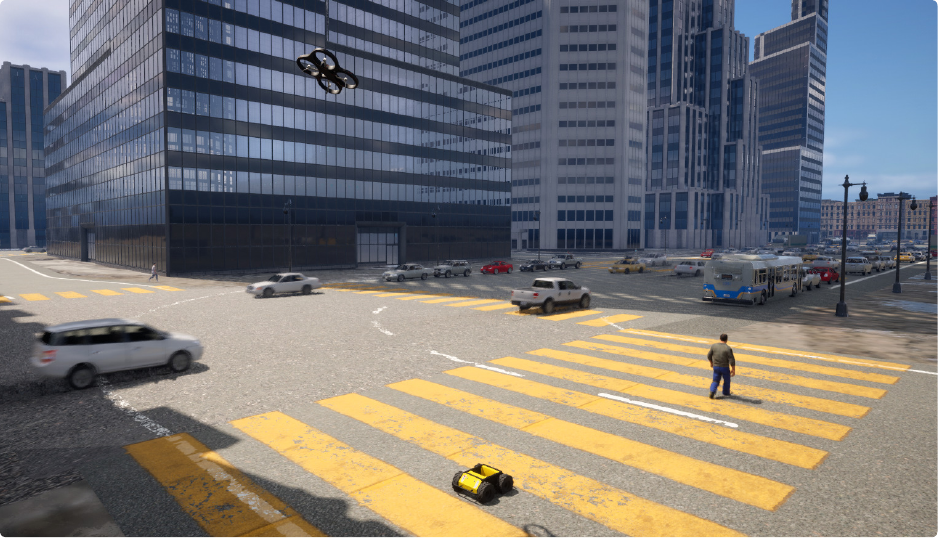}
    \end{minipage}%
    \hfill
    \begin{minipage}[c]{0.44\textwidth}
        \centering
        \includegraphics[width=\linewidth]{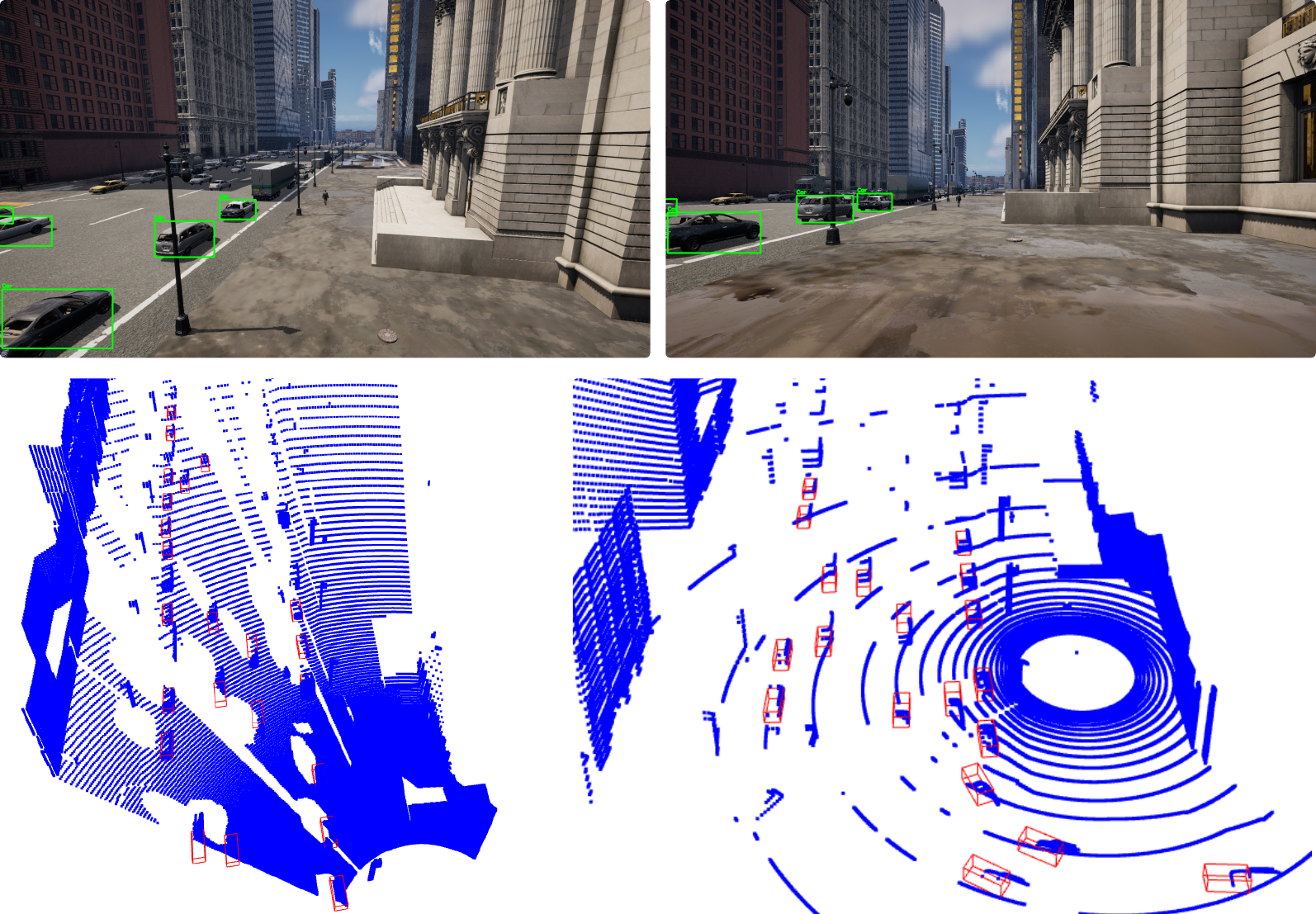}
    \end{minipage}
    \caption{
    Representative sample from the HERCULES cooperative vehicle--infrastructure dataset. 
    (Left) Third-person view showing the heterogeneous robot team: a UAV provides overhead infrastructure-view sensing while a UGV navigates at street level among dynamic traffic and pedestrians.
    (Right) Synchronized sensor data from both platforms. Top row: infrastructure/UAV RGB image (left) and vehicle/UGV RGB image (right) with 2D annotations. Bottom row: corresponding LiDAR point clouds with 3D bounding boxes from infrastructure view (left) and vehicle view (right).
    The paired viewpoints provide complementary visibility and geometry, enabling stronger 3D object detection.}
    \label{fig:synthetic_v2x_data}
\end{figure*}

\begin{table*}[t!]
\centering
\scriptsize
\setlength{\tabcolsep}{2.6pt}
\renewcommand{\arraystretch}{1.15}
\caption{Sim-to-sim baseline: cooperative 3D Car detection trained and evaluated on HERCULES synthetic data ($\text{IoU}=0.50$).
KITTI-style AP11/AP40 are reported across distance bins and overall (unweighted mean over bins).}
\label{tab:vic_iou050_1class}
\resizebox{\textwidth}{!}{%
\begin{tabular}{|l|cc|cc|cc|cc|cc|cc|cc|cc|}
\hline\hline
\multirow{3}{*}{\textbf{Fusion}} &
\multicolumn{8}{c|}{\textbf{KITTI $AP_{3D}$}} &
\multicolumn{8}{c|}{\textbf{KITTI $AP_{BEV}$}} \\
\cline{2-17}
&
\multicolumn{2}{c|}{\textbf{0--30\,m}} &
\multicolumn{2}{c|}{\textbf{30--50\,m}} &
\multicolumn{2}{c|}{\textbf{50--100\,m}} &
\multicolumn{2}{c|}{\textbf{Overall}} &
\multicolumn{2}{c|}{\textbf{0--30\,m}} &
\multicolumn{2}{c|}{\textbf{30--50\,m}} &
\multicolumn{2}{c|}{\textbf{50--100\,m}} &
\multicolumn{2}{c|}{\textbf{Overall}} \\
\cline{2-17}
&
\textbf{AP11} & \textbf{AP40} &
\textbf{AP11} & \textbf{AP40} &
\textbf{AP11} & \textbf{AP40} &
\textbf{AP11} & \textbf{AP40} &
\textbf{AP11} & \textbf{AP40} &
\textbf{AP11} & \textbf{AP40} &
\textbf{AP11} & \textbf{AP40} &
\textbf{AP11} & \textbf{AP40} \\
\hline
UGV-only
  & 72.34 & 72.17 & 52.31 & 54.58 & 17.19 & 13.99 & 47.28 & 46.92
  & 72.47 & 72.29 & 61.10 & 57.90 & 17.69 & 16.54 & 50.42 & 48.91 \\
\hline
UAV-only
  & 54.39 & 52.38 & 62.79 & 59.26 & 62.98 & 69.08 & 60.06 & 60.24
  & 54.41 & 54.84 & 63.00 & 59.46 & 71.86 & 69.25 & 63.09 & 61.18 \\
\hline
Late Fusion
  & \textbf{71.24} & \textbf{73.32} & \textbf{69.65} & \textbf{73.67} & \textbf{62.57} & \textbf{63.81} & \textbf{67.82} & \textbf{70.27}
  & \textbf{71.29} & \textbf{73.39} & \textbf{70.09} & \textbf{74.25} & \textbf{62.76} & \textbf{66.35} & \textbf{68.05} & \textbf{71.33} \\
\hline\hline
\end{tabular}%
}
\end{table*}

\subsubsection{Discussion.} 
These findings demonstrate that HERCULES can accurately reproduce heterogeneous multi-robot SLAM conditions at a kilometer scale while maintaining precise ground-truth benchmarking for quantitative evaluation. This experiment highlights that the photorealistic rendering and physically accurate sensor simulation in HERCULES allow researchers to create a variety of challenging environmental conditions, including perceptual aliasing, poor features, and realistic lighting. HERCULES supports even more configurations beyond those included in this experiment; including day/night cycles, dynamic environmental phenomena (e.g., fires, floods), and additional dynamic agents (e.g., cars, pedestrians). Thus, HERCULES enables researchers to investigate future challenges in heterogeneous multi-robot SLAM within a convenient simulation environment, largely reducing the need for time-consuming and labor-intensive real-world data collection. The successes on simpler sequences and failures on more challenging sequences also  validate the simulator’s ability to evaluate SLAM pipelines and identify critical failure modes before real-world deployment.

\subsection{Collaborative Perception: 3D Object Detection}
\label{sec:experiments-collab-perception}
We demonstrate HERCULES’s utility for heterogeneous multi-robot collaborative perception through a 3D object detection task. Analogous to vehicle-infrastructure cooperation (VIC) in autonomous driving, this experiment fuses UGV and UAV sensor data to improve overall detection performance. Additionally, we validate the sim-to-real fidelity of HERCULES by using our synthetic data to augment the real-world DAIR-V2X benchmark dataset~\citep{yu2022dair}, demonstrating its effectiveness in advancing real-world computer vision models.

To mimic the VIC setting, we construct a heterogeneous team with one UGV on the sidewalk and one UAV overhead in the \textit{City} world, where the UGV takes the vehicle role and the UAV takes the infrastructure role. Their trajectories follow the Leader--Follower motion pattern, with the UAV maintaining overhead coverage along the UGV path.
Sensor types, rates, and extrinsics mirror the DAIR-V2X configurations to ensure compatibility with its protocols and metrics. In total, we generate 6,000 time-synchronized UGV–UAV frame pairs containing RGB images, LiDAR point clouds, sensor poses, and ground-truth 3D bounding boxes (see Figure~\ref{fig:synthetic_v2x_data}). As the data interface matches DAIR-V2X's VIC-Sync conventions (paired frames with sub-10\,ms skew and cooperative annotations), models trained on our data can be directly resumed on DAIR-V2X data.

\subsubsection{Experimental Setup and Training Protocol.} 
We adopt the DAIR-V2X late-fusion PointPillars \citep{lang2019pointpillars} backbone as the baseline to evaluate two key capabilities of our simulator: (i) the efficacy of multi-robot sensor fusion over single-robot perception using HERCULES synthetic data, and (ii) the sim-to-real transferability of features learned via HERCULES pre-training when fine-tuned on real-world data. To assess these capabilities, we conduct a \emph{sim-to-sim} experiment where PointPillars is trained and evaluated on the HERCULES synthetic dataset, and a \emph{sim-to-real} experiment, where PointPillars is trained using both synthetic and real-world data and evaluated on real-world data. We focus on the Car class in both experiments, while HERCULES supports annotation generation for all object categories. 
Specifically, for the \emph{sim-to-sim} experiment, we train PointPillars from scratch on the HERCULES synthetic dataset, select the best checkpoint on the synthetic validation set, and evaluate it on a held-out synthetic test set.

For the \emph{sim-to-real} experiment, we compare models trained using the following strategies:
\begin{itemize}
    \item \textbf{From-Scratch:} trained on DAIR-V2X real data only.
    \item \textbf{FT-Unfrozen:} pretrained on HERCULES synthetic data, then finetuned on DAIR-V2X training data with all layers unfrozen.
    \item \textbf{FT-Frozen:} pretrained on HERCULES, then finetuned on DAIR-V2X with the backbone frozen for the first five epochs before unfreezing.
\end{itemize}
All three models are trained for the same total epoch budget using identical loss functions, augmentations, and learning-rate schedules.
Following standard practice in 3D detection, we select the best-performing checkpoint on the validation set for each method and report final results on the held-out test set.
We evaluate at both the standard $\text{IoU}=0.50$ and the strict $\text{IoU}=0.70$ threshold, using KITTI-style AP40 as the primary metric, and report average precision for both 3D and bird's-eye-view (BEV) detection.

\begin{table}[t!]
\centering
\scriptsize
\setlength{\tabcolsep}{4pt}
\renewcommand{\arraystretch}{1.15}
\caption{Sim-to-real transfer: late-fusion Car detection on the DAIR-V2X test set (KITTI-style AP40, $\text{IoU}=0.70$). All models are trained using the same total epoch budget; the reported epoch is the best validation checkpoint.
The best results are shown in bold; $\Delta$ is relative to From-Scratch.}
\label{tab:main_results_pretraining}
\begin{tabular}{|l|c|c|c|c|c|}
\hline\hline
\textbf{Method} & \textbf{BEV AP40} & \textbf{$\Delta$ BEV} & \textbf{3D AP40} & \textbf{$\Delta$ 3D} & \textbf{Epoch} \\
\hline
From-Scratch            & 50.21 & ---   & 42.62 & ---   & 19 \\
FT-Unfrozen    & 54.24 & $+4.03$ & 46.67 & $+4.05$ & 19 \\
\textbf{FT-Frozen}      & \textbf{54.23} & $\mathbf{+4.02}$ & \textbf{46.73} & $\mathbf{+4.11}$ & 20 \\
\hline\hline
\end{tabular}
\end{table}

\begin{table}[t!]
\centering
\scriptsize
\setlength{\tabcolsep}{3.8pt}
\renewcommand{\arraystretch}{1.15}
\caption{Sim-to-real transfer: the pretraining finding evaluated under two evaluation protocols (Car 3D AP, $\text{IoU}=0.70$). KITTI-style uses 40-point interpolation (AP40); DAIR-V2X uses continuous AP. Both protocols confirm that synthetic pretraining improves performance, with FT-Frozen yielding the largest gain.}
\label{tab:eval_crossval}
\begin{tabular}{|l|cc|cc|}
\hline\hline
\multirow{2}{*}{\textbf{Method}} &
\multicolumn{2}{c|}{\textbf{KITTI (AP40)}} &
\multicolumn{2}{c|}{\textbf{DAIR-V2X (AP)}} \\
\cline{2-5}
 & \textbf{3D AP} & \textbf{$\Delta$} & \textbf{3D AP} & \textbf{$\Delta$} \\
\hline
From-Scratch          & 42.62 & ---     & 43.36 & ---     \\
FT-Unfrozen  & 46.67 & $+4.05$ & 45.76 & $+2.40$ \\
\textbf{FT-Frozen}    & \textbf{46.73} & $\mathbf{+4.11}$ & \textbf{45.87} & $\mathbf{+2.52}$ \\
\hline\hline
\end{tabular}
\end{table}

\subsubsection{Results.}
\emph{Sim-to-sim baseline.}
Table~\ref{tab:vic_iou050_1class} reports cooperative detection trained and evaluated on HERCULES synthetic data at $\text{IoU}=0.50$. Late fusion outperforms both single-robot configurations overall, but the more informative pattern is across distance: the UGV is strongest at close range and falls off sharply past 50\,m, while the elevated UAV viewpoint is weakest up close and strongest at long range. The two platforms are thus complementary rather than redundant, and late fusion inherits the stronger of the two at every range, yielding the most balanced performance across all distance bins. Distance-binned results at $\text{IoU}=0.70$ and $0.25$ are reported in Appendix~\ref{app:supplementary-detection} (Tables~\ref{tab:vic_iou070_bins} and \ref{tab:vic_iou025_bins}).

\emph{Sim-to-real transfer.}
Table~\ref{tab:main_results_pretraining} reports our main sim-to-real result. Pretraining on HERCULES synthetic data improves real-world DAIR-V2X detection by $+4.11$ Car 3D~AP40 over training from scratch at the strict $\text{IoU}=0.70$ threshold, with the same gain reflected in BEV. The two finetuning strategies, freezing the pretrained backbone during warm-up versus unfreezing throughout, reach effectively identical accuracy, indicating that the synthetic features transfer regardless of the unfreezing schedule. Table~\ref{tab:eval_crossval} reproduces this result under a second evaluation protocol (KITTI-style AP40 and DAIR-V2X continuous AP), confirming the improvement is not an artifact of the AP computation. Full results across all three fusion configurations and both IoU thresholds are reported in Appendix~\ref{app:sim2real-detection} (Tables~\ref{tab:sim2real_iou070_full} and \ref{tab:sim2real_iou050_full}).

\begin{figure}[!t]
  \centering

  \begin{subfigure}[t]{0.49\linewidth}
    \centering
    \includegraphics[width=\linewidth]{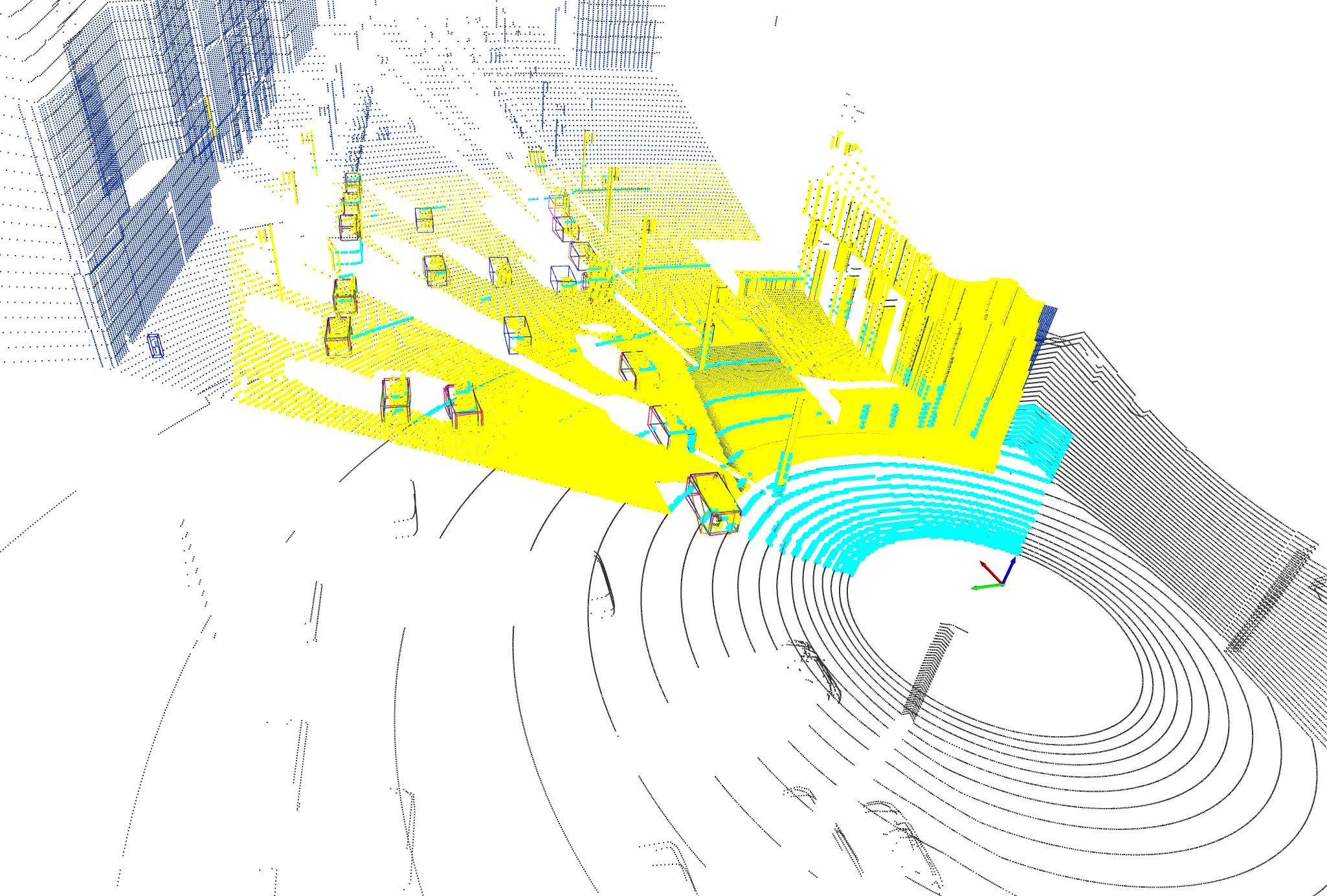}
    \label{fig:my2x2_a}
  \end{subfigure}\hfill
  \begin{subfigure}[t]{0.49\linewidth}
    \centering
    \includegraphics[width=\linewidth]{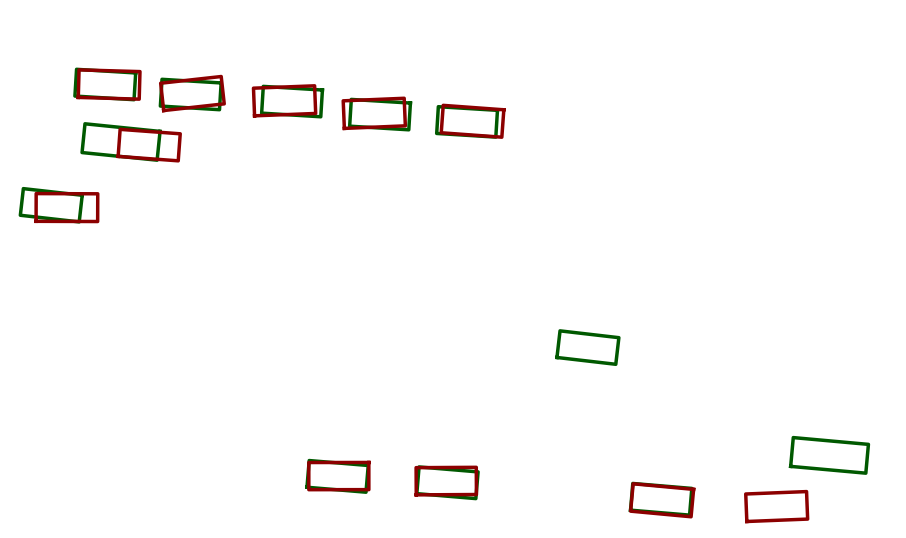}
    \label{fig:my2x2_b}
  \end{subfigure}

  \vspace{0.4em}

  \begin{subfigure}[t]{0.49\linewidth}
    \centering
    \includegraphics[width=\linewidth]{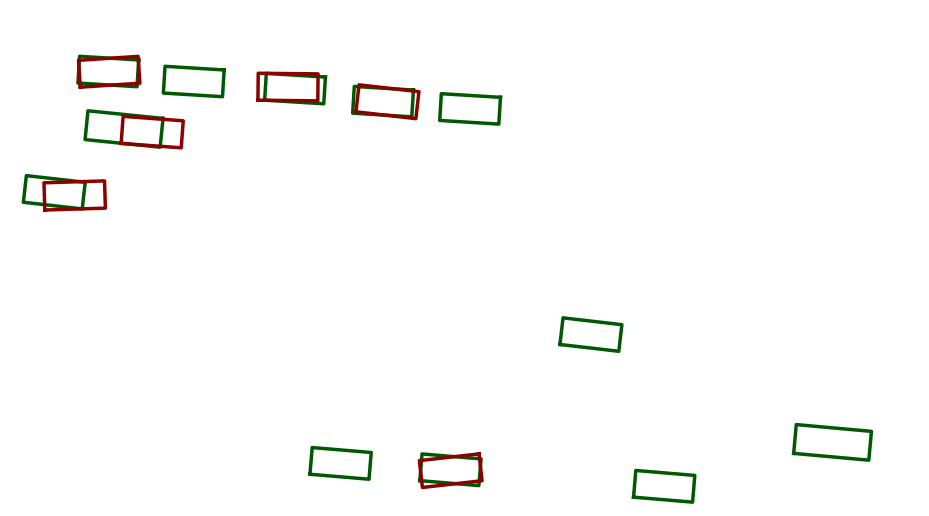}
    \label{fig:my2x2_c}
  \end{subfigure}\hfill
  \begin{subfigure}[t]{0.49\linewidth}
    \centering
    \includegraphics[width=\linewidth]{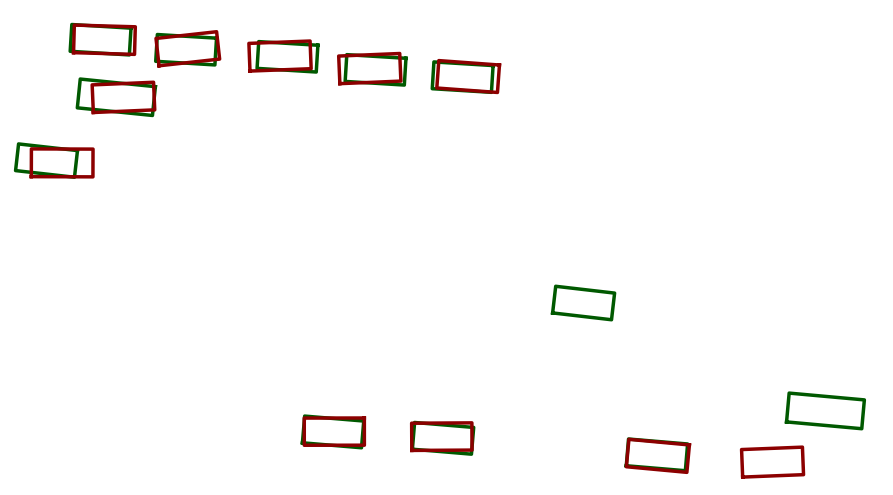}
    \label{fig:my2x2_d}
  \end{subfigure}

  \caption{Qualitative cooperative 3D Car detection late-fusion results on the DAIR-V2X test set. 
  (Top left)~Ground-truth annotations in 3D and BEV.
  (Top right)~Late-fusion detections combining vehicle and infrastructure views.
  (Bottom left)~Vehicle-only detections.
  (Bottom right)~Infrastructure-only detections.
  Late fusion recovers objects missed by individual agents, particularly in occluded and long-range regions.}
  \label{fig:my2x2}
\end{figure}

Table~\ref{tab:fusion_ablation} shows that the pretraining gain holds under cooperative fusion: late fusion improves over both single-agent detectors, and the improvement over training from scratch appears in all three configurations rather than in a single agent. The benefit of synthetic pretraining therefore carries through to the fused system. Figure~\ref{fig:my2x2} shows this qualitatively, with late fusion recovering objects that either agent alone misses in occluded and long-range regions.

\begin{table}[t!]
\centering
\scriptsize
\setlength{\tabcolsep}{4pt}
\renewcommand{\arraystretch}{1.15}
\caption{Sim-to-real transfer: late-fusion Car detection results of FT-Frozen compared with single-agent detection (Car 3D AP40 at IoU = 0.70). $\Delta$ is reported relative to From-Scratch trained on real data only, showing consistent improvements across all configurations with HERCULES pretraining.}
\label{tab:fusion_ablation}
\begin{tabular}{|l|c|c|}
\hline\hline
\textbf{Configuration} & \textbf{3D AP40} & \textbf{$\Delta$} \\
\hline
Vehicle-only            & 20.63 & $+1.98$      \\
Infrastructure-only     & 28.33 & $+0.06$  \\
\textbf{Late Fusion}    & \textbf{46.73} & $\mathbf{+2.11}$ \\
\hline\hline
\end{tabular}
\end{table}

\subsubsection{Discussion.} 
The gain comes from stronger features, not from how the model is trained or fused. Freezing the backbone for a warm-up versus unfreezing it from the start changes the final score by only 0.06~AP, so pretraining helps regardless of the unfreezing schedule. The same gain appears when we run the stock DAIR-V2X late-fusion pipeline unmodified, which puts the benefit in the per-agent detectors rather than the fusion step. The practical payoff is for late fusion specifically: early fusion can score higher on VIC-Sync but costs too much bandwidth to deploy, while simulation pretraining recovers much of that accuracy without any extra transmission between agents.

\subsubsection{Reproducibility details.} 
We release the HERCULES scenario, sensor configs (rates/FOVs), calibration files, and data-export scripts that produce the 6,000-pair pretraining set; training uses the same loss functions, augmentations, and schedules as the original PointPillars VIC baselines, and we retain DAIR-V2X's evaluation protocol and ranges.

\subsection{Closed-Loop Exploration}
\label{sec:active_perception}

The above two experiments replay fixed trajectories to keep the datasets repeatable. Here, we instead let a UAV--UGV team plan and re-plan online from its own observations, showing that HERCULES enables closed-loop simulation.

\noindent\subsubsection{Setup.} One UAV and one UGV operate in the desert environment with no pre-planned trajectory. Each robot tracks what it has seen by projecting its depth camera into the world using ground-truth poses (returns past a $12$\,m cutoff are dropped) and adding the observed surface to a shared $0.5$\,m voxel grid. Every $3$\,s, each robot picks a new short-horizon goal as the nearest unobserved cell within a local search radius and drives toward it: the UAV via \texttt{moveToPositionAsync} and the UGV via the proportional controller described in Sec.~\ref{sec:ugv_controller}. We test the two coordination patterns in Sec.~\ref{sec:motion_patterns}. In \emph{Complementary Coverage}, each robot's goal score is penalized for being close to the other, so the team spreads out. In \emph{Leader--Follower}, the UGV explores and the UAV is pinned to a fixed offset directly above it. This is intentionally a simple scheme: the goal is to show the closed-loop pipeline works end to end, not to propose a new exploration method. The shared clock and synchronized sensor pipeline are the same as in Sec.~\ref{sec:dataset_collection}.

\begin{figure}[t]
    \centering
    \begin{subfigure}[t]{0.49\linewidth}
        \centering
        \includegraphics[width=\linewidth]{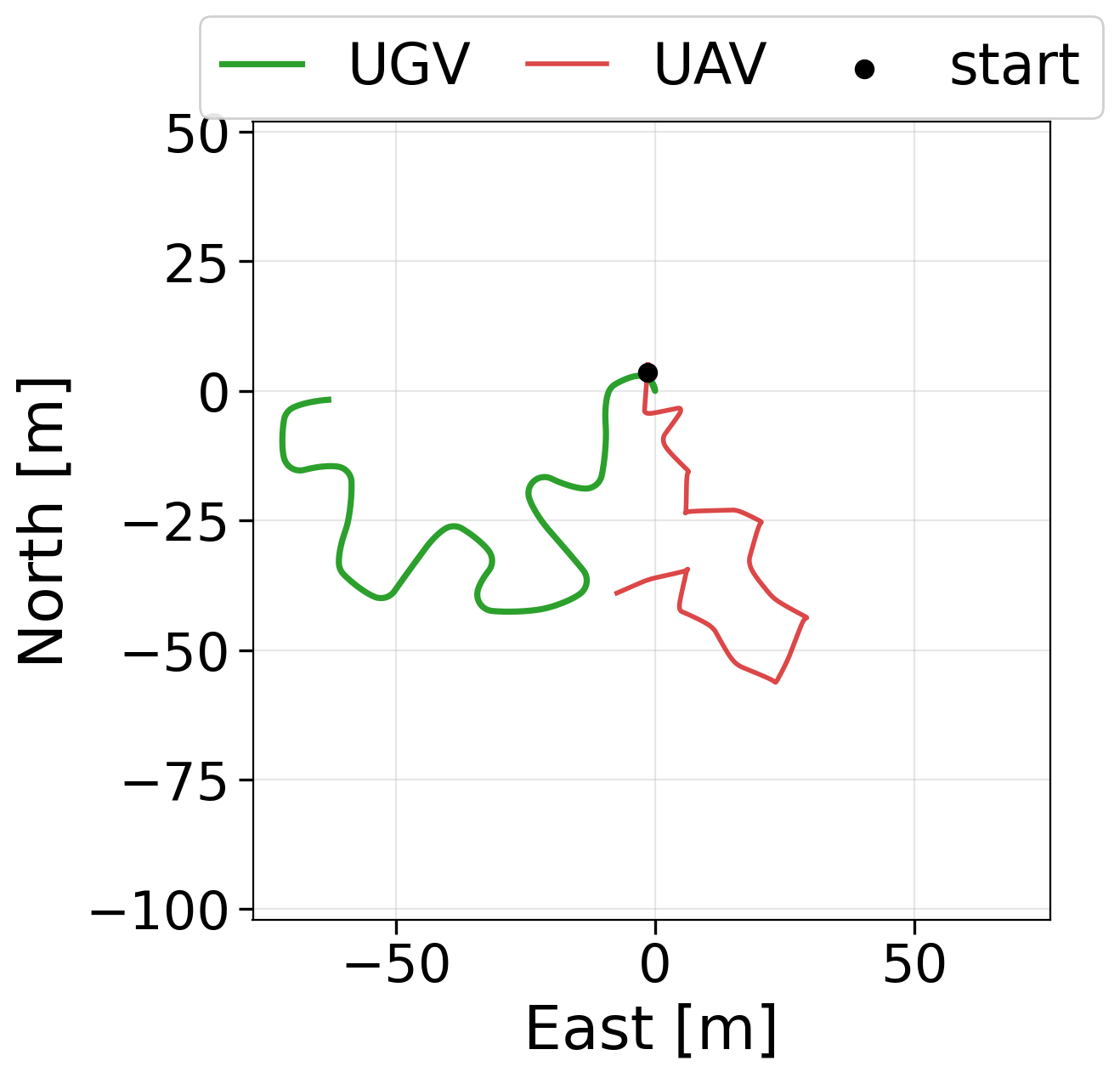}
        \caption{Complementary Coverage: the UGV and UAV diverge into separate regions.}
        \label{fig:ap_traj_comp}
    \end{subfigure}
    \hfill
    \begin{subfigure}[t]{0.49\linewidth}
        \centering
        \includegraphics[width=\linewidth]{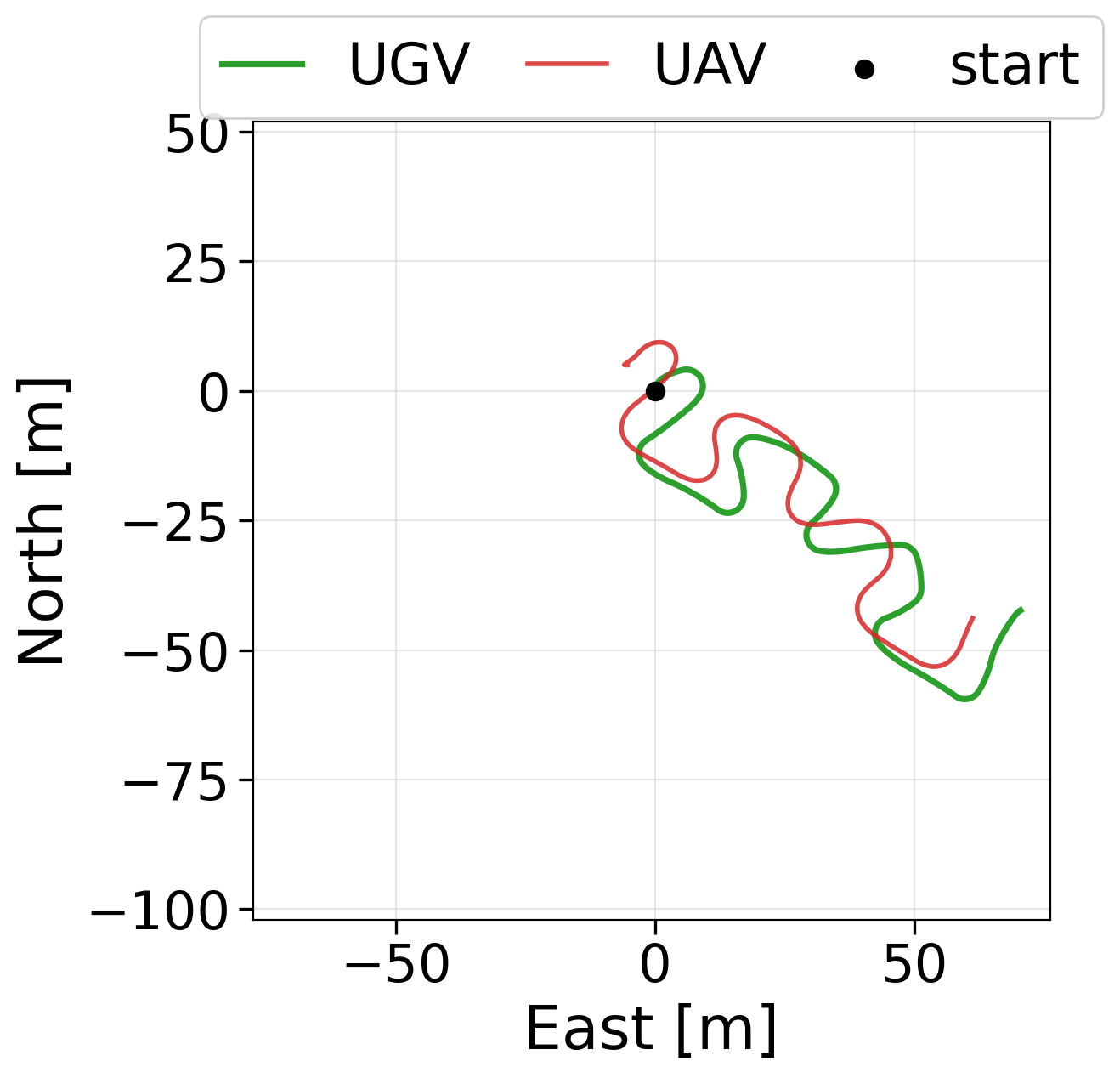}
        \caption{Leader--Follower: the UAV tracks the UGV, maintaining sensing overlap.}
        \label{fig:ap_traj_lf}
    \end{subfigure}
    \caption{Representative single-run UAV (red) and UGV (green) trajectories in the desert environment under the two coordination modes. The divergence in (a) versus the overlap in (b) is the qualitative mechanism behind the inter-agent overlap $\eta$ reported in Table~\ref{tab:active_perception}.}
    \label{fig:ap_trajectories}
\end{figure}

\begin{figure}[t!]
    \centering
    \includegraphics[width=\linewidth]{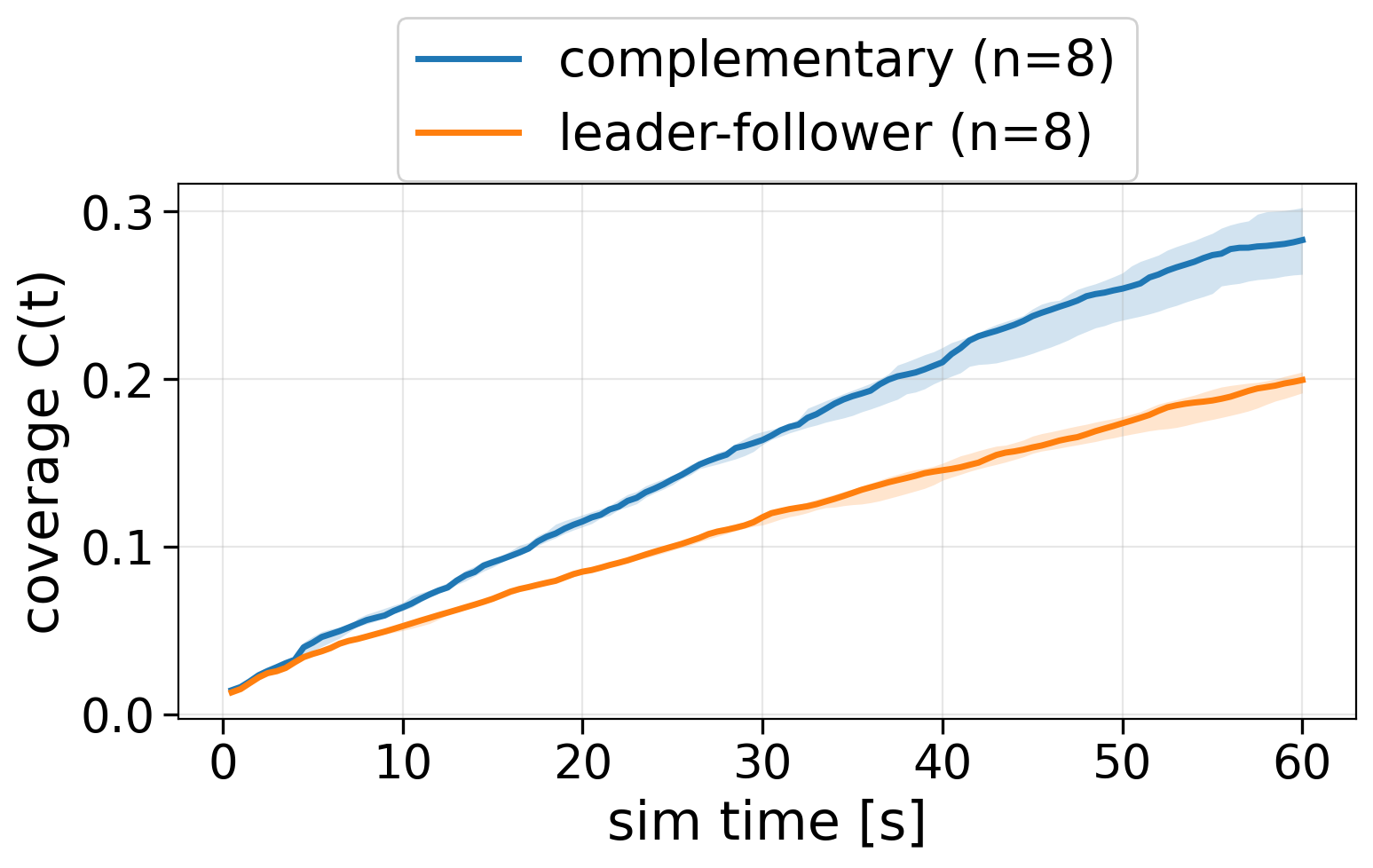}
    \caption{Ground-truth coverage $C(t)$ versus mission time for the heterogeneous UAV--UGV team in the desert environment, over $n=8$ seeds per mode (median, with shaded interquartile range). Complementary Coverage attains consistently higher coverage at equal mission time than Leader--Follower, with non-overlapping IQR bands throughout the run.}
    \label{fig:ap_coverage}
\end{figure}

\noindent\subsubsection{Metrics.} We report (i) the fraction of the ground-truth surface within the operating region that has been observed as a function of mission time, $C(t) = |\mathcal{M}_{\text{obs}}(t)| / |\mathcal{M}_{\text{gt}}|$; (ii) the inter-agent sensing-footprint overlap $\eta(t)$ averaged over the run, which should be low for Complementary Coverage and high for Leader--Follower; and (iii) the achieved real-time factor (RTF) of the closed-loop pipeline, which characterizes whether online planning and sensor streaming keep up with simulation time on the heterogeneous team.

\noindent\subsubsection{Results.} Figure~\ref{fig:ap_coverage} shows coverage over time for both modes across $n=8$ seeds. Complementary Coverage reaches a median final coverage of $0.283$ (IQR $[0.262, 0.302]$) versus $0.199$ (IQR $[0.192, 0.204]$) for Leader--Follower at $t=60$\,s, a relative gain of $42\%$, with non-overlapping interquartile bands throughout the run, confirming that dispersing the agents explores more of the map at equal mission time. The trajectory examples in Figure~\ref{fig:ap_trajectories} show the underlying behavior: the agents separate under Complementary Coverage and overlap under Leader--Follower. Table~\ref{tab:active_perception} reports the corresponding inter-agent sensing-footprint overlap $\eta$ and real-time factor: $\eta$ is low for Complementary Coverage ($0.060$) and high for Leader--Follower ($0.484$), as designed, and both modes sustain RTF $\geq 0.30$, confirming that HERCULES runs the heterogeneous team closed-loop at usable speed.

\noindent\subsubsection{Discussion.} The goal of this experiment is to demonstrate closed-loop operation in HERCULES, rather than to evaluate a specific exploration method. The greedy nearest-frontier scheme is intentionally simple and is distinct from the kinodynamic planner in Appendix~\ref{app:krrt-details}. The same heterogeneous core, mapping, planning, and control components used to generate the offline datasets in Sec.~\ref{sec:experiments-collab-slam} and \ref{sec:experiments-collab-perception} also run online, showing that HERCULES supports active-coordination research in addition to dataset collection.

\begin{table}[t]
\centering
\footnotesize
\setlength{\tabcolsep}{6pt}
\renewcommand{\arraystretch}{1.2}
\caption{Closed-loop multi-robot exploration in the desert environment ($n=8$ seeds; median, with IQR for coverage). $C(60\text{s})$ is final ground-truth coverage; $\eta$ is the time-averaged inter-agent sensing-footprint overlap (ground discs of radius $12$\,m); RTF is the achieved real-time factor.}
\label{tab:active_perception}
\begin{tabular}{|l|c|c|c|}
\hline
\textbf{Mode} & \textbf{$C(60\text{s})$} & \textbf{$\eta$} & \textbf{RTF} \\
\hline
Complementary Coverage & $0.283$ & $0.060$ & $0.31$ \\
Leader--Follower       & $0.199$ & $0.484$ & $0.32$ \\
\hline
\end{tabular}
\end{table}

\section{Conclusion}
We presented HERCULES, an open-source UE5-based simulator and data-collection tool for heterogeneous multi-robot autonomy that extends Cosys-AirSim with concurrent UAV--UGV operation, a shared navigation stack, derived thermal and low-light imaging, and photorealistic environments with dynamic agents and environmental phenomena. 
Three sets of experiments demonstrated HERCULES’s broad utility in multi-robot research. First, we benchmarked state-of-the-art multi-robot SLAM methods on diverse HERCULES sequences, demonstrating its ability to generate photorealistic, large-scale environments with varied sensing conditions for SLAM evaluation. The performance degradation of ROMAN on heterogeneous robot pairs underscores the need for more robust multi-robot SLAM methods, a challenge that HERCULES is designed to support as a foundational development tool. Second, the collaborative 3D detection experiment demonstrated the efficacy of HERCULES as a synthetic data generator for collaborative perception research, while the sim-to-real experiment further validated the high fidelity of HERCULES synthetic data. Finally, HERCULES can also be used as a closed-loop simulation for multi-robot planning, active perception, and exploration, as demonstrated in the last experiment. Together, these capabilities make advanced multi-robot research more accessible, reproducible, and scalable for the broader research community.

{
\expandafter\def\expandafter\UrlBreaks\expandafter{\UrlBreaks\do\/\do\-\do\.\do\a\do\b\do\c\do\d\do\e\do\f\do\g\do\h\do\i\do\j\do\k\do\l\do\m\do\n\do\o\do\p\do\q\do\r\do\s\do\t\do\u\do\v\do\w\do\x\do\y\do\z\do\A\do\B\do\C\do\D\do\E\do\F\do\G\do\H\do\I\do\J\do\K\do\L\do\M\do\N\do\O\do\P\do\Q\do\R\do\S\do\T\do\U\do\V\do\W\do\X\do\Y\do\Z\do\0\do\1\do\2\do\3\do\4\do\5\do\6\do\7\do\8\do\9}

\FloatBarrier

\begingroup
\raggedright
\bibliographystyle{template/SageH}
\bibliography{bib/IEEEfull, bib/strings-full, refs.bib}
\endgroup
}

\onecolumn
\appendix

\section{Supplementary Sim-to-Real Detection Results}
\label{app:sim2real-detection}
Tables~\ref{tab:sim2real_iou070_full} and \ref{tab:sim2real_iou050_full} extend the main-text sim-to-real results (Tables~\ref{tab:main_results_pretraining}--\ref{tab:fusion_ablation}) by reporting all three training strategies across all three fusion configurations at IoU=0.70 and IoU=0.50, respectively.

\begin{table}[H]
\centering
\scriptsize
\setlength{\tabcolsep}{3.0pt}
\renewcommand{\arraystretch}{1.15}
\caption{Sim-to-real transfer: Car detection on the DAIR-V2X test set at IoU=0.70 (strict).
All three training strategies and fusion configurations are shown.
KITTI-style AP11/AP40 and DAIR-V2X continuous AP are reported for both $AP_{3D}$ and $AP_{BEV}$.
$\Delta$ columns show change in AP40 (KITTI) or AP (DAIR-V2X) relative to from-scratch within each fusion type.
Bold marks the best result per fusion type.}
\label{tab:sim2real_iou070_full}
\resizebox{\textwidth}{!}{%
\begin{tabular}{|l|l|ccc|ccc|cc|cc|c|}
\hline\hline
\multirow{2}{*}{\textbf{Fusion}} &
\multirow{2}{*}{\textbf{Method}} &
\multicolumn{3}{c|}{\textbf{KITTI $AP_{3D}$}} &
\multicolumn{3}{c|}{\textbf{KITTI $AP_{BEV}$}} &
\multicolumn{2}{c|}{\textbf{DAIR-V2X $AP_{3D}$}} &
\multicolumn{2}{c|}{\textbf{DAIR-V2X $AP_{BEV}$}} &
\multirow{2}{*}{\textbf{Ep.}} \\
\cline{3-12}
& &
\textbf{AP11} & \textbf{AP40} & \textbf{$\Delta$} &
\textbf{AP11} & \textbf{AP40} & \textbf{$\Delta$} &
\textbf{AP} & \textbf{$\Delta$} &
\textbf{AP} & \textbf{$\Delta$} & \\
\hline
\multirow{3}{*}{Vehicle-only}
  & From-Scratch         & 24.56 & 18.65 & ---     & 26.90 & 24.54 & ---     & 18.75 & ---     & 23.90 & ---     & 19 \\
  & FT-Unfrozen            & \textbf{24.79} & \textbf{20.64} & $\mathbf{+1.99}$ & 26.91 & 24.58 & $+0.04$ & 19.11 & $+0.36$ & 24.27 & $+0.37$ & 19 \\
  & \textbf{FT-Frozen}   & 24.71 & 20.63 & $+1.98$ & \textbf{26.95} & \textbf{24.63} & $\mathbf{+0.09}$ & \textbf{19.13} & $\mathbf{+0.38}$ & \textbf{24.30} & $\mathbf{+0.40}$ & 20 \\
\hline
\multirow{3}{*}{Infra-only}
  & From-Scratch         & 26.26 & 28.27 & ---     & 35.87 & 32.10 & ---     & 26.70 & ---     & 31.21 & ---     & 19 \\
  & \textbf{FT-Unfrozen}   & 26.28 & \textbf{28.35} & $\mathbf{+0.08}$ & \textbf{35.91} & \textbf{32.13} & $\mathbf{+0.03}$ & 26.85 & $+0.15$ & \textbf{31.31} & $\mathbf{+0.10}$ & 19 \\
  & FT-Frozen            & \textbf{26.30} & 28.33 & $+0.06$ & 35.84 & 32.08 & $-0.02$ & \textbf{26.86} & $\mathbf{+0.16}$ & 31.14 & $-0.07$ & 20 \\
\hline
\multirow{3}{*}{Late Fusion}
  & From-Scratch         & 43.13 & 44.62 & ---     & 53.86 & 54.21 & ---     & 45.36 & ---     & 54.87 & ---     & 19 \\
  & FT-Unfrozen            & 43.21 & 46.67 & $+2.05$ & 53.86 & \textbf{54.24} & $+0.03$ & 45.76 & $+0.40$ & \textbf{55.33} & $\mathbf{+0.46}$ & 19 \\
  & \textbf{FT-Frozen}   & \textbf{43.26} & \textbf{46.73} & $\mathbf{+2.11}$ & \textbf{53.87} & 54.23 & $+0.02$ & \textbf{45.87} & $\mathbf{+0.51}$ & 55.13 & $+0.26$ & 20 \\
\hline\hline
\end{tabular}%
}
\end{table}

\begin{table}[H]
\centering
\scriptsize
\setlength{\tabcolsep}{3.0pt}
\renewcommand{\arraystretch}{1.15}
\caption{Sim-to-real transfer: Car detection on the DAIR-V2X test set at IoU=0.50 (loose).
All three training strategies and fusion configurations are shown.
KITTI-style AP11/AP40 and DAIR-V2X continuous AP are reported for both $AP_{3D}$ and $AP_{BEV}$.
$\Delta$ columns show change in AP40 (KITTI) or AP (DAIR-V2X) relative to from-scratch within each fusion type.
Bold marks the best result per fusion type.}
\label{tab:sim2real_iou050_full}
\resizebox{\textwidth}{!}{%
\begin{tabular}{|l|l|ccc|ccc|cc|cc|c|}
\hline\hline
\multirow{2}{*}{\textbf{Fusion}} &
\multirow{2}{*}{\textbf{Method}} &
\multicolumn{3}{c|}{\textbf{KITTI $AP_{3D}$}} &
\multicolumn{3}{c|}{\textbf{KITTI $AP_{BEV}$}} &
\multicolumn{2}{c|}{\textbf{DAIR-V2X $AP_{3D}$}} &
\multicolumn{2}{c|}{\textbf{DAIR-V2X $AP_{BEV}$}} &
\multirow{2}{*}{\textbf{Ep.}} \\
\cline{3-12}
& &
\textbf{AP11} & \textbf{AP40} & \textbf{$\Delta$} &
\textbf{AP11} & \textbf{AP40} & \textbf{$\Delta$} &
\textbf{AP} & \textbf{$\Delta$} &
\textbf{AP} & \textbf{$\Delta$} & \\
\hline
\multirow{3}{*}{Vehicle-only}
  & From-Scratch         & 26.93 & 24.61 & ---     & 27.14 & 24.84 & ---     & 24.34 & ---     & 25.38 & ---     & 19 \\
  & FT-Unfrozen            & 26.96 & 24.62 & $+0.01$ & 27.14 & \textbf{27.19} & $\mathbf{+2.35}$ & 24.70 & $+0.36$ & \textbf{25.80} & $\mathbf{+0.42}$ & 19 \\
  & \textbf{FT-Frozen}   & \textbf{26.97} & \textbf{24.66} & $\mathbf{+0.05}$ & \textbf{27.17} & 27.19 & $+2.35$ & \textbf{24.74} & $\mathbf{+0.40}$ & 25.76 & $+0.38$ & 20 \\
\hline
\multirow{3}{*}{Infra-only}
  & From-Scratch         & \textbf{36.05} & \textbf{32.25} & ---     & 36.32 & 32.47 & ---     & 31.84 & ---     & 32.78 & ---     & 19 \\
  & FT-Unfrozen            & 36.04 & 32.23 & $-0.02$ & 36.32 & 32.47 & $0.00$  & \textbf{31.87} & $\mathbf{+0.03}$ & 32.83 & $+0.05$ & 19 \\
  & FT-Frozen            & 36.03 & 32.21 & $-0.04$ & 36.32 & 32.47 & $0.00$  & 31.85 & $+0.01$ & \textbf{32.92} & $\mathbf{+0.14}$ & 20 \\
\hline
\multirow{3}{*}{Late Fusion}
  & From-Scratch         & 54.05 & 56.72 & ---     & 54.44 & 57.30 & ---     & 55.93 & ---     & 57.85 & ---     & 19 \\
  & FT-Unfrozen            & 54.01 & \textbf{56.75} & $+0.03$ & 54.43 & 57.30 & $0.00$  & \textbf{56.33} & $\mathbf{+0.40}$ & \textbf{58.35} & $\mathbf{+0.50}$ & 19 \\
  & \textbf{FT-Frozen}   & \textbf{54.03} & 56.75 & $\mathbf{+0.03}$ & \textbf{54.44} & \textbf{57.31} & $\mathbf{+0.01}$ & 56.30 & $+0.37$ & 58.33 & $+0.48$ & 20 \\
\hline\hline
\end{tabular}%
}
\end{table}

\section{Supplementary Sim-to-Sim Detection Results}
\label{app:supplementary-detection}
Tables~\ref{tab:vic_iou070_bins} and \ref{tab:vic_iou025_bins} complement Table~\ref{tab:vic_iou050_1class} by providing distance-binned results for the sim-to-sim cooperative Car detection baseline at IoU=0.70 and IoU=0.25, respectively.

\begin{table}[H]
\centering
\scriptsize
\setlength{\tabcolsep}{3.2pt}
\renewcommand{\arraystretch}{1.15}
\caption{Sim-to-sim baseline: cooperative Car detection at IoU=0.70, distance-binned. AP (AP11/AP40) for $AP_{3D}$ and $AP_{BEV}$.}
\label{tab:vic_iou070_bins}
\resizebox{\textwidth}{!}{%
\begin{tabular}{|l|cc|cc|cc|cc|cc|cc|}
\hline\hline
\multirow{3}{*}{\textbf{Fusion}} &
\multicolumn{6}{c|}{\textbf{$AP_{3D}$ (IoU=0.70)}} &
\multicolumn{6}{c|}{\textbf{$AP_{BEV}$ (IoU=0.70)}} \\
\cline{2-13}
&
\multicolumn{2}{c|}{\textbf{0--30\,m}} &
\multicolumn{2}{c|}{\textbf{30--50\,m}} &
\multicolumn{2}{c|}{\textbf{50--100\,m}} &
\multicolumn{2}{c|}{\textbf{0--30\,m}} &
\multicolumn{2}{c|}{\textbf{30--50\,m}} &
\multicolumn{2}{c|}{\textbf{50--100\,m}} \\
\cline{2-13}
&
\textbf{AP11} & \textbf{AP40} &
\textbf{AP11} & \textbf{AP40} &
\textbf{AP11} & \textbf{AP40} &
\textbf{AP11} & \textbf{AP40} &
\textbf{AP11} & \textbf{AP40} &
\textbf{AP11} & \textbf{AP40} \\
\hline
Veh.-only
  & 50.58 & 52.21 & 20.66 & 20.33 &  9.09 & 4.45
  & 62.98 & 64.20 & 40.47 & 41.28 & 16.04 & 12.73 \\
\hline
Inf.-only
  & 44.52 & 46.05 & 41.85 & 42.79 & 60.42 & 57.33
  & 45.27 & 49.37 & 51.58 & 49.75 & 62.03 & 60.99 \\
\hline
Late Fusion
  & 57.02 & 54.05 & 45.61 & 42.21 & 51.50 & 49.63
  & 62.16 & 65.46 & 57.25 & 56.12 & 53.16 & 57.71 \\
\hline\hline
\end{tabular}%
}
\end{table}

\begin{table}[H]
\centering
\scriptsize
\setlength{\tabcolsep}{3.2pt}
\renewcommand{\arraystretch}{1.15}
\caption{Sim-to-sim baseline: cooperative Car detection at IoU=0.25, distance-binned. AP (AP11/AP40) for $AP_{3D}$ and $AP_{BEV}$.}
\label{tab:vic_iou025_bins}
\resizebox{\textwidth}{!}{%
\begin{tabular}{|l|cc|cc|cc|cc|cc|cc|}
\hline\hline
\multirow{3}{*}{\textbf{Fusion}} &
\multicolumn{6}{c|}{\textbf{$AP_{3D}$ (IoU=0.25)}} &
\multicolumn{6}{c|}{\textbf{$AP_{BEV}$ (IoU=0.25)}} \\
\cline{2-13}
&
\multicolumn{2}{c|}{\textbf{0--30\,m}} &
\multicolumn{2}{c|}{\textbf{30--50\,m}} &
\multicolumn{2}{c|}{\textbf{50--100\,m}} &
\multicolumn{2}{c|}{\textbf{0--30\,m}} &
\multicolumn{2}{c|}{\textbf{30--50\,m}} &
\multicolumn{2}{c|}{\textbf{50--100\,m}} \\
\cline{2-13}
&
\textbf{AP11} & \textbf{AP40} &
\textbf{AP11} & \textbf{AP40} &
\textbf{AP11} & \textbf{AP40} &
\textbf{AP11} & \textbf{AP40} &
\textbf{AP11} & \textbf{AP40} &
\textbf{AP11} & \textbf{AP40} \\
\hline
Veh.-only
  & 72.56 & 72.38 & 62.14 & 61.05 & 17.88 & 19.02
  & 72.56 & 74.80 & 62.18 & 61.08 & 17.88 & 19.07 \\
\hline
Inf.-only
  & 54.53 & 54.99 & 63.31 & 62.15 & 72.00 & 69.33
  & 54.53 & 54.99 & 63.31 & 62.15 & 72.00 & 69.33 \\
\hline
Late Fusion
  & 71.41 & 73.53 & 78.82 & 77.22 & 62.87 & 66.52
  & 71.41 & 73.53 & 78.82 & 77.22 & 62.87 & 66.52 \\
\hline\hline
\end{tabular}%
}
\end{table}

\section{Kinodynamic RRT Planner Formulation}
\label{app:krrt-details}

This appendix provides the full mathematical formulation of the kinodynamic RRT (KRRT) planner summarized in Sec.~\ref{sec:kino-traj-planning}.

\subsection*{Dynamics used for propagation}
\emph{UGV (unicycle):}
\begin{align}
  \dot{\mathbf{x}}^{g}(t) &=
  \begin{bmatrix}
    v^{g}(t)\cos\theta^{g}(t)\\
    v^{g}(t)\sin\theta^{g}(t)\\
    \omega^{g}(t)
  \end{bmatrix}, \quad
  \mathbf{x}^{g} = [x^{g},y^{g},\theta^{g}]^{\!\top}, \\
  |v^{g}(t)| &\le v^{\max}_{g}, \qquad
  |\omega^{g}(t)| \le \omega^{\max}_{g}.
  \label{eq:app-rrt-ugv}
\end{align}

\emph{UAV (planar double-integrator at altitude $h_{\mathrm{UAV}}$):}
\begin{align}
  \ddot{\mathbf{p}}^{a}\!(t)=\mathbf{u}^{a}\!(t),\quad
  \|\mathbf{u}^{a}\|\!\le\! a^{\max}_{a},\;
  \|\dot{\mathbf{p}}^{a}\|\!\le\! v^{\max}_{a}, \\
  \mathbf{p}^{a}\!=\![x^{a},y^{a}]^{\!\top} \label{eq:app-rrt-uav}
\end{align}

With time step $\Delta t$, we use the discrete-time update
$\mathbf{x}_{k+1} = F(\mathbf{x}_{k},\mathbf{u}_{k})$
to numerically integrate the continuous-time dynamics.

\subsection*{Task-aware sampling}
At each iteration, a target $x_{\mathrm{rand}}$ is drawn from a mixture distribution:
\begin{align}
  x_{\mathrm{rand}}\!\sim\!
  \begin{cases}
    \text{GoalDistribution} & \text{w.p. } p_{\mathrm{goal}},\\
    \text{FrontierDistribution}(\mathbf{o}_{k}) & \text{w.p. } p_{\mathrm{front}},\\
    \text{Uniform}(X_{\mathrm{free}}) & \text{w.p. } 1-p_{\mathrm{goal}}-p_{\mathrm{front}},
  \end{cases}
  \label{eq:app-rrt-mixture}
\end{align}
where $\mathbf{o}_{k}$ is the occupancy map at step $k$, and frontier samples lie near known/unknown boundaries (optionally weighted by a local information score).

For SLAM-oriented dataset collection, a fraction of the sampling mass is assigned to a \emph{revisit distribution} concentrated on previously observed landmarks or keyframe poses, so that KRRT periodically steers robots back through already-mapped regions. This induces both intra-robot loop closures and inter-robot trajectory overlap.

\subsection*{Nearest neighbor and steering metric}
Let $\mathcal{V}$ be current tree vertices. We select
\begin{align}
x_{\mathrm{near}} &= \arg\min_{x\in\mathcal{V}}\rho(x,x_{\mathrm{rand}}),
\end{align}
with
\begin{align}
\begin{split}
  \rho(x_{1},x_{2}) 
    &= w_{p}\,\|p_{1}-p_{2}\|_{2}
     + w_{\theta}\,d_{\!\angle}(\theta_{1},\theta_{2}) \\
    &\quad+ w_{v}\,\|v_{1}-v_{2}\|_{2}
\end{split} \label{eq:app-rrt-metric}
\end{align}
where $p$ is planar position, $\theta$ is heading (UGV only), $v$ is speed (when modeled), $d_{\angle}$ is wrapped angular distance, and $w_{p},w_{\theta},w_{v}\!\ge\!0$ are platform-specific weights.

\subsection*{Steering objective with safety and information}
From $x_{\mathrm{near}}$, candidate controls $u$ and durations $\tau$ are scored by
\begin{align}
  J(u,\tau)
  = \alpha\!\left[-\rho(\tilde{x},x_{\mathrm{rand}})\right]
    + \beta\,\widehat{\mathcal{I}}(\tilde{x}\mid \mathbf{o}_{k})
    + \gamma\,\phi\!\big(D(\tilde{x})\big),
  \label{eq:app-rrt-J}
\end{align}
where $\tilde{x}=\Phi(x_{\mathrm{near}},u,\tau)$ is the forward-integrated state; 
$\alpha,\beta,\gamma\!\ge\!0$ weight target-proximity, information, and safety; 
$\widehat{\mathcal{I}}$ counts unknown cells expected within the sensing footprint (fast information surrogate via ray-casting); 
$D(\cdot)$ is obstacle clearance obtained from an Euclidean signed distance field (SDF) derived from OctoMap/elevation; and
\begin{align}
  \phi(d) = \min\{d-r_{\mathrm{safe}},\,0\}
\end{align}
penalizes proximity to obstacles for safety radius $r_{\mathrm{safe}}$.
The optimal control is
\begin{align}
  (u^{\star},\tau^{\star})
  &\in \arg\max_{u\in\mathcal{U},\,\tau\in\mathcal{T}} J(u,\tau) \\
  &\text{s.t.}\;\;
    D\!\big(\Phi(x_{\mathrm{near}},u,t)\big)\!\ge\! r_{\mathrm{safe}},
    \;\;\forall t\!\in\![0,\tau]. \label{eq:app-rrt-argmax}
\end{align}

\subsection*{Adaptive step and mode-aware bias}
Steps are shortened in clutter and lengthened in open spaces via
\begin{align}
  \tau = \min\{\tau_{\max},\, \kappa\, D(x_{\mathrm{near}})\}.
  \label{eq:app-rrt-adapt}
\end{align}
Mode logic biases \eqref{eq:app-rrt-mixture} and \eqref{eq:app-rrt-J}: in Complementary Coverage we set $p_{\mathrm{front}}>0$ and $\beta>0$ to encourage frontier sampling, while in Leader--Follower we set $p_{\mathrm{goal}}\!\gg\!0$ toward UGV waypoints and augment $J$ with a visibility/overlap bonus.

\subsection*{Information surrogate}
\begin{align}
  \widehat{\mathcal{I}}\!\big(\tilde{x}\mid\mathbf{o}_{k}\big)
  &= \sum_{j\in\mathcal{N}(\tilde{x},\rho_{\mathrm{sens}})}
     \mathds{1}_{\{o^{j}_{k}=-1\}}, \label{eq:app-rrt-gain}
\end{align}
i.e., the count of unknown cells within sensor radius at $\tilde{x}$.

\section{UGV Pure-Pursuit Controller Formulation}
\label{app:ugv-controller-details}

This appendix provides the full mathematical formulation of the UGV pure-pursuit controller summarized in Sec.~\ref{sec:ugv_controller}.

Let $\gamma(s)$ be the reference path parameterized by arc-length $s$. At time $t$, let $s^\star(t)$ be the index of the closest point to the UGV position $\mathbf{p}^{g}(t)=[x^{g}(t),y^{g}(t)]^{\!\top}$. The forward lookahead target is selected with a distance threshold $L_{\text{la}}>0$:
\begin{align}
  s_{\text{la}}(t)
  &= \min\!\Big\{s\ge s^\star(t)\;:\;
      \big\|\gamma(s)-\mathbf{p}^{g}(t)\big\| \ge L_{\text{la}}\Big\}, \nonumber\\[-0.25ex]
  \mathbf{p}_{\text{la}}(t) &= \gamma\!\big(s_{\text{la}}(t)\big),
  \label{eq:app-pp-lookahead}
\end{align}
with forward-only index stepping to avoid oscillation. The heading error toward the target is
\begin{align}
  \begin{split}
    e_{\psi}(t) \;=\;
    \operatorname{wrap}_{(-\pi,\pi]}\!\Big(
      \operatorname{atan2}\!\big((\mathbf{p}_{\text{la}}\!-\!\mathbf{p}^{g})_{y}, \\
      \qquad\qquad (\mathbf{p}_{\text{la}}\!-\!\mathbf{p}^{g})_{x}\big)
      - \psi^{g}(t)\Big)
  \end{split} \label{eq:app-pp-heading}
\end{align}
where $\psi^{g}(t)$ is the UGV yaw. Steering (yaw-rate) and speed commands are
\begin{align}
  \omega^{g}(t)
  &= \operatorname{sat}_{[-\omega^{\max}_{g},\,\omega^{\max}_{g}]}
     \!\big(K_{p,\text{steer}}\,e_{\psi}(t)\big), \label{eq:app-pp-steer}\\[-0.5ex]
  &\quad K_{p,\text{steer}}\approx 1.0 \nonumber \\[1ex]
  v^{g}(t)
  &= \operatorname{sat}_{[0,\,v^{\max}_{g}]}
     \!\Big(v_{\text{des}}(t)
     + K_{p,\text{spd}}\big(v_{\text{des}}(t)-\|\dot{\mathbf{p}}^{g}(t)\|\big)\Big), 
     \label{eq:app-pp-speed}\\[-0.5ex]
  &\quad K_{p,\text{spd}}\approx 0.5 \nonumber
\end{align}
where $v_{\text{des}}(t)$ is the desired cruise speed for the current path segment. The proportional gains $K_{p,\text{steer}}$ and $K_{p,\text{spd}}$ regulate heading and speed responses: higher values yield faster convergence but may increase oscillation on low-friction surfaces. The chosen nominal values provide stable tracking across terrain types and can be tuned based on maximum curvature and velocity bounds.

Commands are issued at a fixed rate $f_{\text{ctrl}}$ until end-of-path or timeout, after which a braking command is applied. This controller is consistent with the unicycle dynamics used in planning: the planner ensures curvature and velocity feasibility, while the tracker provides geometric convergence without requiring state-feedback linearization.

\textit{Implementation notes.}
The steering cap is $\delta_{\max}\!\approx\!0.5$\,rad, and the lookahead distance $L_{\text{la}}$ is proportional to speed (smaller at low speed for accuracy, larger at high speed for stability). 
Saturations in \eqref{eq:app-pp-steer}--\eqref{eq:app-pp-speed} prevent excessive control actions on low-friction terrain. 
The same controller is used for both the Husky-style UGV and the simulated SUV, with only geometric parameters (wheelbase, steering limits) adjusted.

\end{document}